\RequirePackage{fix-cm}
\documentclass[twocolumn]{svjour3}          
\smartqed  
\usepackage[usenames,dvipsnames]{xcolor}
\usepackage{graphicx}
\usepackage[draft]{hyperref}
\usepackage{svg}
\usepackage[space]{cite}
\usepackage{placeins}
\usepackage{subfig}
\usepackage{tabularx}
\usepackage{booktabs}
\usepackage{courier}
\usepackage{multirow}
\usepackage{pdflscape}
\usepackage{afterpage}
\usepackage[clockwise]{rotating}
\usepackage{morefloats}
\usepackage{breakcites}
\usepackage{etoolbox}

\newtoggle{smallfigs}
\toggletrue{smallfigs}

\newcommand{\ignore}[1]{}

\newcommand{\attribute}[1]{\texttt{{#1}}}
\newcommand{\object}[1]{\texttt{{#1}}}
\newcommand{\qa}[1]{\texttt{{#1}}}
\newcommand{\synset}[1]{\texttt{{#1}}}
\newcommand{\predicate}[1]{\texttt{{#1}}}
\newcommand{\relationship}[3]{\textit{#2}(\textit{#1}, \textit{#3})}

\hypersetup{final}

\begin{document}
\sloppy
\title{Visual Genome}
\subtitle{Connecting Language and Vision Using Crowdsourced Dense Image Annotations}


\author{Ranjay Krishna        \and
        Yuke Zhu       \and
        Oliver Groth       \and
        Justin Johnson \\       \and
        Kenji Hata      \and
        Joshua Kravitz      \and
        Stephanie Chen       \and
        Yannis Kalantidis \\      \and
        Li-Jia Li     \and
        David A. Shamma       \and
        Michael S. Bernstein       \and
        Li Fei-Fei
}


\institute{Ranjay Krishna \at
              Stanford University, Stanford, CA, USA \\
              \email{ranjaykrishna@cs.stanford.edu}
            \and
           Yuke Zhu \at
              Stanford University, Stanford, CA, USA
            \and
           Oliver Groth \at
              Dresden University of Technology, Dresden, Germany
            \and
           Justin Johnson \at
              Stanford University, Stanford, CA, USA
            \and
           Kenji Hata \at
              Stanford University, Stanford, CA, USA
            \and
           Joshua Kravitz \at
              Stanford University, Stanford, CA, USA
            \and
           Stephanie Chen \at
              Stanford University, Stanford, CA, USA
            \and
           Yannis Kalantidis \at
              Yahoo Inc., San Francisco, CA, USA
            \and
           Li-Jia Li\at
              Snapchat Inc., Los Angeles, CA, USA
            \and
           David A. Shamma \at
              Yahoo Inc., San Francisco, CA, USA
            \and
           Michael S. Bernstein \at
              Stanford University, Stanford, CA, USA
            \and
           Li Fei-Fei \at
              Stanford University, Stanford, CA, USA
}

\date{Received: date / Accepted: date}

\maketitle

\begin{abstract}
Despite progress in perceptual tasks such as image classification, computers still perform poorly on cognitive tasks such as image description and question answering. Cognition is core to tasks that involve not just recognizing, but reasoning about our visual world. However, models used to tackle the rich content in images for cognitive tasks are still being trained using the same datasets designed for perceptual tasks. To achieve success at cognitive tasks, models need to understand the interactions and relationships between objects in an image. When asked ``What vehicle is the person riding?'', computers will need to identify the objects in an image as well as the relationships \textit{riding(man, carriage)} and \textit{pulling(horse, carriage)} in order to answer correctly that ``the person is riding a horse-drawn carriage.''

In this paper, we present the Visual Genome dataset to enable the modeling of such relationships. We collect dense annotations of objects, attributes, and relationships within each image to learn these models. Specifically, our dataset contains over $100$K images where each image has an average of $21$ objects, $18$ attributes, and $18$ pairwise relationships between objects. 
We canonicalize the objects, attributes, relationships, and noun phrases in region descriptions and questions answer pairs to WordNet synsets. Together, these annotations represent the densest and largest dataset of image descriptions, objects, attributes, relationships, and question answers.

\keywords{Computer Vision \and Dataset \and Image \and Scene Graph \and Question Answering \and Objects \and Attributes \and Relationships \and Knowledge \and Language \and Crowdsourcing}
\end{abstract}


\section{Introduction}
\label{sec:introduction}

\begin{figure*}[t]%
    \centering
    \includegraphics[width=0.82\textwidth]{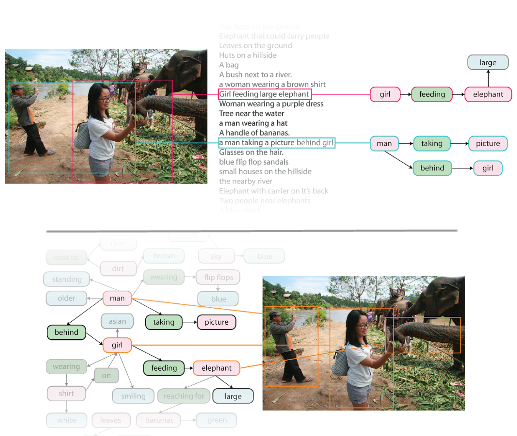}
    \caption{An overview of the data needed to move from perceptual awareness to cognitive understanding of images. We present a dataset of images densely annotated with numerous region descriptions, objects, attributes, and relationships. Region descriptions (e.g. ``girl feeding large elephant'' and ``a man taking a picture behind girl'') are shown (top). The objects (e.g. \object{elephant}), attributes (e.g. \attribute{large}) and relationships (e.g. \predicate{feeding}) are shown (bottom). Our dataset also contains image related question answer pairs (not shown).}%
    \label{fig:pipeline}%
\end{figure*}

A holy grail of computer vision is the complete understanding of visual scenes: a model that is able to name and detect objects, describe their attributes, and recognize their relationships and interactions. Understanding scenes would enable important applications such as image search, question answering, and robotic interactions. Much pro\-gress has been made in recent years towards this goal, including image classification~\cite{deng2009imagenet, perronnin2010improving, simonyan2014very, krizhevsky2012imagenet, szegedy2014going} and object detection~\cite{everingham2010pascal, girshick2014rich, sermanet2013overfeat, girshick2015fast, ren2015faster}. An important contributing factor is the availability of a large amount of data that drives the statistical models that underpin today's advances in computational visual understanding. While the progress is exciting, we are still far from reaching the goal of comprehensive scene understanding. As Figure~\ref{fig:pipeline} shows, existing models would be able to detect discreet objects in a photo but would not be able to explain their interactions or the relationships between them. Such explanations tend to be \textit{cognitive} in nature, integrating \textit{perceptual} information into conclusions about the relationships between objects in a scene~\cite{bruner1990culture, firestone2015cognition}. A cognitive understanding of our visual world thus requires that we complement computers' ability to detect objects with abilities to describe those objects~\cite{isola2015discovering} and understand their interactions within a scene~\cite{sadeghi2011recognition}.

There is an increasing effort to put together the next generation of datasets to serve as training and benchmarking datasets for these deeper, cognitive scene understanding and reasoning tasks, the most notable being MS-COCO~\cite{lin2014microsoft} and VQA~\cite{antol2015vqa}. The MS-COCO dataset consists of $300$K real-world photos collected from Flickr. For each image, there is pixel-level segmentation of $91$ object classes (when present) and $5$ independent, user-generated sentences describing the scene. VQA adds to this a set of $614$K question-answer pairs related to the visual contents of each image (see more details in Section~\ref{sec:datasets}). With this information, MS-COCO and VQA provide a fertile training and testing ground for models aimed at tasks for accurate object detection, segmentation, and summary-level image captioning~\cite{kiros2014multimodal, mao2014explain, karpathy2014deep, vinyals2014show} as well as basic QA~\cite{ren2015image, antol2015vqa, malinowski2015ask, gao2015you, malinowski2014multi}. For example, a state-of-the-art model~\cite{karpathy2014deep} provides a description of one MS-COCO image in Figure~\ref{fig:pipeline} as ``two men are standing next to an elephant.'' But what is missing is the further understanding of where each object is, what each person is doing, what the relationship between the person and elephant is, etc. Without such relationships, these models fail to differentiate this image from other images of people next to elephants.

To understand images thoroughly, we believe three key elements need to be added to existing datasets: a \textbf{grounding of visual concepts to language}~\cite{kiros2014multimodal}, a more \textbf{complete set of descriptions and QAs} for each image based on multiple image regions~\cite{Johnson2015CVPR}, and a \textbf{formalized representation} of the components of an image~\cite{hayes1978naive}. In the spirit of mapping out this complete information of the visual world, we introduce the Visual Genome dataset. The first release of the Visual Ge\-nome dataset uses $108,249$ images from the intersection of the YFCC100M~\cite{thomee2015yfcc100m} and MS-COCO~\cite{lin2014microsoft}. Section~\ref{sec:dataset_statistics} provides a more detailed description of the dataset. We highlight below the motivation and contributions of the three key elements that set Visual Ge\-nome apart from existing datasets.

The Visual Genome dataset regards relationships and attributes as first-class citizens of the annotation space, in addition to the traditional focus on objects. Recognition of relationships and attributes is an important part of the complete understanding of the visual scene, and in many cases, these elements are key to the story of a scene (e.g., the difference between ``a dog chasing a man'' versus ``a man chasing a dog''). The Visual Genome dataset is among the first to provide a detailed labeling of object interactions and attributes, \textbf{grounding visual concepts to language}\footnotemark.

\footnotetext{The Lotus Hill Dataset~\cite{yao2007introduction} also provides a similar annotation of object relationships, see Sec~\ref{sec:datasets}.}

An image is often a rich scenery that cannot be fully described in one summarizing sentence. The scene in Figure~\ref{fig:pipeline} contains multiple ``stories'': ``a man taking a photo of elephants,'' ``a woman feeding an elephant,'' ``a river in the background of lush grounds,'' etc. Existing datasets such as Flickr 30K~\cite{young2014image} and MS-COCO~\cite{lin2014microsoft} focus on high-level descriptions of an image\footnotemark. Instead, for each image in the Visual Genome dataset, we collect more than 42 descriptions for different regions in the image, providing a much denser and \textbf{complete set of descriptions of the scene}. In addition, inspired by VQA~\cite{antol2015vqa}, we also collect an average of $17$ question-answer pairs based on the descriptions for each image. Region-based question answers can be used to jointly develop NLP and vision models that can answer questions from either the description or the image, or both of them. 

\footnotetext{COCO has multiple sentences generated independently by different users, all focusing on providing an overall, one sentence description of the scene.}

With a set of dense descriptions of an image and the explicit correspondences between visual pixels (i.e.\ bounding boxes of objects) and textual descriptors (i.e.\ relationships, attributes), the Visual Ge\-nome dataset is poised to be the first image dataset that is capable of providing a structured \textbf{formalized representation} of an image, in the form that is widely used in knowledge base representations in NLP~\cite{zhou122007tree, guodong2005exploring, culotta2004dependency, socher2012semantic}. For example, in Figure~\ref{fig:pipeline}, we can formally express the relationship \predicate{holding} between the \object{woman} and \object{food} as \relationship{woman}{holding}{food)}. Putting together all the objects and relations in a scene, we can represent each image as a scene graph~\cite{Johnson2015CVPR}. The scene graph representation has been shown to improve semantic image retrieval~\cite{Johnson2015CVPR, schustergenerating}  and image captioning~\cite{farhadi2009describing, chang2014semantic, gupta2008beyond}. Furthermore, all objects, attributes and relationships in each image in the Visual Genome dataset are canonicalized to its corresponding WordNet~\cite{miller1995WordNet} ID (called a synset ID). This mapping connects all images in Visual Genome and provides an effective way to consistently query the same concept (object, attribute, or relationship) in the dataset. It can also potentially help train models that can learn from contextual information from multiple images.


In this paper, we introduce the Visual Genome dataset with the aim of training and benchmarking the next generation of computer models for comprehensive scene understanding. The paper proceeds as follows: In Section~\ref{sec:data_representation}, we provide a detailed description of each component of the dataset. Section~\ref{sec:related_works} provides a literature review of related datasets as well as related recognition tasks. Section~\ref{sec:crowdsourcing_pipeline} discusses the crowdsourcing strategies we deployed in the ongoing effort of collecting this dataset. Section~\ref{sec:dataset_statistics} is a collection of data analysis statistics, showcasing the key properties of the Visual Genome dataset. Last but not least, Section~\ref{sec:experiments} provides a set of experimental results that use Visual Genome as a benchmark.

Further visualizations, API, and additional information on the Visual Genome dataset can be found online\footnote{\url{https://visualgenome.org}}.

\begin{figure*}[t]%
    \centering
    
     \iftoggle{smallfigs}{
        \includegraphics[width=0.82\textwidth]{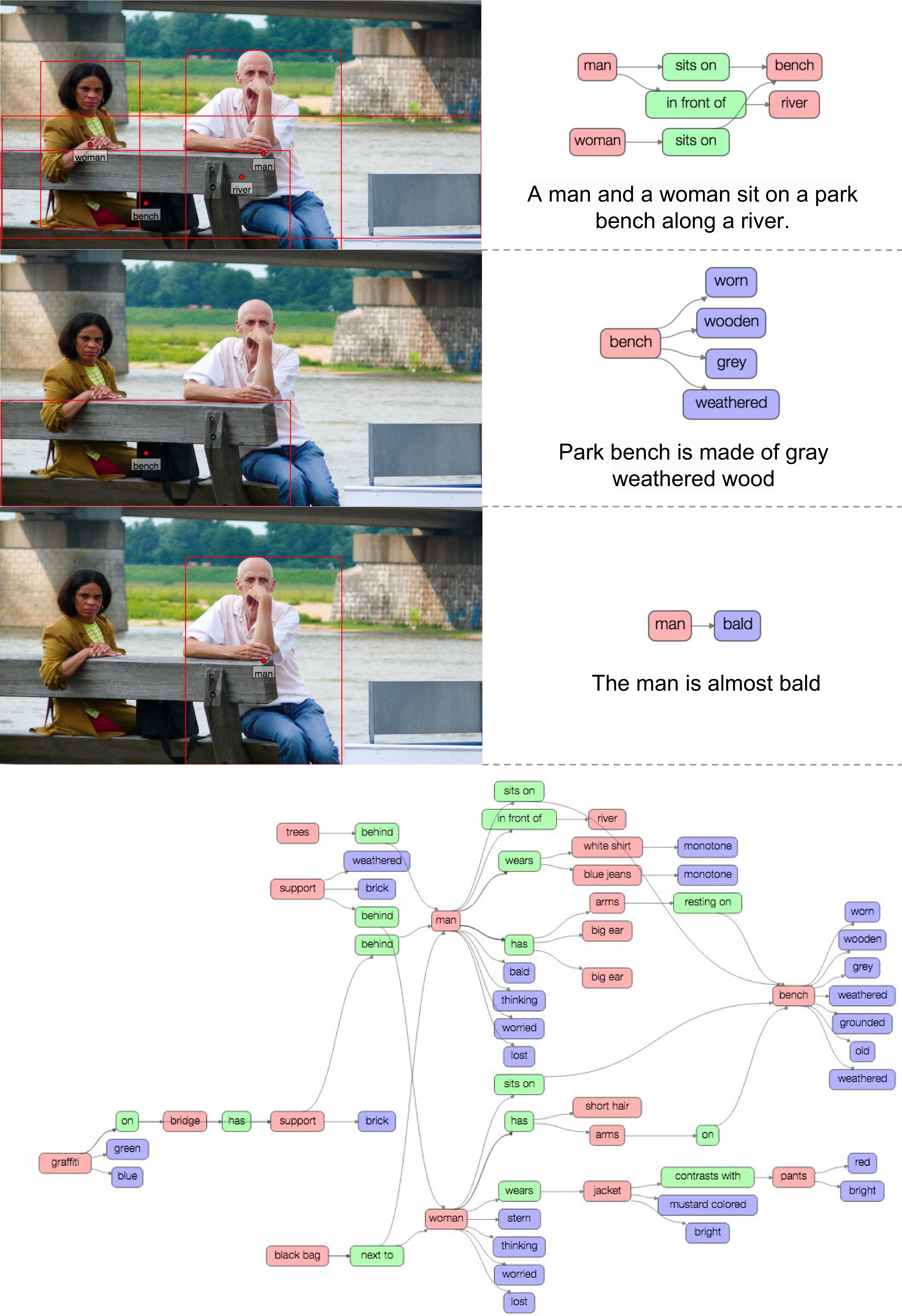}
     }{
        \includegraphics[width=0.82\textwidth]{png_graphics/scene_graph_1.png}
     }

    \caption{An example image from the Visual Genome dataset. We show 3 region descriptions and their corresponding region graphs. We also show the connected scene graph collected by combining all of the image's region graphs. The top region description is ``a man and a woman sit on a park bench along a river.'' It contains the objects: \object{man}, \object{woman}, \object{bench} and \object{river}. The relationships that connect these objects are: \relationship{man}{sits\_on}{bench}, \relationship{man}{in\_front\_of}{river}, and \relationship{woman}{sits\_on}{bench}.}%
    \label{fig:scene_graph_1}%
\end{figure*}

\FloatBarrier

\begin{figure*}[t]%
    \centering
    \iftoggle{smallfigs}{
        \includegraphics[width=0.8\textwidth]{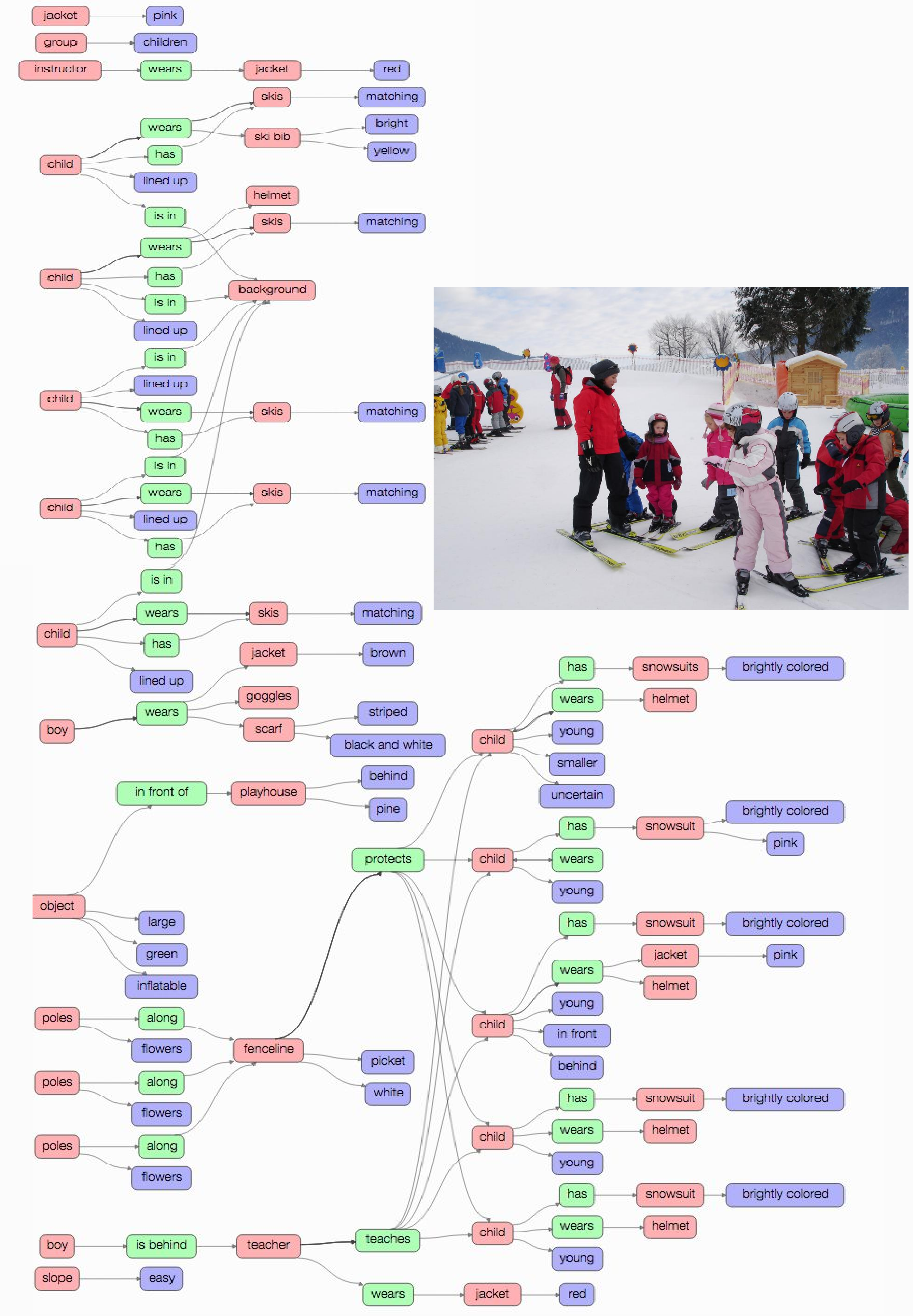}
    }{
        \includegraphics[width=0.8\textwidth]{png_graphics/scene_graph_2.png}
    }
    \caption{An example image from our dataset along with its scene graph representation. The scene graph contains objects (\object{child}, \object{instructor}, \object{helmet}, etc.) that are localized in the image as bounding boxes (not shown). These objects also have attributes: \attribute{large}, \attribute{green}, \attribute{behind}, etc. Finally, objects are connected to each other through relationships: \relationship{child}{wears}{helmet}, \relationship{instructor}{wears}{jacket}, etc.}
    \label{fig:scene_graph_2}%
\end{figure*}

\begin{figure*}[ht]
 \centering
    \iftoggle{smallfigs}{
        \includegraphics[width=\textwidth]{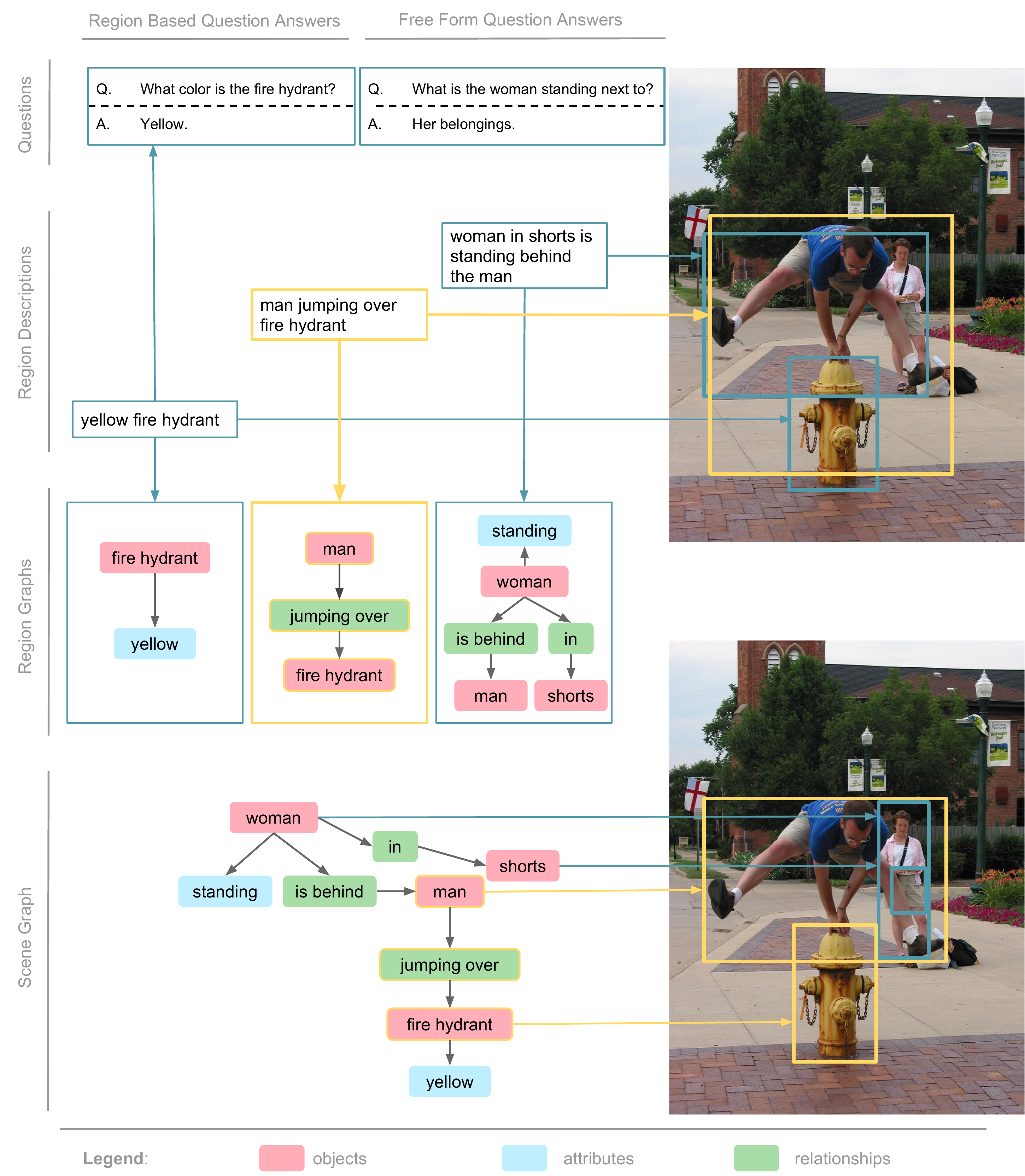}
    }{
        \includegraphics[width=\textwidth]{png_graphics/data_representation.png}
    }
    \caption{A representation of the Visual Genome dataset. Each image contains region descriptions that describe a localized portion of the image. We collect two types of question answer pairs (QAs): freeform QAs and region-based QAs. Each region is converted to a region graph representation of objects, attributes, and pairwise relationships. Finally, each of these region graphs are combined to form a scene graph with all the objects grounded to the image. \textit{Best viewed in color}}
\label{fig:data_representation}
\end{figure*}

\FloatBarrier


\section{Visual Genome Data Representation}
\label{sec:data_representation}

The Visual Genome dataset consists of seven main components: \textit{region descriptions}, \textit{objects}, \textit{attributes}, \textit{relationships}, \textit{region graphs}, \textit{scene graphs}, and \textit{question-answer pairs}. Figure~\ref{fig:data_representation} shows examples of each component for one image. To enable research on comprehensive understanding of images, we begin by collecting descriptions and question answers. These are raw texts without any restrictions on length or vocabulary. Next, we extract objects, attributes and relationships from our descriptions. Together, objects, attributes and relationships fabricate our scene graphs that represent a formal representation of an image. In this section, we break down Figure~\ref{fig:data_representation} and explain each of the seven components. In Section~\ref{sec:crowdsourcing_pipeline}, we will describe in more detail how data from each component is collected through a crowdsourcing platform.

\subsection{Multiple regions and their descriptions}
\begin{figure}[t]%
    \centering
    \includegraphics[width=0.46\textwidth]{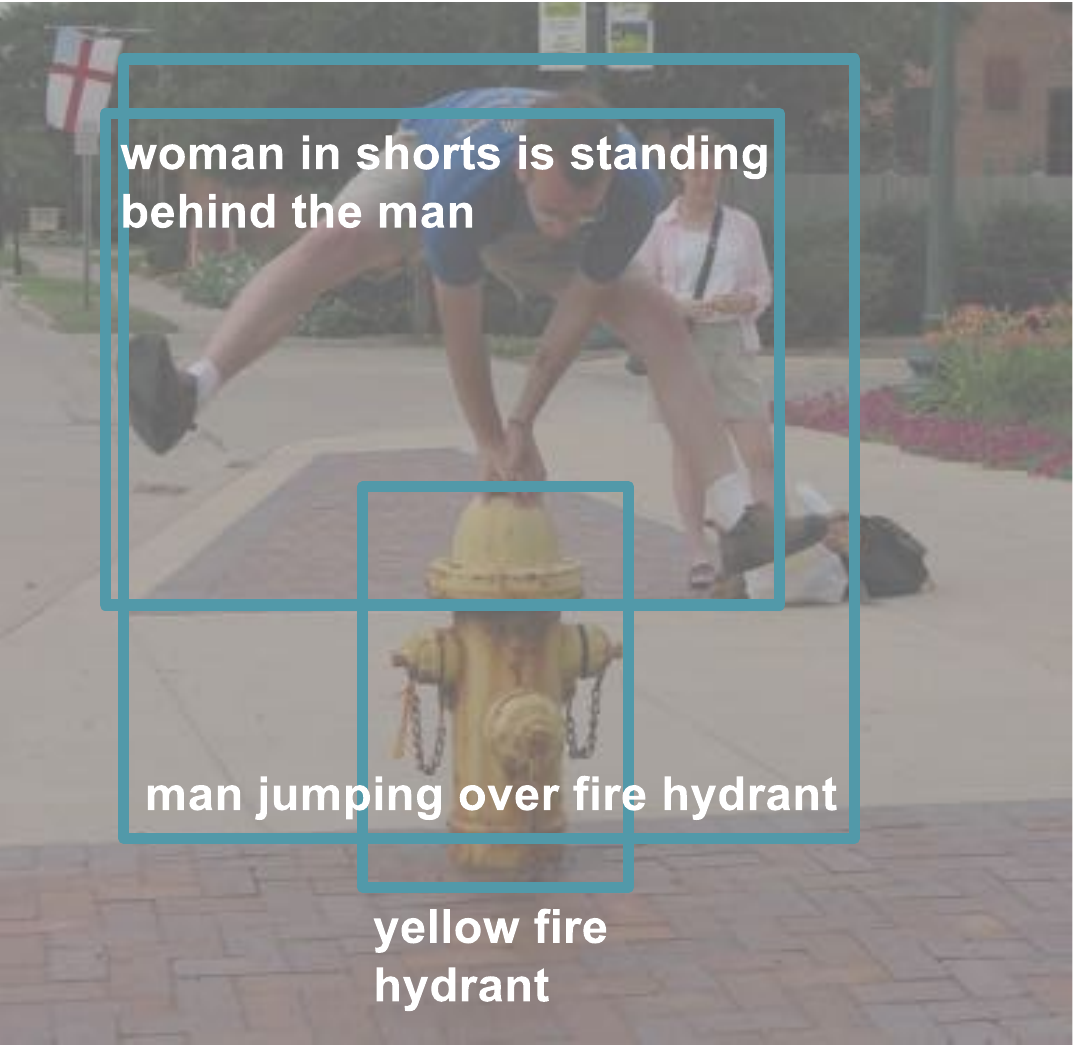}
    \caption{To describe all the contents of and interactions in an image, the Visual Genome dataset includes multiple human-generated image regions descriptions, with each region localized by a bounding box. Here, we show three regions descriptions: ``man jumping over a fire hydrant,'' ``yellow fire hydrant,'' and ``woman in shorts is standing beghind the man.''}
    \label{fig:data_representation_regions}
\end{figure}

In a real-world image, one simple summary sentence is often insufficient to describe all the contents of and interactions in an image. Instead, one natural way to extend this might be a collection of descriptions based on different regions of a scene. In Visual Genome, we collect human-generated image region descriptions, with each region localized by a bounding box. In Figure~\ref{fig:data_representation_regions}, we show three examples of region descriptions. Regions are allowed to have a high degree of overlap with each other when the descriptions differ. For example, ``yellow fire hydrant'' and ``woman in shorts is standing behind the man'' have very little overlap, while ``man jumping over fire hydrant'' has a very high overlap with the other two regions. Our dataset contains on average a total of $42$ region descriptions per image. Each description is a phrase ranging from $1$ to $16$ words in length describing that region. 

\begin{figure}[ht]%
    \centering
    \includegraphics[width=0.45\textwidth]{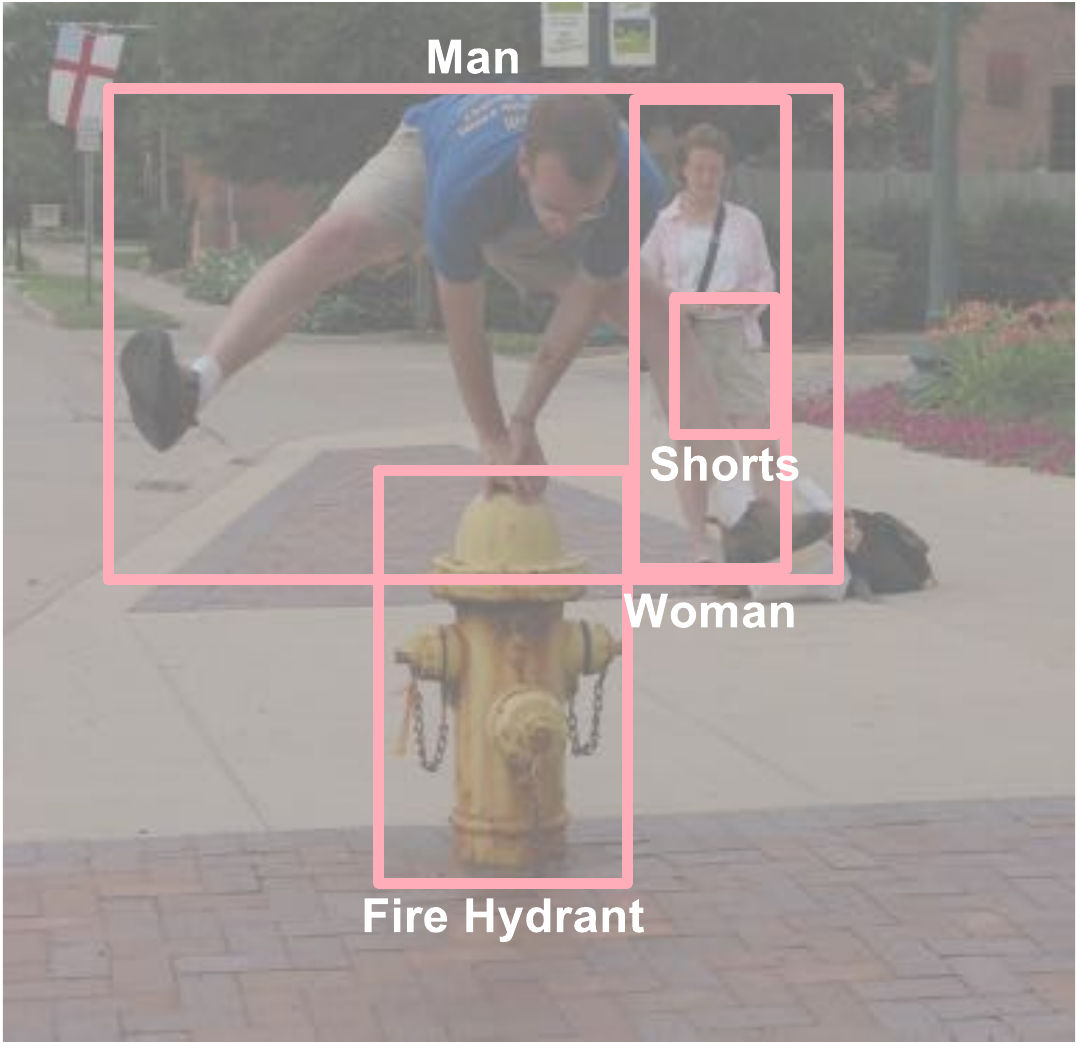}
    \caption{From all of the region descriptions, we extract all objects mentioned. For example, from the region description ``man jumping over a fire hydrant,'' we extract \object{man} and \object{fire hydrant}.}%
    \label{fig:data_representation_objects}%
\end{figure}

\subsection{Multiple objects and their bounding boxes} 
Each image in our dataset consists of an avarege of $21$ objects, each delineated by a tight bounding box (Figure~\ref{fig:data_representation_objects}). Furthermore, each object is canonicalized to a synset ID in WordNet~\cite{miller1995WordNet}. For example, \object{man} and \object{person} would get mapped to \synset{man.n.03 (the generic use of the word to refer to any human being)}. Similarly, \object{person} gets mapped to \synset{person.n.01 (a human being)}. Afterwards, these two concepts can be joined to \synset{person.n.01} since this is a hypernym of \synset{man.n.03}. This is an important standardization step to avoid multiple names for one object (e.g.\ man, person, human), and to connect information across images.

\begin{figure}[t]%
    \centering
    \includegraphics[width=0.4\textwidth]{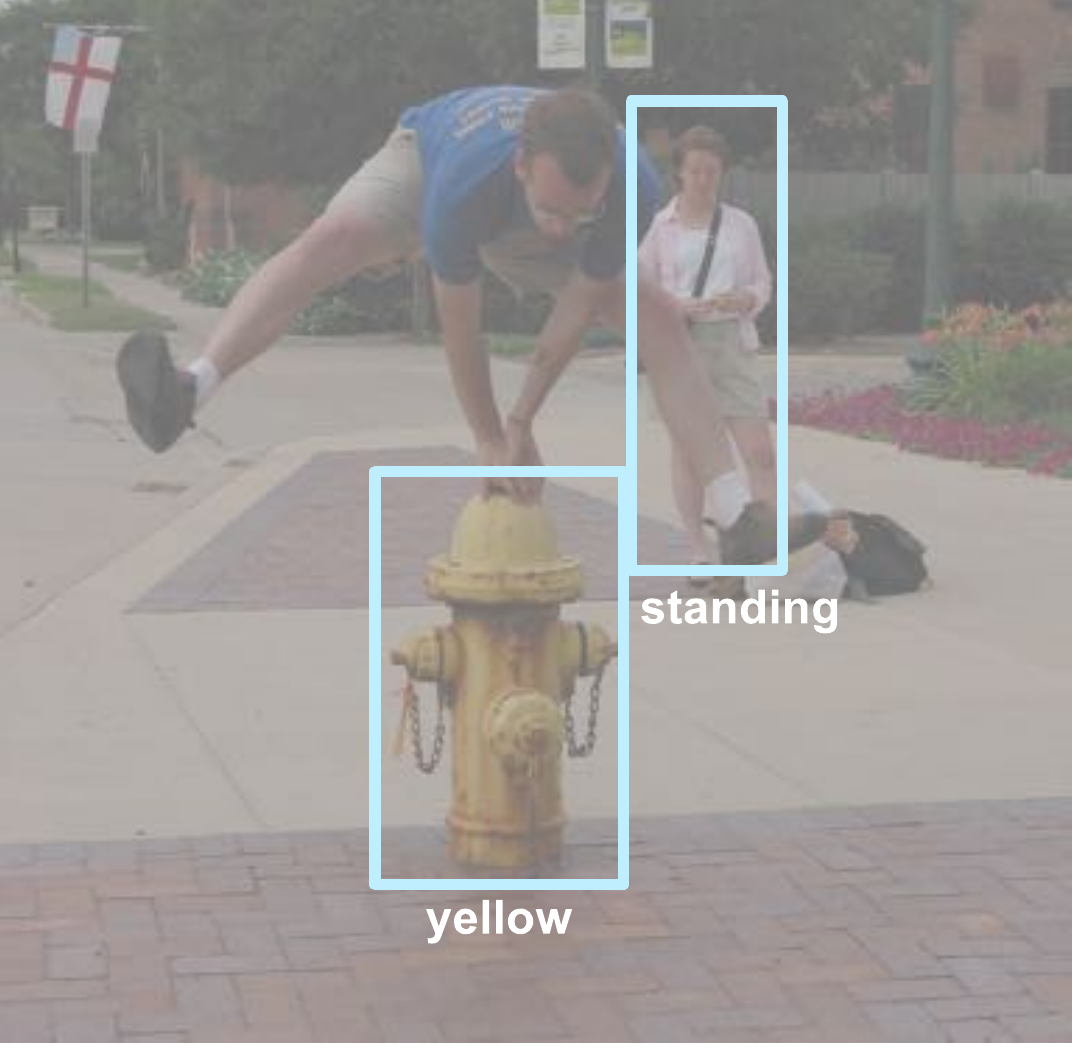}
    \caption{Some descriptions also provide attributes for objects. For example, the region description ``yellow fire hydrant'' adds that the \object{fire hydrant} is \attribute{yellow}. Here we show two attributes: \attribute{yellow} and \attribute{standing}.}
    \label{fig:data_representation_attributes}
\end{figure}

\subsection{A set of attributes} 
Each image in Visual Genome has an average of 16 attributes. Objects can have zero or more attributes associated with them. Attributes can be color (\attribute{yellow}), states (\attribute{standing}), etc. (Figure~\ref{fig:data_representation_attributes}). Just like we extract objects from region descriptions, we also extract the attributes attached to these objects. In Figure~\ref{fig:data_representation_attributes}, from the phrase ``yellow fire hydrant,'' we extract the attribute \attribute{yellow} for the \object{fire hydrant}. As with objects, we canonicalize all attributes to WordNet~\cite{miller1995WordNet}; for example, \attribute{yellow} is mapped to \synset{yellow.s.01 (of the color intermediate between green and orange in the color spectrum; of something resembling the color of an egg yolk)}. 

\subsection{A set of relationships} 
\begin{figure}[t]%
    \centering
    \includegraphics[width=0.42\textwidth]{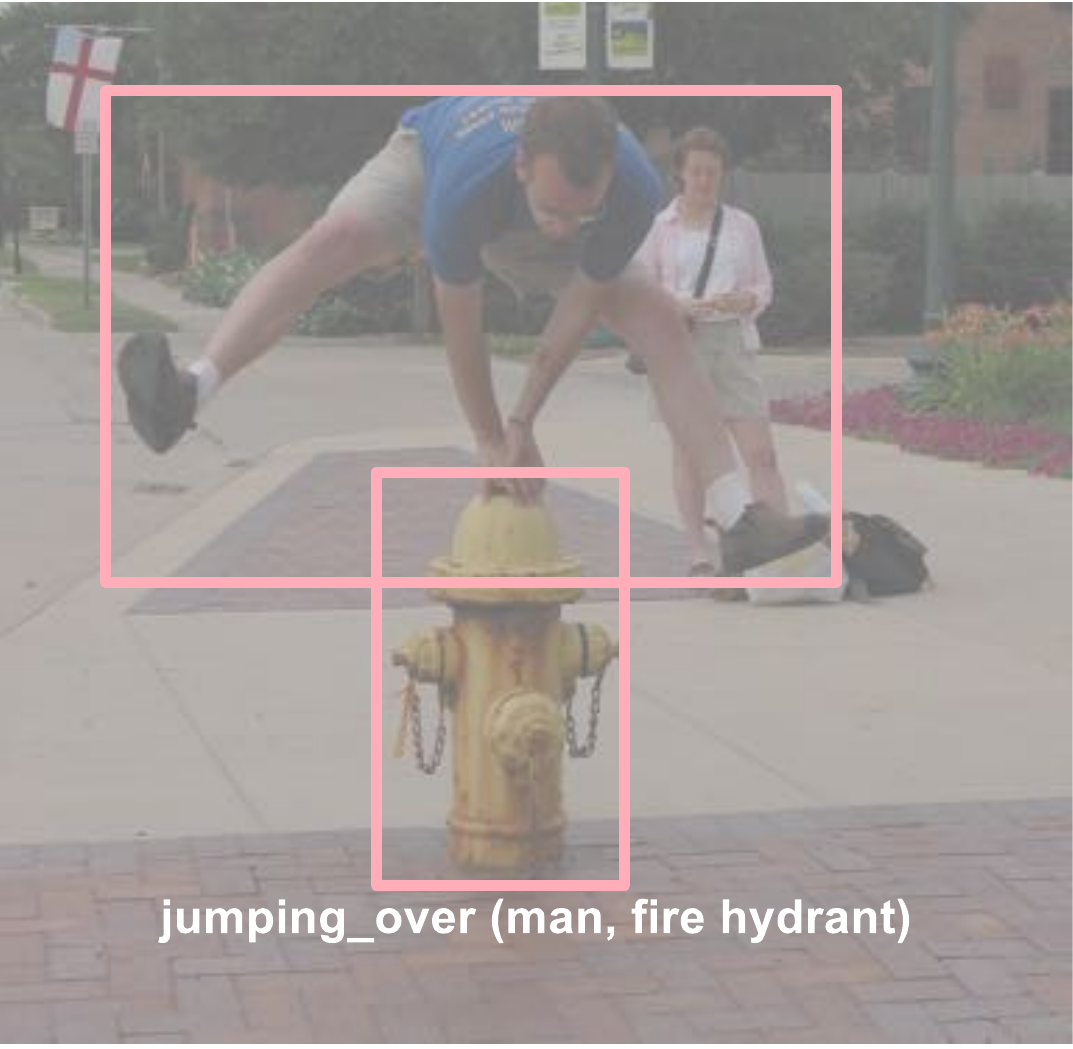}    \caption{Our dataset also captures the relationships and interactions between objects in our images. In this example, we show the relationship \predicate{jumping over} between the objects \object{man} and \object{fire hydrant}.}%
    \label{fig:data_representation_relationships}%
\end{figure}

Relationships connect two objects together. These relationships can be actions (\predicate{jumping over}), spatial (\predicate{is behind}), verbs (\predicate{wear}), prepositions (\predicate{with}), comparative (\predicate{taller than}), or prepositional phrases (\predicate{drive on}). For example, from the region description ``man jumping over fire hydrant,'' we extract the relationship \predicate{jumping over} between the objects \object{man} and \object{fire hydrant} (Figure~\ref{fig:data_representation_relationships}). These relationships are directed from one object, called the subject, to another, called the object. In this case, the subject is the \object{man}, who is performing the relationship \predicate{jumping over} on the object \object{fire hydrant}. Each relationship is canonicalized to a WordNet~\cite{miller1995WordNet} synset ID; i.e.\ \predicate{jumping} is canonicalized to \synset{jump.a.1 (move forward by leaps and bounds)}. On average, each image in our dataset contains 18 relationships.

\subsection{A set of region graphs} 
Combining the objects, attributes, and relationships extracted from region descriptions, we create a directed graph representation for each of the $42$ regions. Examples of region graphs are shown in Figure~\ref{fig:data_representation}. Each region graph is a structured representation of a part of the image. The nodes in the graph represent objects, attributes, and relationships. Objects are linked to their respective attributes while relationships link one object to another. The links connecting two objects in Figure~\ref{fig:data_representation} point from the subject to the relationship and from the relationship to the other object.

\subsection{One scene graph} 
While region graphs are localized representations of an image, we also combine them into a single scene graph representing the entire image (Figure~\ref{fig:scene_graph_2}). 
The scene graph is the \emph{union} of all region graphs and contains all objects, attributes, and relationships from each region description. By doing so, we are able to combine multiple levels of scene information in a more coherent way. For example in Figure~\ref{fig:data_representation}, the leftmost region description tells us that the ``fire hydrant is yellow,'' while the middle region description tells us that the ``man is jumping over the fire hydrant.'' Together, the two descriptions tell us that the ``man is jumping over a yellow fire hydrant.''

\subsection{A set of question answer pairs} 
We have two types of QA pairs associated with each image in our dataset: \textit{freeform QAs}, based on the entire image, and \textit{region-based QAs}, based on selected regions of the image. We collect 6 different types of questions per image: \qa{what}, \qa{where}, \qa{how}, \qa{when}, \qa{who}, and \qa{why}. In Figure~\ref{fig:data_representation}, ``Q. What is the woman standing next to?; A. Her belongings'' is a freeform QA\@. Each image has at least one question of each type listed above. Region-based QAs are collected by prompting workers with region descriptions. For example, we use the region ``yellow fire hydrant'' to collect the region-based QA: ``Q. What color is the fire hydrant?; A. Yellow.'' Region based QAs allow us to independently study methods that use NLP and vision priors to answer questions.


\section{Related Work}
\label{sec:related_works}

We discuss existing datasets that have been released and used by the vision community for classification and object detection. We also mention work that has improved object and attribute detection models. Then, we explore existing work that has utilized representations similar to our relationships between objects. In addition, we dive into literature related to cognitive tasks like image description, question answering, and knowledge representation.

\subsection{Datasets} 
\label{sec:datasets}
Datasets (Table~\ref{tab:all_datasets}) have been growing in size as researchers have begun tackling increasingly complicated problems. \textit{Caltech 101}~\cite{fei2007learning}  was one of the first datasets hand-curated for image classification, with 101 object categories and $15$-$30$ of examples per category. One of the biggest criticisms of Caltech 101 was the lack of variability in its examples. \textit{Caltech 256}~\cite{griffin2007caltech} increased the number of categories to 256, while also addressing some of the shortcomings of Caltech 101. However, it still had only a handful of examples per category, and most of its images contained only a single object. \textit{LabelMe}~\cite{russell2008labelme} introduced a dataset with multiple objects per category. They also provided a web interface that experts and novices could use to annotate additional images. This web interface enabled images to be labeled with polygons, helping create datasets for image segmentation. The \textit{Lotus Hill dataset}~\cite{yao2007introduction} contains a hierarchical decomposition of objects (vehicles, man-made objects, animals, etc.) along with segmentations. Only a small part of this dataset is freely available. \textit{SUN}~\cite{xiao2010sun}, just like LabelMe~\cite{russell2008labelme} and Lotus Hill~\cite{yao2007introduction}, was curated for object detection. 
Pushing the size of datasets even further, \textit{$80$ Million Tiny Images}~\cite{torralba200880} created a significantly larger dataset than its predecessors. It contains tiny (i.e.\ $32\times32$ pixels) images that were collected using WordNet~\cite{miller1995WordNet} synsets as queries. However, because the data in $80$ Million Images were not human-verified, they contain numerous errors. \textit{YFCC100M}~\cite{thomee2015yfcc100m} is another large database of $100$ million images that is still largely unexplored. It contains human generated and machine generated tags. 

\textit{Pascal VOC}~\cite{everingham2010pascal} pushed research from classification to object detection with a dataset containing $20$ semantic categories in $11,000$ images. \textit{Imagenet}~\cite{deng2009imagenet} took WordNet synsets and crowdsourced a large dataset of 14 million images. They started the ILSVRC~\cite{ILSVRC15} challenge for a variety of computer vision tasks. ILSVRC and PASCAL provide a test bench for object detection, image classification, object segmentation, person layout, and action classification. \textit{MS-COCO}~\cite{lin2014microsoft} recently released its dataset, with over $328,000$ images with sentence descriptions and segmentations of $91$ object categories. The current largest dataset for QA, \textit{VQA}~\cite{antol2015vqa}, contains $204,721$ images annotated with one or more question answers. They collected a dataset of $614,163$ freeform questions with $6.1$M ground truth answers and provided a baseline approach in answering questions using an image and a textual question as the input.

\begin{sidewaystable*}
\vspace{45pc}
\centering
\mbox{%
\setlength{\extrarowheight}{5pt}
\scriptsize  
\begin{tabular}{lrrrrrrrrrr}
 &        & Descriptions & Total   & \# Object  & Objects    & \# Attributes  & Attributes  & \# Relationship  & Relationships & Question \\ [0ex]
 & Images & per Image    & Objects & Categories & per Image  & Categories     & per Image   & Categories       & per Image     & Answers \\
\toprule
YFCC100M~\cite{thomee2015yfcc100m}              & 100,000,000   & - & -             & -         & -         & -     & -     & -     & - & - \\ 
Tiny Images~\cite{torralba200880}               & 80,000,000    & - & -             & 53,464    & 1         & -     & -     & -     & - & - \\ 
ImageNet~\cite{deng2009imagenet}                & 14,197,122    & - & 14,197,122    & 21,841    & 1         & -     & -     & -     & - & - \\
ILSVRC Detection (2012)~\cite{ILSVRC15}              & 476,688       & - & 534,309       & 200       & 2.5       & -     & -     & -     & - & - \\
MS-COCO~\cite{2015arXiv150602203R}       & 328,000       & 5 & 27,472        & 91        & -         & -     & -     & -     & - & - \\
Flickr 30K~\cite{young2014image}                & 30,000        & 5 & -             & -         & -         & -     & -     & -     & - & - \\
Caltech 101~\cite{fei2007learning}              & 9,144         & - & 9,144         & 102       & 1         & -     & -     & -     & - & - \\
Caltech 256~\cite{griffin2007caltech}           & 30,608        & - & 30,608        & 257       & 1         & -     & -     & -     & - & - \\
Caltech Pedestrian~\cite{dollar2012pedestrian}  & 250,000       & - & 350,000       & 1         & 1.4       & -     & -     & -     & - & - \\
Pascal Detection~\cite{everingham2010pascal}    & 11,530        & - & 27,450        & 20        & 2.38      & -     & -     & -     & - & -\\
Abstract Scenes~\cite{zitnick2013bringing}      & 10,020        & - & 58            & 11        & 5         & -     & -     & -     & - & - \\
aPascal~\cite{farhadi2009describing}            & 12,000        & - & -             & -         & -         & 64    & -     & -     & - & - \\
Animal Attributes~\cite{lampert2009learning}    & 30,000        & - & -             & -         & -         & 1,280 & -     & -     & - & -\\
SUN Attributes~\cite{patterson2014sun}          & 14,000        & - & -             & -         & -         & 700   & 700  & -     & - & - \\
Caltech Birds~\cite{wah2011caltech}             & 11,788        & - & -             & -         & -         & 312   & 312  & -     & - & - \\
COCO Actions~\cite{2015arXiv150602203R}         & 10,000        & - & -             & -         & -         & -     & -     & 140   & - & - \\
Visual Phrases~\cite{sadeghi2011recognition}    & -             & - & -             & -         & -         & -     & -     & 17    & 1 & - \\
Viske~\cite{sadeghi2015viske}                   & -             & - & -             & -         & -         & -     & -     & 6500  & - & - \\
DAQUAR~\cite{malinowski2014multi}               & 1,449         & - & -             & -         & -         & -     & -     & -     & - & 12,468 \\
COCO QA~\cite{ren2015image}                     & 123,287       & - & -             & -         & -         & -     & -     & -     & - & 117,684 \\
Baidu~\cite{gao2015you}                         & 120,360       & - & -             & -         & -         & -     & -     & -     & - & 250,569 \\
VQA~\cite{antol2015vqa}                         & 204,721       & - & -             & -         & -         & -     & -     & -     & - & 614,163 \\
\midrule
\textbf{Visual Genome}                          & 108,000       & 50 & 4,102,818    & 76,340    & 16        & 15,626 & 16   & 47    & 18 & 1,773,258 \\
\bottomrule
\end{tabular}
}
\caption{A comparison of existing datasets with Visual Genome. We show that Visual Genome has an order of magnitude more descriptions and question answers. It also has a more diverse set of object, attribute, and relationship classes. Additionally, Visual Genome contains a higher density of these annotations per image.}
\label{tab:all_datasets}
\end{sidewaystable*}

\textit{Visual Genome} aims to bridge the gap between all these datasets, collecting not just annotations for a large number of objects but also scene graphs, region descriptions, and question answer pairs for image regions. Unlike previous datasets, which were collected for a single task like image classification, the Visual Genome dataset was collected to be a general-purpose representation of the visual world, without bias toward a particular task. Our images contain an average of $21$ objects, which is almost an order of magnitude more dense than any existing vision dataset. Similarly, we contain an average of $18$ attributes and $18$ relationships per image. We also have an order of magnitude more unique objects, attributes, and relationships than any other dataset. Finally, we have 1.7 million question answer pairs, also larger than any other dataset for visual question answering.

\subsection{Image Descriptions} 
One of the core contributions of Visual Genome is its descriptions for multiple regions in an image. As such, we mention other image description datasets and models in this subsection. Most work related to describing images can be divided into two categories: retrieval of human-generated captions and generation of novel captions. Methods in the first category use similarity metrics between image features from predefined models to retrieve similar sentences~\cite{ordonez2011im2text, hodosh2013framing}. Other methods map both sentences and their images to a common vector space~\cite{ordonez2011im2text} or map them to a space of triples~\cite{farhadi2010every}. Among those in the second category, a common theme has been to use recurrent neural networks to produce novel captions~\cite{kiros2014multimodal, mao2014explain, karpathy2014deep, vinyals2014show}. More recently, researchers have also used a visual attention model~\cite{xu2015show}. 

One drawback of these approaches is their attention to describing only the most salient aspect of the image. This problem is amplified by datasets like Flickr 30K~\cite{young2014image} and MS-COCO~\cite{lin2014microsoft}, whose sentence desriptions tend to focus, somewhat redundantly, on these salient parts. For example, ``an elephant is seen wandering around on a sunny day,'' ``a large elephant in a tall grass field,'' and ``a very large elephant standing alone in some brush'' are 3 descriptions from the MS-COCO dataset, and all of them focus on the salient elephant in the image and ignore the other regions in the image. Many real-world scenes are complex, with multiple objects and interactions that are best described using multiple descriptions~\cite{karpathy2014deep, lebret2015phrase}. Our dataset pushes toward a complete understanding of an image by collecting a dataset in which we capture not just scene-level descriptions but also myriad of low-level descriptions, the ``grammar'' of the scene.

\subsection{Objects} 
Object detection is a fundamental task in computer vision, with applications ranging from identification of faces in photo software to identification of other cars by self-driving cars on the road. It involves classifying an object into a semantic category and localizing the object in the image. Visual Genome uses objects as a core component on which each visual scene is built. Early datasets include the face detectio~ \cite{huang2008labeled} and pedestrian datasets~\cite{dollar2012pedestrian}. The PASCAL VOC and ILSVRC's detection dataset~\cite{deng2009imagenet} pushed research in object detection. But the images in these datasets are iconic and do not capture the settings in which these objects usually co-occur. To remedy this problem, MS-COCO~\cite{lin2014microsoft} annotated real-world scenes that capture object contexts. However, MS-COCO was unable to describe all the objects in its images, since they annotated only 91 object categories. In the real world, there are many more objects that the ones captured by existing datasets. Visual Genome aims at collecting annotations for all visual elements that occur in images, increasing the number of semantic categories to over 17,000.

\vspace{-0.3cm}
\subsection{Attributes} 
The inclusion of attributes allows us to describe, compare, and more easily categorize objects. Even if we haven't seen an object before, attributes allow us to infer something about it; for example, ``yellow and brown spotted with long neck'' likely refers to a giraffe. Initial work in this area involved finding objects with similar features~\cite{malisiewicz2008recognition} using examplar SVMs. Next, textures were used to study objects~\cite{varma2005statistical}, while other methods learned to predict colors~\cite{ferrari2007learning}. Finally, the study of attributes was explicitly demonstrated to lead to improvements in object classification~\cite{farhadi2009describing}. Attributes were defined to be paths (``has legs''), shapes (``spherical''), or materials (``furry'') and could be used to classify new categories of objects. Attributes have also played a large role in improving fine-grained recognition~\cite{goering2014nonparametric} on fine-grained attribute datasets like CUB-2011~\cite{wah2011caltech}. In Visual Genome, we use a generalized formulation~\cite{Johnson2015CVPR}, but we extend it such that attributes are not image-specific binaries but rather object-specific for each object in a real-world scene. We also extend the types of attributes to include size (``small''), pose (``bent''), state (``transparent''), emotion (``happy''), and many more.

\subsection{Relationships} 
Relationship extraction has been a traditional problem in information extraction and in natural language processing. Syntactic features~\cite{zhou122007tree, guodong2005exploring}, dependency tree methods~\cite{culotta2004dependency, bunescu2005shortest}, and deep neural networks~\cite{socher2012semantic, zeng2014relation} have been employed to extract relationships between two entities in a sentence. However, in computer vision, very little work has gone into learning or predicting relationships. Instead, relationships have been implicitly used to improve other vision tasks. Relative layouts between objects have improved scene categorization~\cite{izadinia2014incorporating}, and 3D spatial geometry between objects has helped object detection~\cite{choi2013understanding}. Comparative adjectives and prepositions between pairs of objects have been used to model visual relationships and improved object localization~\cite{gupta2008beyond}. 

Relationships have already shown their utility in improving cognitive tasks. A meaning space of relationships has improved the mapping of images to sentences~\cite{farhadi2010every}. Relationships in a structured representation with objects have been defined as a graph structure called a \textit{scene graph}, where the nodes are objects with attributes and edges are relationships between objects. This representation can be used to generate indoor images from sentences and also to improve image search~\cite{chang2014semantic, Johnson2015CVPR}. We use a similar scene graph representation of an image that generalizes across all these previous works~\cite{Johnson2015CVPR}. Recently, relationships have come into focus again in the form of question answering about associations between objects~\cite{sadeghi2015viske}. These questions ask if a relationship, involving generally two objects, is true, e.g. ``do dogs eat ice cream?''. We believe that relationships will be necessary for higher-level cognitive tasks~\cite{Johnson2015CVPR, luvisualrelationship}, so we collect the largest corpus of them in an attempt to improve tasks by actually understanding relationships between objects.

\subsection{Question Answering} 
Visual question answering (QA) has been recently proposed as a proxy task of evaluating a computer vision system's ability to understand an image beyond object recognition~\cite{geman2015visual,malinowski2014multi}. Several visual QA benchmarks have been proposed in the last few months. The DAQUAR~\cite{malinowski2014multi} dataset was the first toy-sized QA benchmark built upon indoor scene RGB-D images of NYU Depth v2~\cite{SilbermanECCV12}. Most new datasets~\cite{VisualMadlibs,ren2015image,antol2015vqa,gao2015you} have collected QA pairs on MS-COCO images, either generated automatically by NLP tools~\cite{ren2015image} or written by human workers~\cite{VisualMadlibs,antol2015vqa,gao2015you}.

In previous datasets, most questions concentrated on simple recognition-based questions about the salient objects, and answers were often extremely short. For instance, $90\%$ of DAQUAR answers~\cite{malinowski2014multi} and $87\%$ of VQA answers~\cite{antol2015vqa} consist of single-word object names, attributes, and quantities. This shortness limits their diversity and fails to capture the long-tail details of the images. Given the availability of new datasets, an array of visual QA models have been proposed to tackle QA tasks. The proposed models range from SVM classifiers~\cite{antol2015vqa} and probabilistic inference~\cite{malinowski2014multi} to recurrent neural networks~\cite{gao2015you,malinowski2015ask,ren2015image} and convolutional networks~\cite{ma2015cnnQA}. Visual Genome aims to capture the details of the images with diverse question types and long answers. These questions should cover a wide range of visual tasks from basic perception to complex reasoning. Our QA dataset of $1.7$ million QAs is also larger than any currently existing dataset.

\subsection{Knowledge Representation}  
A knowledge representation of the visual world is capable of tackling an array of vision tasks, from action recognition to general question answering. However, it is difficult to answer ``what is the minimal viable set of knowledge needed to understand about the physical world?''~\cite{hayes1978naive}. It was later proposed that there be a certain plurality to concepts and their related axioms~\cite{hayes1985naive}. These efforts have grown to model physical processes~\cite{forbus1984qualitative} or to model a series of actions as scripts~\cite{schank2013scripts} for stories---both of which are not depicted in a single static image but which play roles in an image's story. More recently, NELL~\cite{betteridge2009toward} learns probabilistic horn clauses by extracting information from the web. DeepQA~\cite{ferrucci2010building} proposes a probabilistic question answering architecture involving over $100$ different techniques. Others have used Markov logic networks~\cite{zhu2009statsnowball,niu2012elementary} as their  representation to perform statistical inference for knowledge base construction.  Our work is most similar to that of those~\cite{chen2013neil,zhu2014eccv,zhu2015kb,sadeghi2015viske} who attempt to learn common-sense relationships from images. Visual Genome scene graphs can also be considered a \textit{dense} knowledge representation for images. It is similar to the format used in knowledge bases in NLP.


\section{Crowdsourcing Strategies}
\label{sec:crowdsourcing_pipeline}

Visual Genome was collected and verified entirely by crowd workers from Amazon Mechanical Turk. In this section, we outline the pipeline employed in creating all the components of the dataset. Each component (region descriptions, objects, attributes, relationships, region graphs, scene graphs, questions and answers) involved multiple task stages. We mention the different strategies used to make our data accurate and to enforce diversity in each component. We also provide background information about the workers who helped make Visual Genome possible.

\subsection{Crowd Workers}
We used Amazon Mechanical Turk (AMT) as our primary source of annotations. Overall, a total of over $33,000$ unique workers contributed to the dataset.  The dataset was collected over the course of $6$ months after $15$ months of experimentation and iteration on the data representation. Approximately $800,000$ Human Intelligence Tasks (HITs) were launched on AMT, where each HIT involved creating descriptions, questions and answers, or region graphs. Each HIT was designed such that workers manage to earn anywhere between \$$6$-\$$8$ per hour if they work continuously, in line with  ethical research standards on Mechanical Turk~\cite{salehi2015we}. Visual Genome HITs achieved a $94.1$\% retention rate, meaning that $94.1$\% of workers who completed one of our tasks went ahead to do more. Table~\ref{table:worker_nation} outlines the percentage distribution of the locations of the workers. $93.02$\% of workers contributed from the United States.

\begin{table}[h]
\centering
\begin{tabular}{l r}
\textbf{Country} & \textbf{Distribution} \\
\midrule
United States & 93.02\% \\
Philippines & 1.29\% \\
Kenya & 1.13\% \\
India & 0.94\% \\
Russia & 0.50\% \\
Canada & 0.47\% \\
(Others) & 2.65\% \\
\end{tabular}
\caption{Geographic distribution of countries from where crowd workers contributed to Visual Genome.}
\label{table:worker_nation}
\end{table}

\begin{figure*}[t]%
    \centering
    \subfloat[]{{\includegraphics[width=0.46\textwidth]{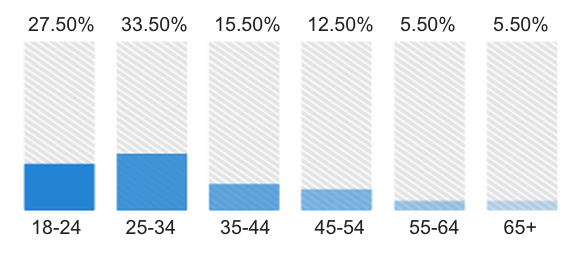} }}%
    \qquad
    \subfloat[]{{\includegraphics[width=0.46\textwidth]{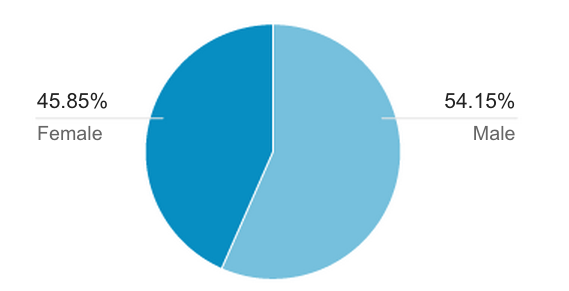} }}%
    \caption{(a) Age and (b) gender distribution of Visual Genome's crowd workers.}%
    \label{fig:worker_demographic}%
\end{figure*}

Figures~\ref{fig:worker_demographic} (a) and (b) outline the demographic distribution of our crowd workers. The majority of our workers were between the ages of $25$ and $34$ years old. Our youngest contributor was $18$ years old and the oldest was $68$ years old. We also had a near-balanced split of $54.15$\% male and $45.85$\% female workers.

\subsection{Region Descriptions}
\label{sec:crowdsourcing_region_description}
Visual Genome's main goal is to enable the study of cognitive computer vision tasks. The next step towards understanding images requires studying relationships between objects in scene graph representations of images. However, we observed that collecting scene graphs directly from an image leads to workers annotating easy, frequently-occurring relationships like \relationship{man}{wearing}{shirt} instead of focusing on salient parts of the image. This is evident from previous datasets~\cite{Johnson2015CVPR, luvisualrelationship} that contain a large number of such relationships. After experimentation, we observed that when asked to describe an image using natural language, crowd workers naturally start with the most salient part of the image and then move to describing other parts of the image one by one. Inspired by this finding, we focused our attention towards collecting a dataset of region descriptions that is diverse in content.

When a new image is added to the crowdsourcing pipeline with no annotations, it is sent to a worker who is asked to draw three bounding boxes and write three descriptions for the region enclosed by each box. Next, the image is sent to another worker along with the previously written descriptions. Workers are explicitly encouraged to write descriptions that have not been written before. This process is repeated until we have collect $50$ region descriptions for each image. To prevent workers from having to skim through a long list of previously written descriptions, we only show them the top seven most similar descriptions. We calculate these most similar descriptions using BLEU~\cite{papineni2002bleu} (n-gram) scores between pairs of sentences. We define the BLEU score between a description $d_i$ and a previous description $d_j$ to be:

\begin{equation}
BLEU_N(d_i, d_j) = b(d_i, d_j) \exp(\frac{1}{N}\sum_{n=1}^{N} \log p_n(d_i, d_j))
\end{equation}
\vspace{0.3cm}
where we enforce a brevity penalty using:
\vspace{0.3cm}
\begin{equation}
b(d_i, d_j) = \left\{ \begin{array}{lr} 1 & \mathrm{if} \; len(d_i) > len(d_j) \\ e^{1-\frac{len(d_j)}{len(d_i)}} & \mathrm{otherwise} \end{array} \right.
\end{equation}
\vspace{0.3cm}
and $p_n$ calculates the percentage of n-grams in $d_i$ that match n-grams in $d_j$.

When a worker writes a new description, we programmatically enforce that it has not been repeated by using BLEU score thresholds set to $0.7$ to ensure that it is dissimilar to descriptions from both of the following two lists:
\begin{enumerate}
\item \textbf{Image-specific descriptions.} A list of all previously written descriptions for that image.
\item \textbf{Global image descriptions.} A list of the top 100 most common written descriptions of all images in the dataset. This prevents very common phrases like ``sky is blue'' from dominating the set of region descriptions.
\end{enumerate}

\begin{figure}[t]%
    \centering
    \iftoggle{smallfigs}{
        \includegraphics[width=0.95\columnwidth]{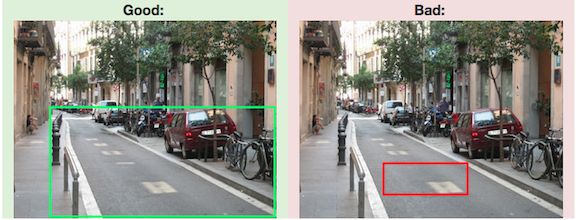}
    }{
        \includegraphics[width=\columnwidth]{png_graphics/bbox_coverage.png}
    }
    \caption{Good (left) and bad (right) bounding boxes for the phrase ``a street with a red car parked on the side,'' judged on \textbf{coverage}.}%
    \label{fig:bbox_coverage}%
\end{figure}

Finally, we ask workers to draw bounding boxes that satisfy one requirement: \textbf{coverage}. The bounding box must cover all objects mentioned in the description. Figure~\ref{fig:bbox_coverage} shows an example of a good box that covers both the \object{street} as well the \object{car} mentioned in the description, as well as an example of a bad box.

\subsection{Objects}
\begin{figure}[t]%
    \centering
    \iftoggle{smallfigs}{
        \includegraphics[width=\columnwidth]{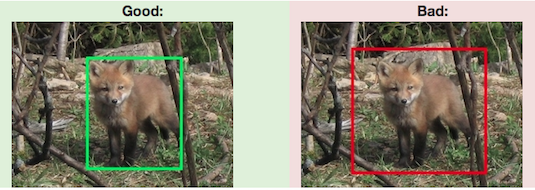}
    }{
        \includegraphics[width=\columnwidth]{png_graphics/bbox_quality.png}
    }
    \caption{Good (left) and bad (right) bounding boxes for the object \object{fox}, judged on both \textbf{coverage} as well as \textbf{quality}.}%
    \label{fig:bbox_quality}.%
\end{figure}

Once $50$ region descriptions are collected for an image, we extract the visual objects from each description. Each description is sent to one crowd worker, who extracts all the objects from the description and grounds each object as a bounding box in the image. For example, from Figure~\ref{fig:data_representation}, let's consider the description ``woman in shorts is standing behind the man.'' A worker would extract three objects: \object{woman}, \object{shorts}, and \object{man}. They would then draw a box around each of the objects. We require each bounding box to be drawn to satisfy two requirements: \textbf{coverage} and \textbf{quality}. Coverage has the same definition as described above in Section~\ref{sec:crowdsourcing_region_description}, where we ask workers to make sure that the bounding box covers the object completely (Figure~\ref{fig:bbox_quality}). Quality requires that each bounding box be as tight as possible around its object such that if the box's length or height were decreased by one pixel, it would no longer satisfy the coverage requirement. Since a one pixel error can be physically impossible for most workers, we relax the definition of quality to four pixels.

Multiple descriptions for an image might refer to the same object, sometimes with different words. For example, a \object{man} in one description might be referred to as \object{person} in another description. We can thus use this crowdsourcing stage to build these co-reference chains. With each region description given to a worker to process, we include a list of previously extracted objects as suggestions. This allows a worker to choose a previously drawn box annotated as \object{man} instead of redrawing a new box for \object{person}.

Finally, to increase the speed with which workers complete this task, we also use Stanford's dependency parser~\cite{manning-EtAl:2014:P14-5} to extract nouns automatically and send them to the workers as suggestions. While the parser manages to find most of the nouns, it sometimes misses compound nouns, so we avoided completely depending on this automated method. By combining the parser with crowdsourcing tasks, we were able to speed up our object extraction process without losing accuracy.

\subsection{Attributes, Relationships, and Region Graphs}
Once all objects have been extracted from each region description, we can extract the attributes and relationships described in the region. We present each worker with a region description along with its extracted objects and ask them to add attributes to objects or to connect pairs of objects with relationships, based on the text of the description. From the description ``woman in shorts is standing behind the man'', workers will extract the attribute \attribute{standing} for the \object{woman} and the relationships \relationship{woman}{in}{shorts} and \relationship{woman}{behind}{man}. Together, objects, attributes, and relationships form the region graph for a region description. Some descriptions like ``it is a sunny day'' do not contain any objects and therefore have no region graphs associated with them. Workers are asked to not generate any graphs for such descriptions. We create scene graphs by combining all the region graphs for an image by combining all the co-referenced objects from different region graphs.

\subsection{Scene Graphs}
\begin{figure}[t]%
    \centering
    \iftoggle{smallfigs}{
        \includegraphics[width=\columnwidth]{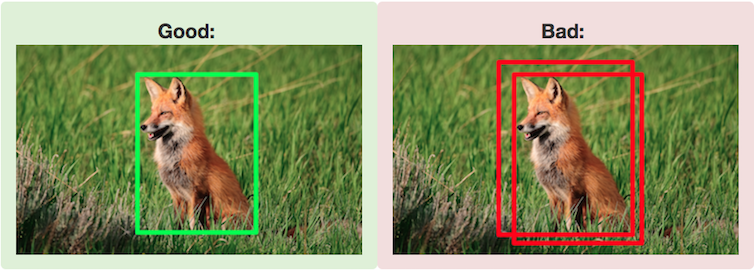}
    }{
        \includegraphics[width=\columnwidth]{png_graphics/bbox_combined.png}
    }
    \caption{Each object (\object{fox}) has only one bounding box referring to it (left). Multiple boxes drawn for the same object (right) are combined together if they have a minimum threshold of $0.9$ intersection over union.}%
    \label{fig:bbox_combined}.%
\end{figure}

The scene graph is the union of all region graphs extracted from region descriptions. We merge nodes from region graphs that correspond to the same object; for example, \object{man} and \object{person} in two different region graphs might refer to the same object in the image. We say that objects from different graphs refer to the same object if their bounding boxes have an overlap over union of $0.8$. However, this heuristic might contain false positives. So, before merging two objects, we ask workers to confirm that a pair of objects with significant overlap are indeed the same object. For example, in Figure~\ref{fig:bbox_combined} (right), the \object{fox} might be extracted from two different region descriptions. These boxes are then combined together (Figure~\ref{fig:bbox_combined} (left)) when constructing the scene graph. Two region graphs are combined together by merging objects that are co-referenced by both the graphs. 

\subsection{Questions and Answers}
To create question answer (QA) pairs, we ask the AMT workers to write pairs of questions and answers about an image. To ensure quality, we instruct the workers to follow three rules: 1) start the questions with one of the ``seven Ws'' (\qa{who}, \qa{what}, \qa{where}, \qa{when}, \qa{why}, \qa{how} and \qa{which}); 2) avoid ambiguous and speculative questions; 3) be precise and unique, and relate the question to the image such that it is clearly answerable if and only if the image is shown.

We collected two separate types of QAs: freeform QAs and region-based QAs. In freeform QA, we ask a worker to look at an image and write eight QA pairs about it. To encourage diversity, we enforce that workers write at least three different Ws out of the seven in their eight pairs. In region-based QA, we ask the workers to write a pair based on a given region. We select the regions that have large areas (more than 5k pixels) and long phrases (more than 4 words). This enables us to collect around twenty region-based pairs at the same cost of the eight freeform QAs. In general, freeform QA tends to yield more diverse QA pairs that enrich the question distribution; region-based QA tends to produce more factual QA pairs at a lower cost.

\subsection{Verification}
All Visual Genome data go through a verification stage as soon as they are annotated. This stage helps eliminate incorrectly labeled objects, attributes, and relationships. It also helps remove region descriptions and questions and answers that might be correct but are vague (``This person seems to enjoy the sun.''), subjective (``room looks dirty''), or opinionated (``Being exposed to hot sun like this may cause cancer''). 

Verification is conducted using two separate strategies: majority voting~\cite{snow2008cheap} and rapid judgments~\cite{krishnaembracing}. All components of the dataset except objects are verified using majority voting. Majority voting\cite{snow2008cheap} involves three unique workers looking at each annotation and voting on whether it is factually correct. An annotation is added to our dataset if at least two (a majority) out of the three workers verify that it is correct. 

We only use rapid judgments to speed up the verification of the objects in our dataset. Meanwhile, rapid judgments~\cite{krishnaembracing} use an interface inspired by rapid serial visual processing that enable verification of objects with an order of magnitude increase in speed than majority voting.

\subsection{Canonicalization}
All the descriptions and QAs that we collect are freeform worker-generated texts. They are not constrained by any limitations. For example, we do not force workers to refer to a man in the image as a \object{man}. We allow them to choose to refer to the man as \object{person}, \object{boy}, \object{man}, etc. This ambiguity makes it difficult to collect all instances of \object{man} from our dataset. In order to reduce the ambiguity in the concepts of our dataset and connect it to other resources used by the research community, we map all objects, attributes, relationships, and noun phrases in region descriptions and QAs to synsets in WordNet~\cite{miller1995WordNet}. In the example above, \object{person}, \object{boy}, and \object{man} would map to the synsets: \synset{person.n.01 (a human being)}, \synset{male\_child.n.01 (a youthful male person)} and \synset{man.n.03 (the generic use of the word to refer to any human being)} respectively. Thanks to the WordNet hierarchy it is now possible to fuse those three expressions of the same concept into \synset{person.n.01 (a human being)} since this is the lowest common ancestor node of all aforementioned synsets.

We use the Stanford NLP tools~\cite{manning-EtAl:2014:P14-5} to extract the noun phrases from the region descriptions and QAs. Next, we map them to their most frequent matching synset in WordNet according to WordNet lexeme counts. We then refine this simple heuristic by hand-crafting mapping rules for the 30  most  common  failure  cases. For example according to WordNet's lexeme counts the most common semantic for ``table'' is \synset{table.n.01 (a set of data arranged in rows and columns)}. However in our data it is more likely to see pieces of furniture and therefore bias the mapping towards \synset{table.n.02 (a piece of furniture having a smooth flat top that is usually supported by one or more vertical legs)}. The objects in our scene graphs are already noun phrases and are mapped to WordNet in the same way.

We normalize each attribute based on morphology (so called ``stemming'') and map them to the WordNet adjectives. We include 15 hand-crafted rules to address common failure cases, which typically occur when the concrete or spatial sense of the word seen in an image is not the most common overall sense. For example, the synset \synset{long.a.02 (of relatively great or greater than average spatial extension)} is less common in WordNet than \synset{long.a.01 (indicating a relatively great or greater than average duration of time)}, even though instances of the word ``long'' in our images are much more likely to refer to that spatial sense.

For relationships, we ignore all prepositions as they are not recognized by WordNet. Since the meanings of verbs are highly dependent upon their morphology and syntactic placement (e.g.\ passive cases, prepositional phrases), we try to find WordNet synsets whose sentence frames match with the context of the relationship. Sentence frames in WordNet are formalized syntactic frames in which a certain sense of a word might appear; for example, \synset{play.v.01: participate in games or sport} occurs in the sentence frames ``Somebody [play]s'' and ``Somebody [play]s something.'' For each verb-synset pair, we then consider the root hypernym of that synset to reduce potential noise from WordNet's fine-grained sense distinctions. The WordNet hierarchy for verbs is segmented and originates from over $100$ root verbs. For example, \synset{draw.v.01: cause to move by pulling} traces back to the root hypernym \synset{move.v.02: cause to move or shift into a new position}, while \synset{draw.v.02: get or derive} traces to the root \synset{get.v.01: come into the possession of something concrete or abstract}. We also include $20$ hand-mapped rules, again to  correct  for  WordNet's  lower  representation  of  concrete or spatial senses.

These mappings are not perfect and still contain some ambiguity. Therefore, we send all our mappings along with the top four alternative synsets for each term to Amazon Mechanical Turk. We ask workers to verify that our mapping was accurate and change the mapping to an alternative one if it was a better fit. We present workers with the concept we want to canonicalize along with our proposed corresponding synset with 4 additional options. To prevent workers from always defaulting to the our proposed synset, we do not explicitly specify which one of the 5 synsets presented is our proposed synset. Section~\ref{sec:canonicalization_stats} provides experimental precision and recall scores for our canonicalization strategy.


\clearpage
\begin{figure}[h]
 \centering
   \includegraphics[width=0.5\textwidth]{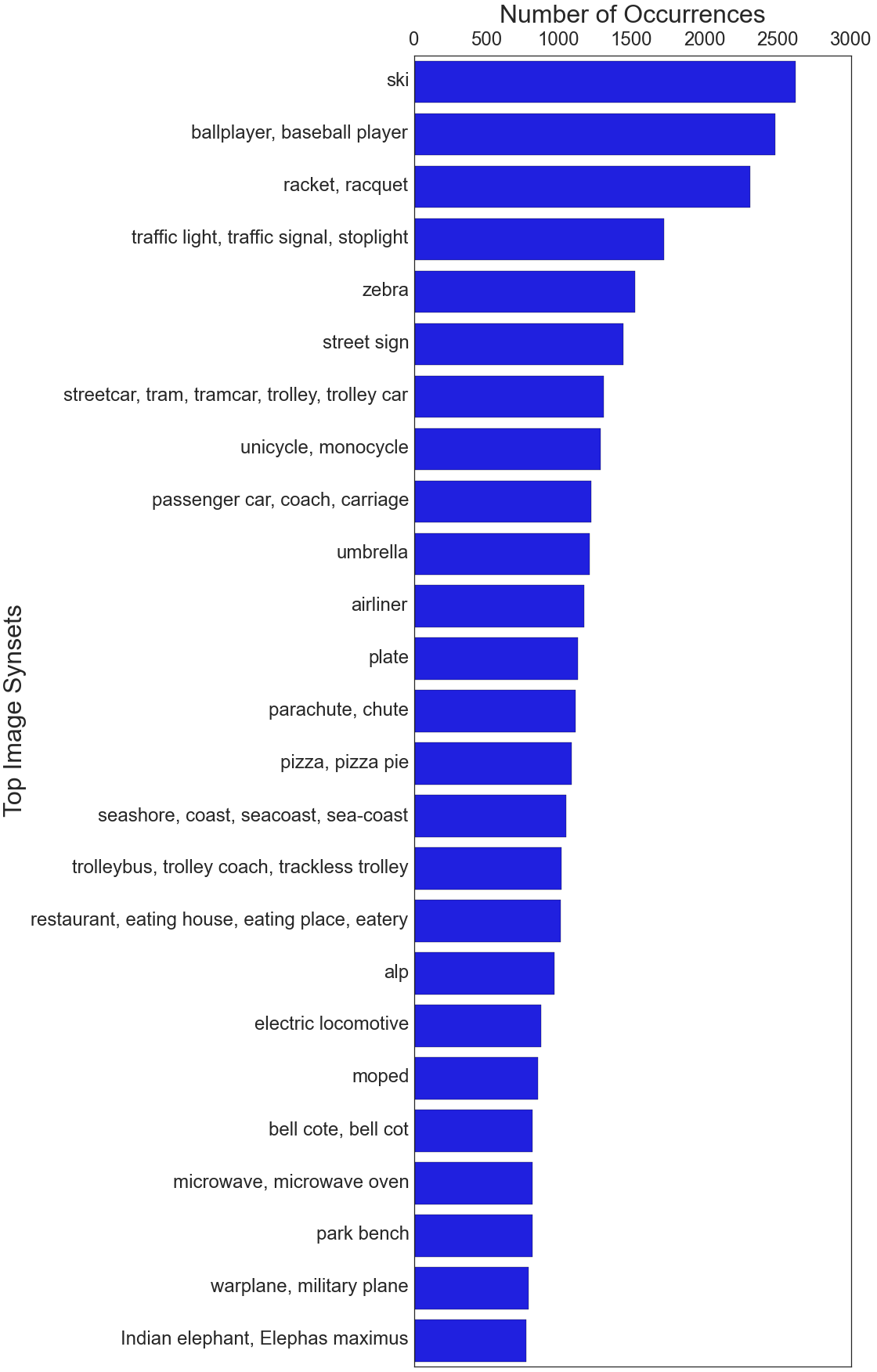}
\caption{A distribution of the top $25$ image synsets in the Visual Genome dataset. A variety of synsets are well represented in the dataset, with the top $25$ synsets having at least $800$ example images each.}
\label{fig:synsets}
\end{figure}

\section{Dataset Statistics and Analysis}
\label{sec:dataset_statistics}

In this section, we provide statistical insights and analysis for each component of Visual Genome. Specifically, we examine the distribution of \textit{images} (Section~\ref{sec:image_stats}) and the collected data for \textit{region descriptions} (Section~\ref{sec:region_stats}) and \textit{questions and answers} (Section~\ref{sec:qa_stats}). We analyze \textit{region graphs} and \textit{scene graphs} together in one section (Section~\ref{sec:graph_stats}), but we also break up these graph structures into their three constituent parts---\textit{objects} (Section~\ref{sec:object_stats}), \textit{attributes} (Section~\ref{sec:attribute_stats}), and \textit{relationships} (Section~\ref{sec:relationship_stats})---and study each part individually. Finally, we describe our canonicalization pipeline and results (Section~\ref{sec:canonicalization_stats}).

\begin{figure}[t]
 \centering
    \iftoggle{smallfigs}{
        \subfloat[]{{\includegraphics[width=0.5\textwidth]{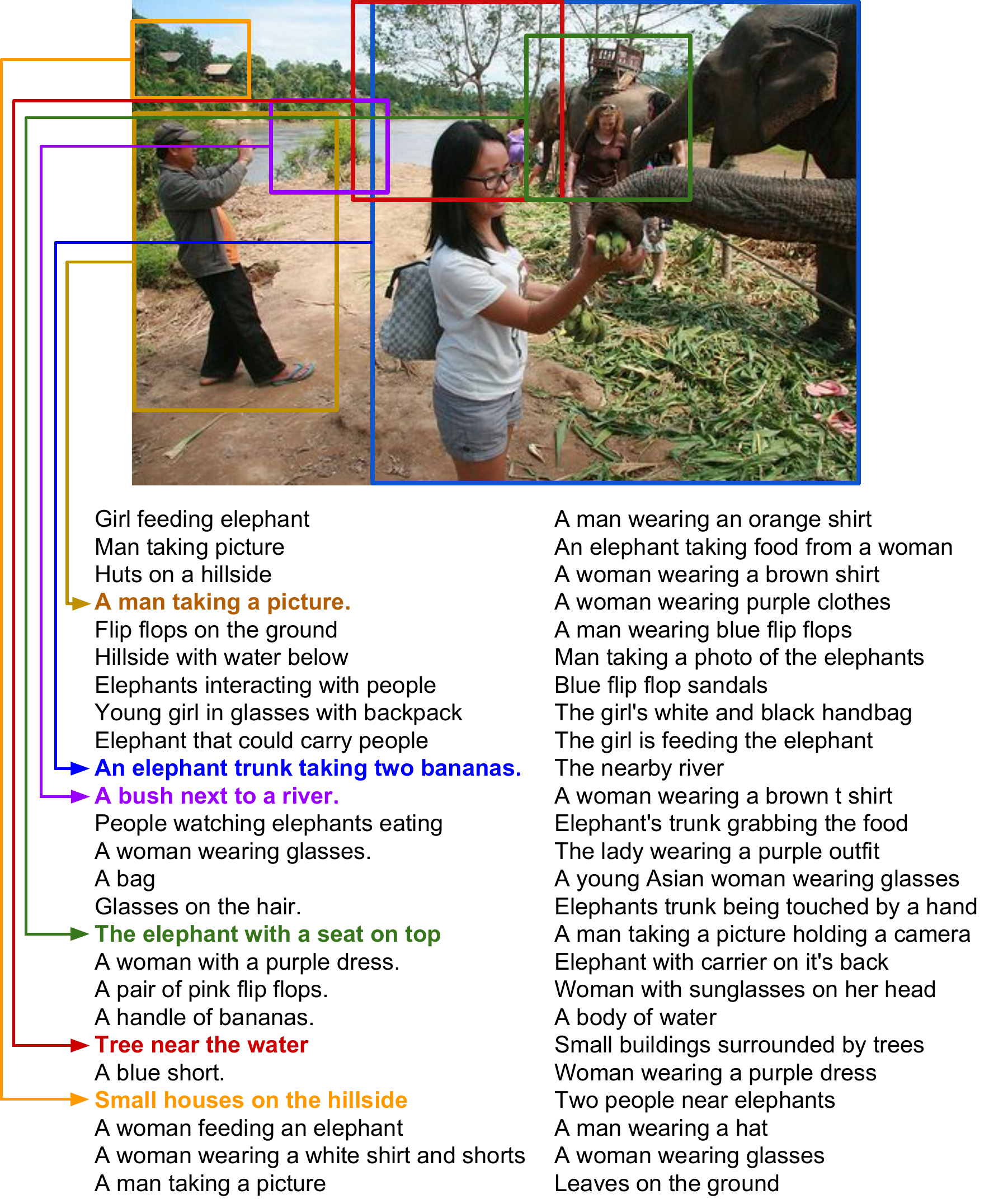} }}%
    }{
        \subfloat[]{{\includegraphics[width=0.5\textwidth]{png_graphics/region_drawing.png} }}%
    }
  \qquad
    \iftoggle{smallfigs}{
        \subfloat[]{{\includegraphics[width=0.35\textwidth]{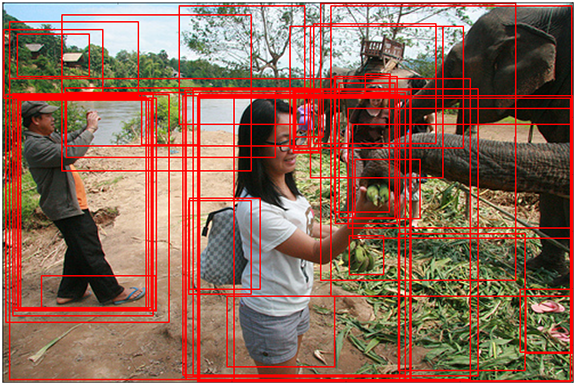} }}%
    }{
        \subfloat[]{{\includegraphics[width=0.35\textwidth]{png_graphics/region_drawing_boxes.png} }}%
    }
    \caption{(a) An example image from the dataset with its region descriptions. We only display localizations for $6$ of the $42$ descriptions to avoid clutter; all 50 descriptions do have corresponding bounding boxes. (b) All $42$ region bounding boxes visualized on the image.}
\label{fig:region_drawing}
\end{figure}

\label{sec:region_stats}
\begin{figure*}[t]%
    \centering
    \subfloat[]{{\includegraphics[width=0.46\textwidth]{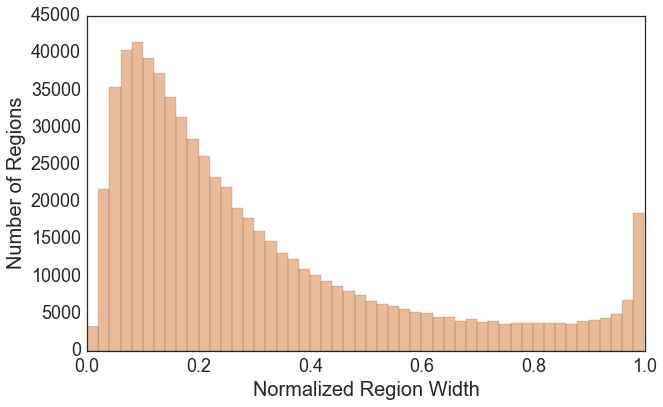} }}%
    \qquad
    \subfloat[]{{\includegraphics[width=0.46\textwidth]{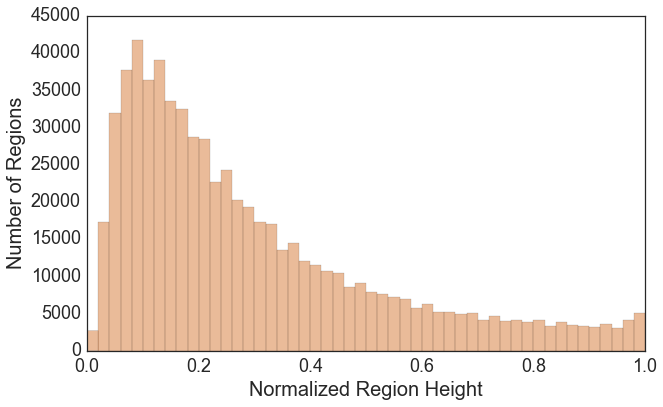} }}%
    \caption{(a) A distribution of the width of the bounding box of a region description normalized by the image width. (b) A distribution of the height of the bounding box of a region description normalized by the image height.}%
    \label{fig:region_distributions}%
\end{figure*}

\subsection{Image Selection}
\label{sec:image_stats}
The Visual Genome dataset consists of all $108,249$ images from the intersection of MS-COCO's~\cite{lin2014microsoft} $328,000$ images and YFCC's~\cite{thomee2015yfcc100m} $100$ million images. These images are real-world, non-iconic images that were uploaded onto Flickr by users. The images range from as small as $72$ pixels wide to as large as $1280$ pixels wide, with an average width of $500$ pixels. We collected the WordNet synsets into which our $108,249$ images can be categorized using the same method as ImageNet~\cite{deng2009imagenet}. Visual Genome images cover $972$ synsets. Figure~\ref{fig:synsets} shows the top synsets to which our images belong. ``ski'' is the most common synset, with $2612$ images; it is followed by ``ballplayer'' and ``racket,'' with all three synsets referring to images of people playing sports. Our dataset is somewhat biased towards images of people, as Figure~\ref{fig:synsets} shows; however, they are quite diverse overall, as the top $25$ synsets each have over $800$ images, while the top $50$ synsets each have over 500 examples.

\begin{figure}[t]
 \centering
   \includegraphics[width=0.5\textwidth]{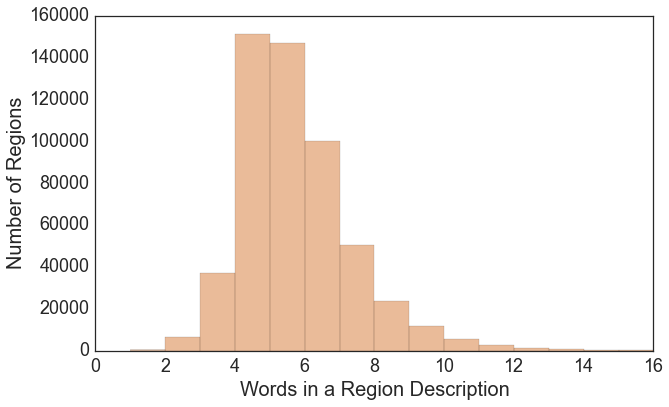}
\caption{A distribution of the number of words in a region description. The average number of words in a region description is $5$, with shortest descriptions of $1$ word and longest descriptions of $16$ words.}
\label{fig:region_length}
\end{figure}

\begin{figure}[t]
 \centering
   \includegraphics[width=0.4\textwidth]{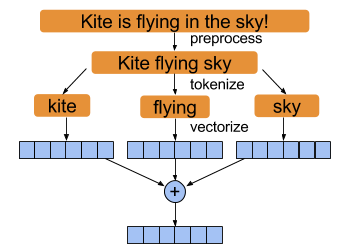}
\caption{The process used to convert a region description into a 300-dimensional vectorized representation.}
\label{fig:region_clustering_pipeline}
\end{figure}

\subsection{Region Description Statistics}

One of the primary components of Visual Genome is its region descriptions. Every image includes an average of $42$ regions with a bounding box and a descriptive phrase. Figure~\ref{fig:region_drawing} shows an example image from our dataset with its $50$ region descriptions. We display bounding boxes for only $6$ out of the $50$ descriptions in the figure to avoid clutter. These descriptions tend to be highly diverse and can focus on a single object, like in ``A bag,'' or on multiple objects, like in ``Man taking a photo of the elephants.'' They encompass the most salient parts of the image, as in ``An elephant taking food from a woman,'' while also capturing the background, as in ``Small buildings surrounded by trees.''

MS-COCO~\cite{lin2014microsoft} dataset is good at generating variations on a single scene-level descriptor. Consider three sentences from MS-COCO dataset on a similar image: ``there is a person petting a very large elephant,'' ``a person touching an elephant in front of a wall,'' and ``a man in white shirt petting the cheek of an elephant.'' These three sentences are single scene-level descriptions. In comparison, Visual Genome descriptions emphasize different regions in the image and thus are less semantically similar. To ensure diversity in the descriptions, we use BLEU score~\cite{papineni2002bleu} thresholds between new descriptions and all previously written descriptions. More information about crowdsourcing can be found in Section~\ref{sec:crowdsourcing_pipeline}.

Region descriptions must be specific enough in an image to describe individual objects, like in the description ``A bag,'' but they must also be general enough to describe high-level concepts in an image, like ``An man being chased by a bear.''  Qualitatively, we note that regions that cover large portions of the image tend to be general descriptions of an image, while regions that cover only a small fraction of the image tend to be more specific. In Figure~\ref{fig:region_distributions} (a), we show the distribution of regions over the width of the region normalized by the width of the image. We see that the majority of our regions tend to be around $10\%$ to $15\%$ of the image width. We also note that there are a large number of regions covering $100\%$ of the image width. These regions usually include elements like ``sky,'' ``ocean,'' ``snow,'' ``mountains,'' etc.\ that cannot be bounded and thus span the entire image width. In Figure~\ref{fig:region_distributions} (b), we show a similar distribution over the normalized height of the region. We see a similar overall pattern, as most of our regions tend to be very specific descriptions of about $10\%$ to $15\%$ of the image height. Unlike the distribution over width, however, we do not see a increase in the number of regions that span the entire height of the image, as there are no common visual equivalents that span images vertically. Out of all the descriptions gathered, only one or two of them tend to be global scene descriptions that are similar to MS-COCO~\cite{lin2014microsoft}.

\begin{figure*}[ht]%
    \centering
    \subfloat[]{{\includegraphics[width=0.48\textwidth]{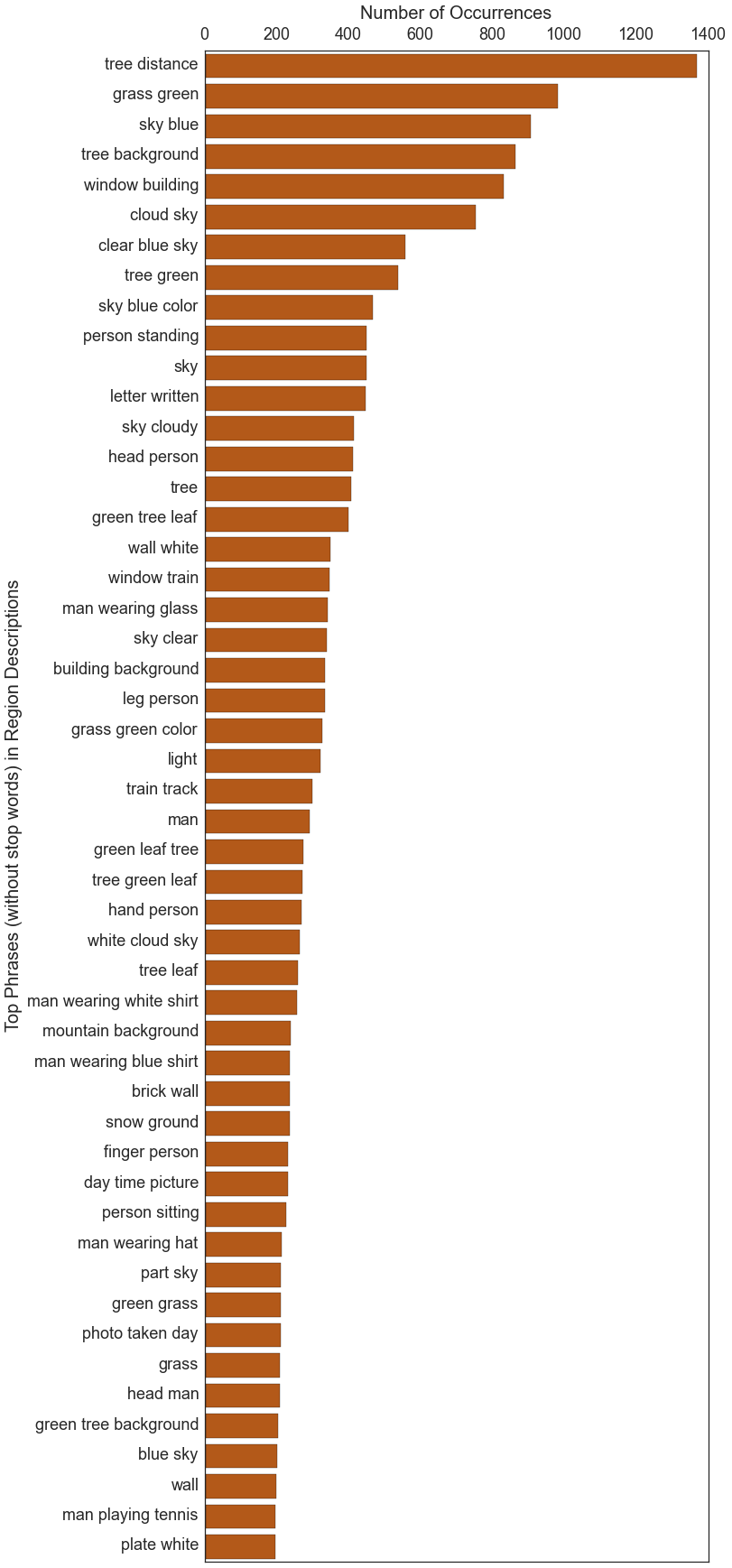} }}%
    \qquad
    \subfloat[]{{\includegraphics[width=0.43\textwidth]{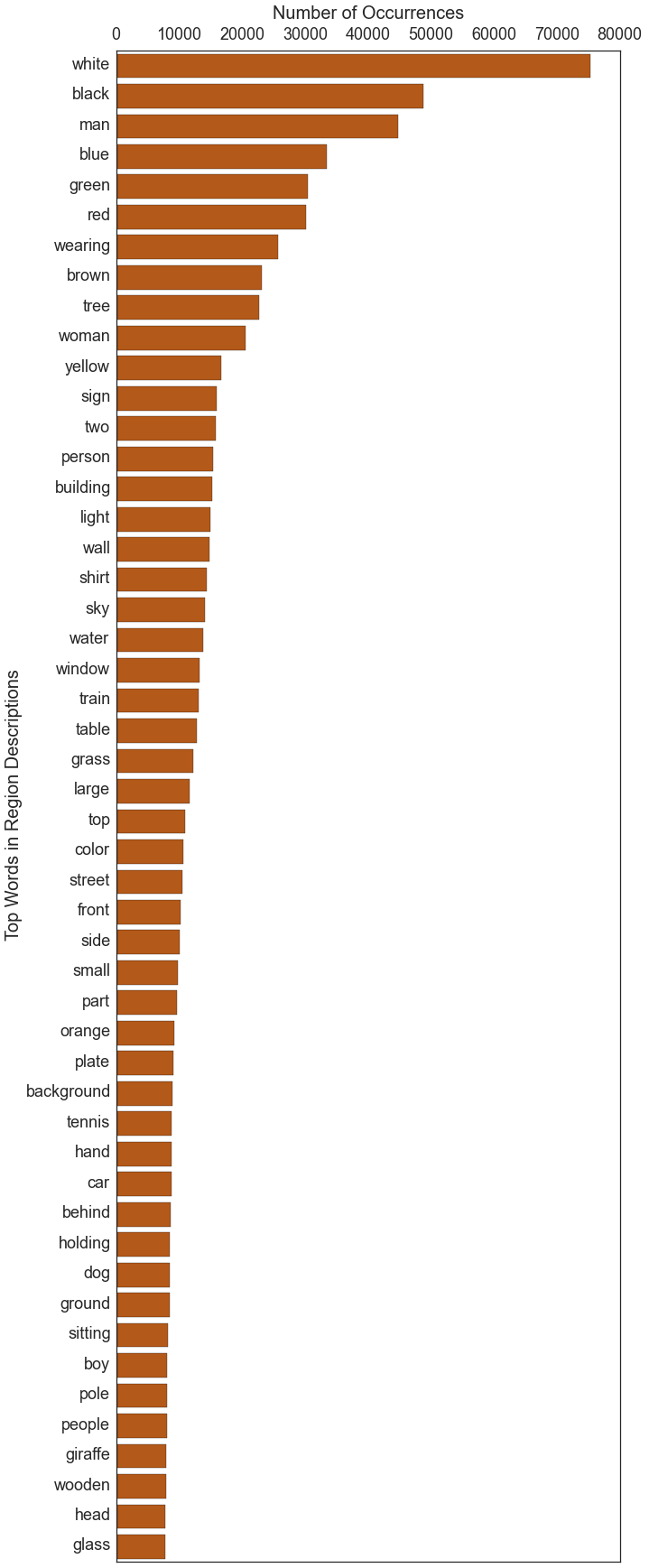} }}%
    \caption{(a) A plot of the most common visual concepts or phrases that occur in region descriptions. The most common phrases refer to universal visual concepts like ``blue sky,'' ``green grass,'' etc. (b) A plot of the most frequently used words in region descriptions. Colors occur the most frequently, followed by common objects like ``man'' and ``dog'' and universal visual concepts like ``sky.''}%
    \label{fig:region_top_phrases_words}%
\end{figure*}

After examining the distribution of the size of the regions described, it is also valuable to look at the semantic information captured by these descriptions. In Figure~\ref{fig:region_length}, we show the distribution of the length (word count) of these region descriptions. The average word count for a description is 5 words, with a minimum of 1 word and a maximum of 12 words. In Figure~\ref{fig:region_top_phrases_words} (a), we plot the most common phrases occurring in our region descriptions, with stop words removed. Common visual elements like ``green grass,'' ``tree [in] distance,'' and ``blue sky'' occur much more often than other, more nuanced elements like ``fresh strawberry.'' We also study descriptions with finer precision in Figure~\ref{fig:region_top_phrases_words} (b), where we plot the most common words used in descriptions. Again, we eliminate stop words from our study. Colors like ``white'' and ``black'' are the most frequently used words to describe visual concepts; we conduct a similar study on other captioning datasets including MS-COCO~\cite{lin2014microsoft} and Flickr 30K~\cite{young2014image} and find a similar distribution with colors occurring most frequently. Besides colors, we also see frequent occurrences of common objects like ``man,'' ``tree,'' and ``sign'' and of universal visual elements like ``sky.''

\begin{figure*}[t]%
    \centering
    \subfloat[]{{
    \iftoggle{smallfigs}{
        \includegraphics[width=0.96\textwidth]{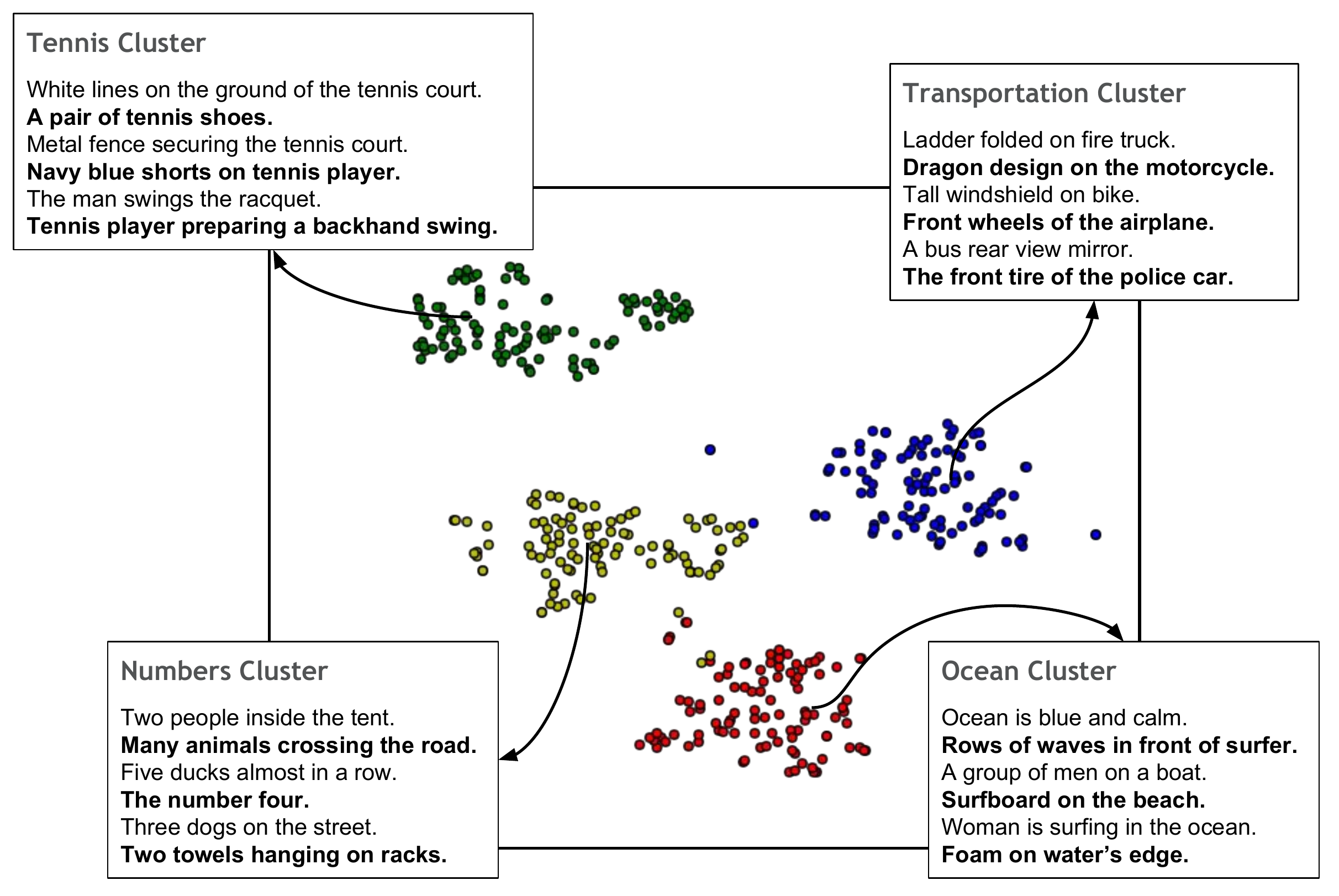}
    }{
        \includegraphics[width=0.96\textwidth]{png_graphics/cluster_infographic.png}
    }
    }}%
    \qquad
    \subfloat[]{{\includegraphics[width=0.46\textwidth]{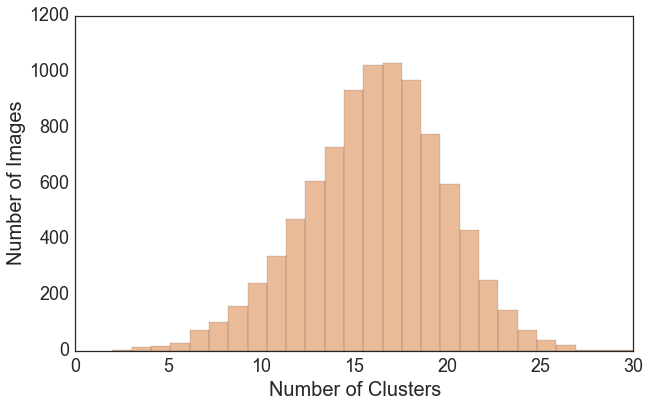} }}%
    \qquad
    \subfloat[]{{\includegraphics[width=0.46\textwidth]{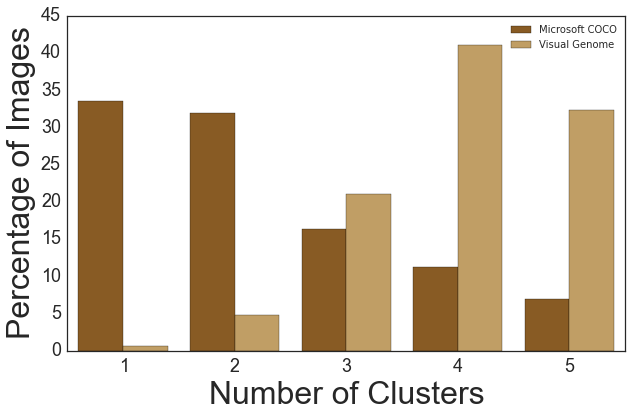} }}%
    \caption{(a) Example illustration showing four clusters of region descriptions and their overall themes. Other clusters not shown due to limited space. (b) Distribution of images over number of clusters represented in each image's region descriptions. (c) We take Visual Genome with 5 random descriptions taken from each image and MS-COCO dataset with all 5 sentence descriptions per image and compare how many clusters are represented in the descriptions. We show that Visual Genome's descriptions are more varied for a given image, with an average of 4 clusters per image, while MS-COCO's images have an average of 3 clusters per image.}
    \label{fig:region_clustering}
\end{figure*}

\paragraph{Semantic diversity.} 
We also study the actual semantic contents of the descriptions. We use an unsupervised approach to analyze the semantics of these descriptions. Specifically, we use word2vec~\cite{mikolov2013efficient} to convert each word in a description to a 300-dimensional vector. Next, we remove stop words and average the remaining words to get a vector representation of the whole region description. This pipeline is outlined in Figure~\ref{fig:region_clustering_pipeline}. We use hierarchical agglomerative clustering on vector representations of each region description and find 71 semantic and syntactic groupings or ``clusters.'' Figure~\ref{fig:region_clustering} (a) shows four such example clusters. One cluster contains all descriptions related to tennis, like ``A man swings the racquet'' and ``White lines on the ground of the tennis court,'' while another cluster contains descriptions related to numbers, like ``Three dogs on the street'' and ``Two people inside the tent.'' To quantitatively measure the diversity of Visual Genome's region descriptions, we calculate the number of clusters represented in a single image's region descriptions. We show the distribution of the variety of descriptions for an image in Figure~\ref{fig:region_clustering} (b). We find that on average, each image contains descriptions from 17 different clusters. The image with the least diverse descriptions contains descriptions from 4 clusters, while the image with the most diverse descriptions contains descriptions from 26 clusters.

Finally, we also compare the descriptions in Visual Genome to the captions in MS-COCO\@. First we aggregate all Visual Genome and MS-COCO descriptions and remove all stop words. After removing stop words, the descriptions from both datasets are roughly the same length. We conduct a similar study, in which we vectorize the descriptions for each image and calculate each dataset's cluster diversity per image. We find that on average, 2 clusters are represented in the captions for each image in MS-COCO, with very few images in which 5 clusters are represented. Because each image in MS-COCO only contains 5 captions, it is not a fair comparison to compare the number of clusters represented in all the region descriptions in the Visual Genome dataset. We thus randomly sample 5 Visual Genome region descriptions per image and calculate the number of clusters in an image. We find that Visual Genome descriptions come from 4 or 5 clusters. We show our comparison results in Figure~\ref{fig:region_clustering} (c). The difference between the semantic diversity between the two datasets is statistically significant ($t=-240$, $p<0.01$).

\clearpage

\newcolumntype{P}[1]{>{\centering\arraybackslash}p{#1}}

\tabcolsep=0.05cm
\begin{table*}[h]
\centering
\begin{tabular}{p{3cm}P{2cm}P{2cm}P{2cm}P{2cm}P{2cm}P{2cm}P{2cm}}
& Visual Genome & ILSVRC Det. \cite{ILSVRC15} & MS-COCO \cite{lin2014microsoft} & Caltech101 \cite{fei2007learning} & Caltech256 \cite{griffin2007caltech} & PASCAL Det. \cite{everingham2010pascal} & Abstract Scenes \cite{zitnick2013bringing} \\ 
 \midrule
Images & 108,249 & 476,688 & 328,000 & 9,144 & 30,608 & 11,530 & 10,020 \\
Total Objects & 255,718 & 534,309 & 2,500,000 & 9,144 & 30,608 & 27,450 & 58 \\
Total Categories & 18,136 & 200 & 91 & 102 & 257 & 20 & 11 \\
Objects per Category & 14.10 & 2671.50 & 27472.50 & 90 & 119 & 1372.50 & 5.27 \\
\bottomrule
\end{tabular}
\caption{Comparison of Visual Genome objects and categories to related datasets.}
\label{fig:objects_and_categories}
\end{table*}

\begin{figure}[t!]
 \centering
 \subfloat[]{{\includegraphics[width=0.45\textwidth]{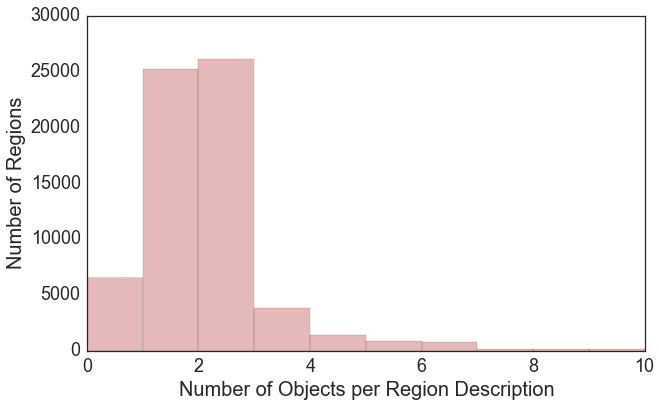} }}%
 \qquad
 \subfloat[]{{\includegraphics[width=0.45\textwidth]{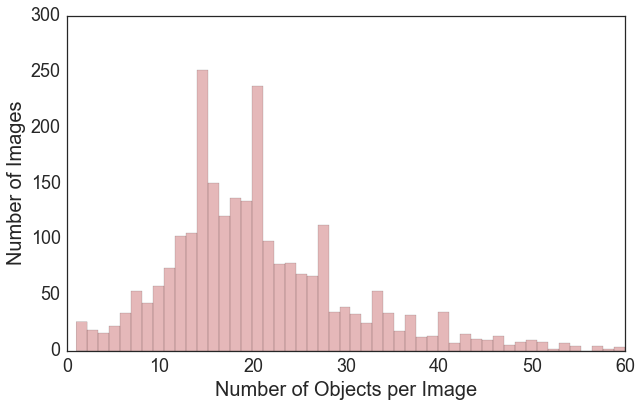} }}%
 \qquad
\caption{(a) Distribution of the number of objects per region. Most regions have between 0 and 2 objects. (b) Distribution of the number of objects per image. Most images contain between 15 and 20 objects.}
\label{fig:objects_region_image}
\end{figure}

\subsection{Object Statistics}
\label{sec:object_stats}
\begin{figure}[t!]
 \centering
 \includegraphics[width=0.5\textwidth]{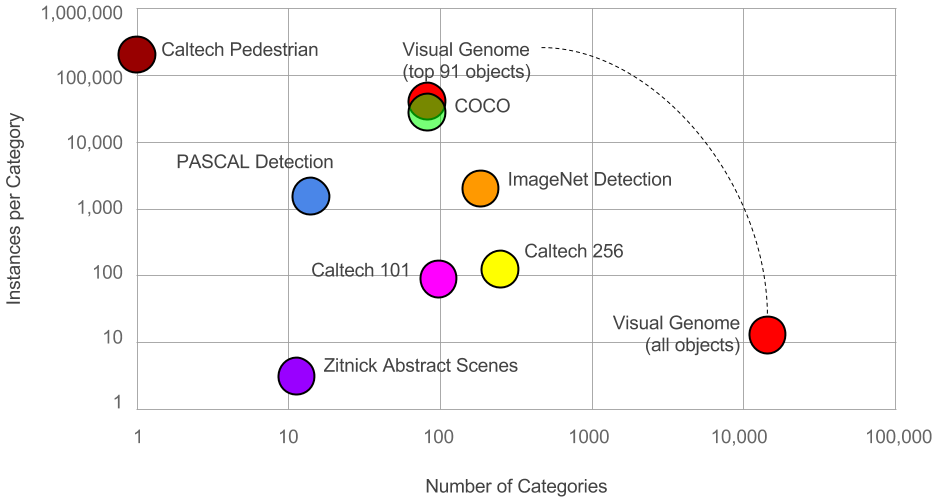}
  \caption{Comparison of object diversity between various datasets. Visual Genome far surpasses other datasets in terms of number of object categories.}
  \label{fig:categories_and_instances}
\end{figure}

In comparison to related datasets, Visual Genome fares well in terms of object density and diversity. Visual Genome contains approximately $21$ objects per image, exceeding ImageNet~\cite{deng2009imagenet}, PASCAL~\cite{everingham2010pascal}, MS-COCO~\cite{lin2014microsoft}, and other datasets by large margins. As shown in Figure~\ref{fig:categories_and_instances}, there are more object categories represented in Visual Genome than in any other dataset. This comparison is especially pertinent with regards to Microsoft MS-COCO~\cite{lin2014microsoft}, which uses the same images as Visual Genome. The lower count of objects per category is a result of our higher number of categories. For a fairer comparison with ILSVRC 2014 Detection~\cite{ILSVRC15}, Visual Genome has about $2239$ objects per category when only the top $200$ categories are considered, which is comparable to ILSVRC's $2671.5$ objects per category. For a fairer comparison with MS-COCO, Visual Genome has about $3768$ objects per category when only the top $91$ categories are considered. This is comparable to MS-COCO's~\cite{lin2014microsoft} when we consider just the $108,249$ MS-COCO images in Visual Genome.

\begin{figure*}[h!]
 \centering
    \iftoggle{smallfigs}{
        \subfloat[]{{\includegraphics[width=0.45\textwidth]{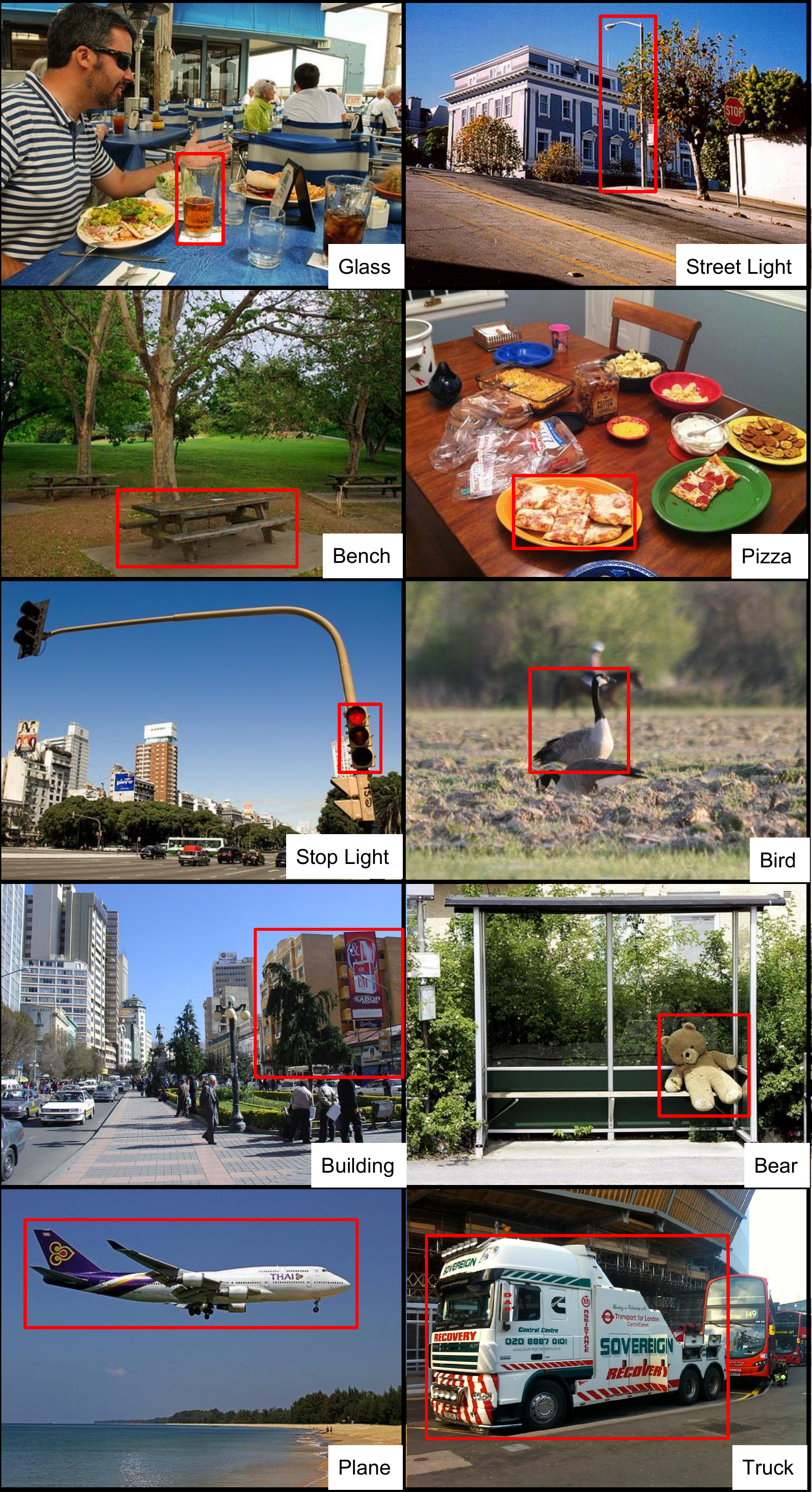} }}%
    }{
        \subfloat[]{{\includegraphics[width=0.45\textwidth]{png_graphics/object_example_classes.png} }}%
    }
   \subfloat[]{{\includegraphics[width=0.40\textwidth]{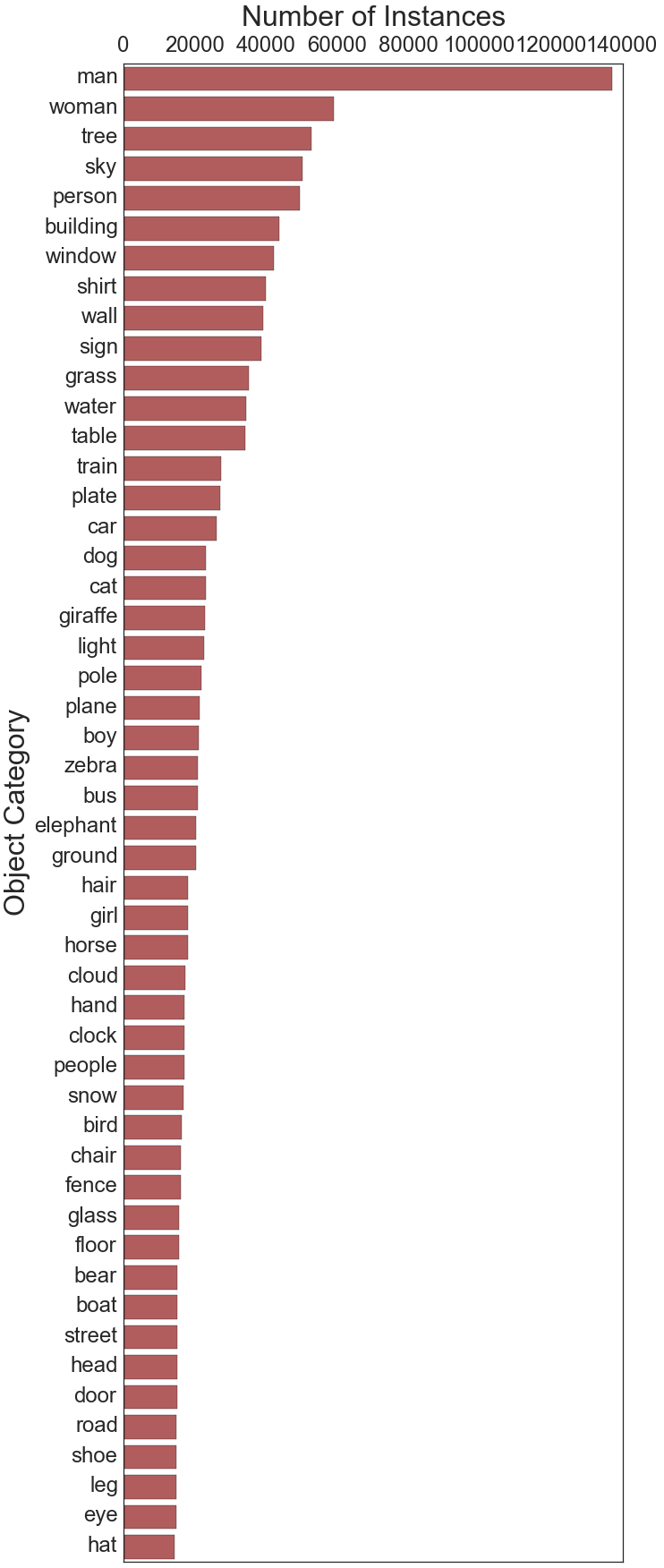} }}%
   \qquad
\caption{(a) Examples of objects in Visual Genome. Each object is localized in its image with a tightly drawn bounding box. (b) Plot of the most frequently occurring objects in images. People are the most frequently occurring objects in our dataset, followed by common objects and visual elements like \object{building}, \object{shirt}, and \object{sky}.}
\label{fig:object_examples_top_objects}
\end{figure*}

Objects in Visual Genome come from a variety of categories. As shown in Figure~\ref{fig:object_examples_top_objects} (b), objects related to WordNet categories such as humans, animals, sports, and scenery are most common; this is consistent with the general bias in image subject matter in our dataset. Common objects like \object{man}, \object{person}, and \object{woman} occur especially frequently with occurrences of $24$K, $17$K, and $11$K. Other objects that also occur in MS-COCO~\cite{lin2014microsoft} are also well represented with around $5000$ instances on average. Figure~\ref{fig:object_examples_top_objects} (a) shows some examples of objects in images. Objects in Visual Genome span a diverse set of Wordnet categories like food, animals, and man-made structures.

It is important to look not only at what types of objects we have but also at the distribution of objects in images and regions. Figure~\ref{fig:objects_region_image} (a) shows, as expected, that we have between 0 and 2 objects in each region on average. It is possible for regions to contain no objects if their descriptions refer to no explicit objects in the image. For example, a region described as ``it is dark outside'' has no objects to extract. Regions with only one object generally have descriptions that focus on the attributes of a single object. On the other hand, regions with two or more objects generally have descriptions that contain both attributes of specific objects and relationships between pairs of objects.

As shown in Figure~\ref{fig:objects_region_image} (b), each image contains on average around $21$ unique objects. Few images have a low number of objects, which we expect since images usually capture more than a few objects. Moreover, few images have an extremely high number of objects (e.g.\ over $40$).

\clearpage


\subsection{Attribute Statistics}
\label{sec:attribute_stats}
Attributes allow for detailed description and disambiguation of objects in our dataset. About $45\%$ of objects in Visual Genome are annotated with at least one attribute; our dataset contains $1.6$ million total attributes with $13,041$ unique attributes. Attributes include colors (\attribute{green}), sizes (\attribute{tall}), continuous action verbs (\attribute{standing}), materials (\attribute{plastic}), etc. Each attribute in our scene graphs belongs to one object, while each object can have multiple attributes. We denote attributes as \attribute{attribute}(\object{object}). 

\begin{figure}[t!]
 \centering
 \subfloat[]{{\includegraphics[width=0.38\textwidth]{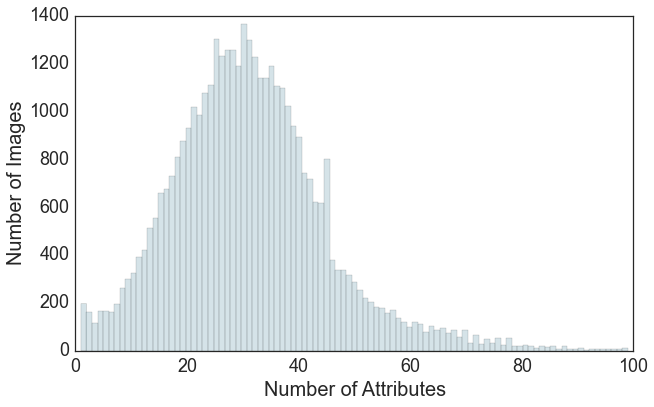}}}
 \qquad
 \subfloat[]{{\includegraphics[width=0.38\textwidth]{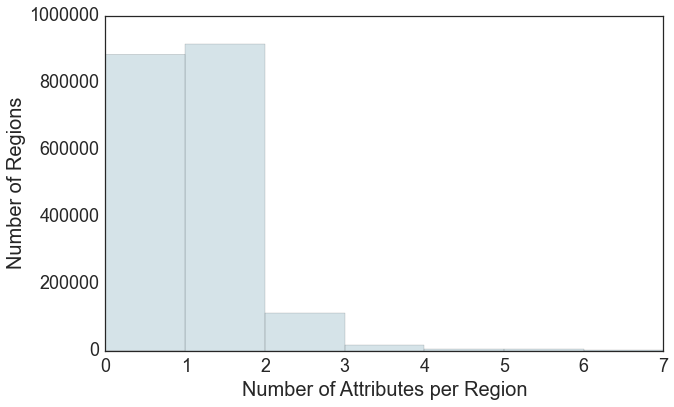}}}
 \qquad
 \subfloat[]{{\includegraphics[width=0.38\textwidth]{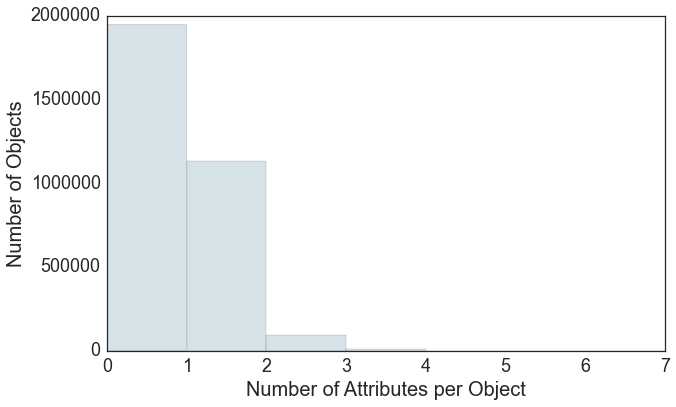}}}
 \qquad
\caption{Distribution of the number of attributes (a) per image, (b) per region description, (c) per object.}
\setlength{\belowcaptionskip}{-0.1cm}
\label{fig:attribute_distribution}
\end{figure}

\begin{figure*}[t]
 \centering
 \subfloat[]{{\includegraphics[width=0.55\textwidth]{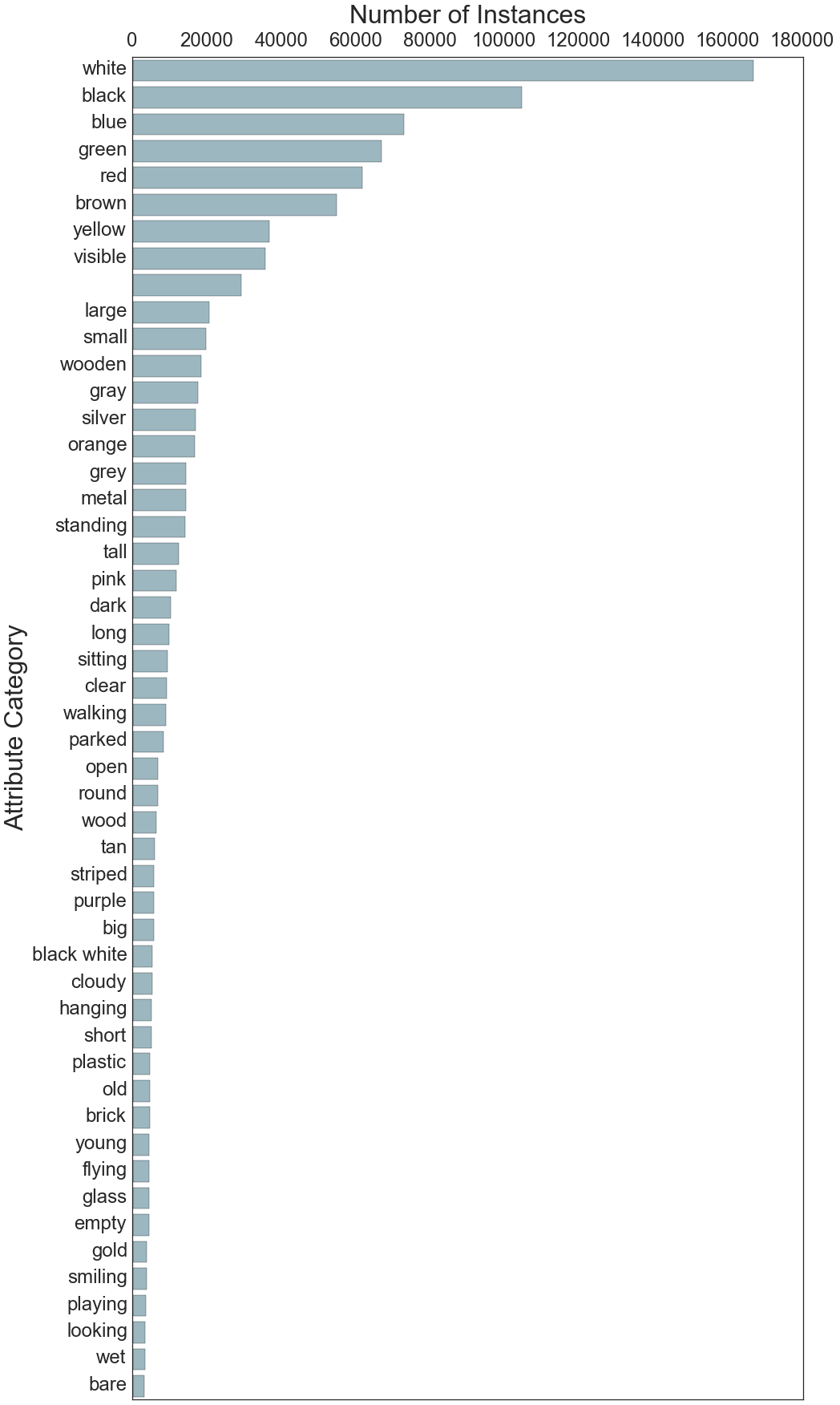}}}
 \subfloat[]{{\includegraphics[width=0.45\textwidth]{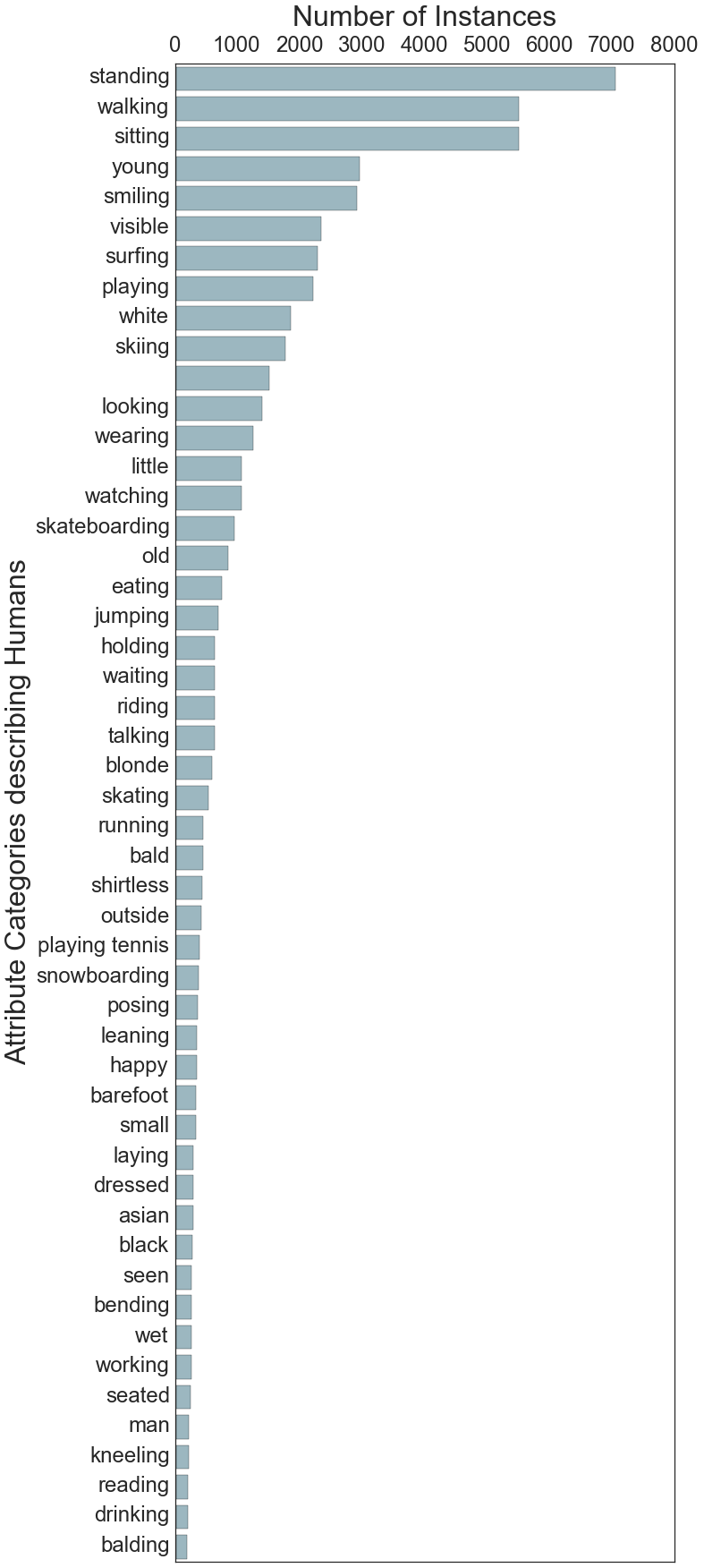}}}
 \qquad
\caption{(a) Distribution showing the most common attributes in the dataset. Colors (\attribute{white}, \attribute{red}) and materials (\attribute{wooden}, \attribute{metal}) are the most common. (b) Distribution showing the number of attributes describing people. State-of-motion verbs (\attribute{standing}, \attribute{walking}) are the most common, while certain sports (\attribute{skiing}, \attribute{surfing}) are also highly represented due to an image source bias in our image set.}
\label{fig:top_attributes}
\end{figure*}

On average, each image in Visual Genome contains $21$ attributes, as shown in Figure~\ref{fig:attribute_distribution}. Each region contains on average $1$ attribute, though about $42\%$ of regions contain no attribute at all; this is primarily because many regions are relationship-focused. Figure~\ref{fig:top_attributes} (a) shows the distribution of the most common attributes in our dataset. Colors (e.g.~\attribute{white}, \attribute{green}) are by far the most frequent attributes. Also common are sizes (e.g.~\attribute{large}) and materials (e.g.~\attribute{wooden}). Figure~\ref{fig:top_attributes} (b) shows the distribution of attributes describing people (e.g.~\object{man}, \object{girls}, and \object{person}). The most common attributes describing people are intransitive verbs describing their states of motion (e.g.\attribute{standing} and \attribute{walking}). Certain sports (\attribute{skiing}, \attribute{surfboarding}) are overrepresented due to a bias towards these sports in our images.

\begin{figure*}[t]
 \centering
 \subfloat[]{{\includegraphics[width=0.9\textwidth]{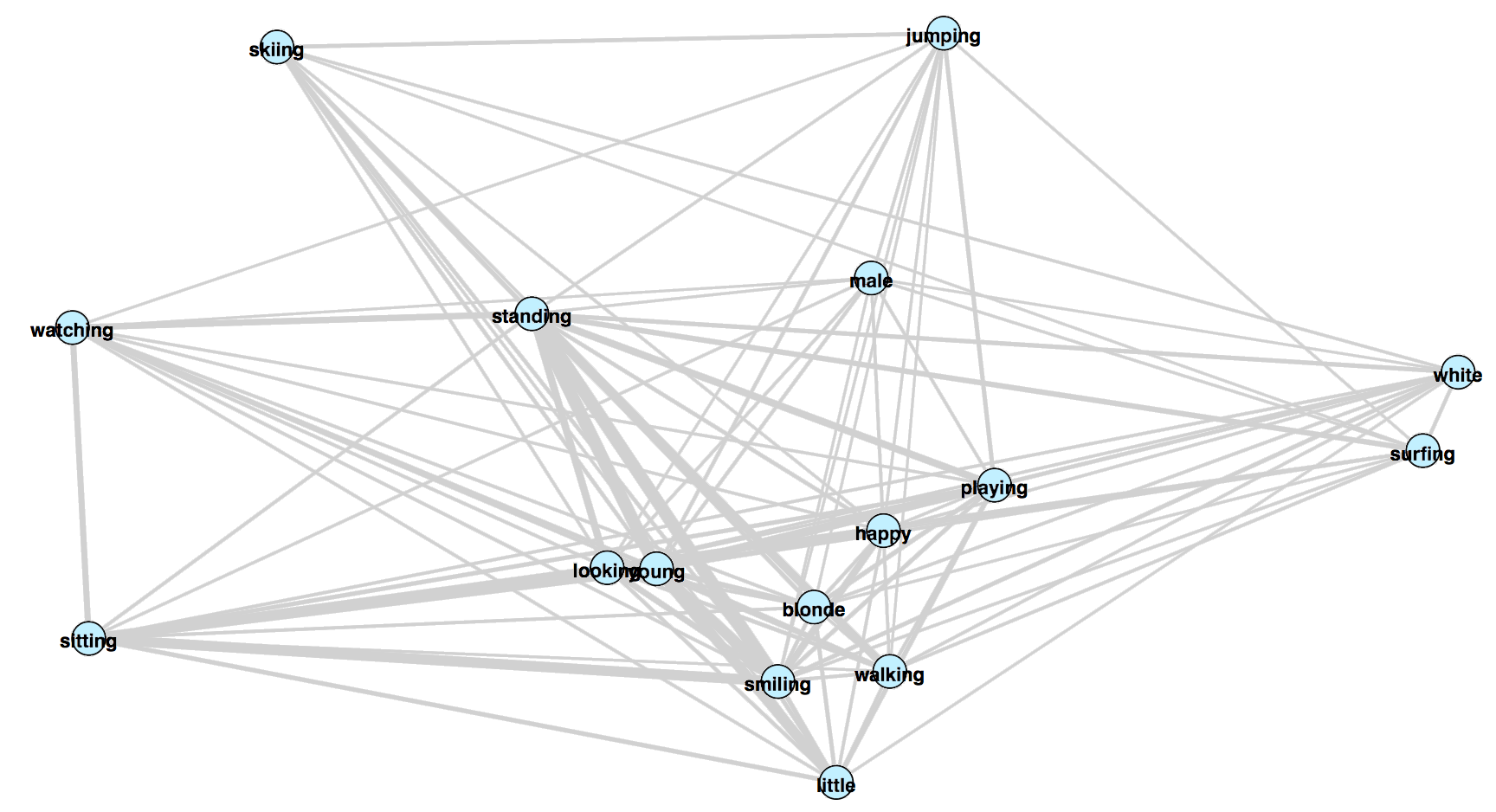}}}
 \qquad
 \subfloat[]{{\includegraphics[width=0.9\textwidth]{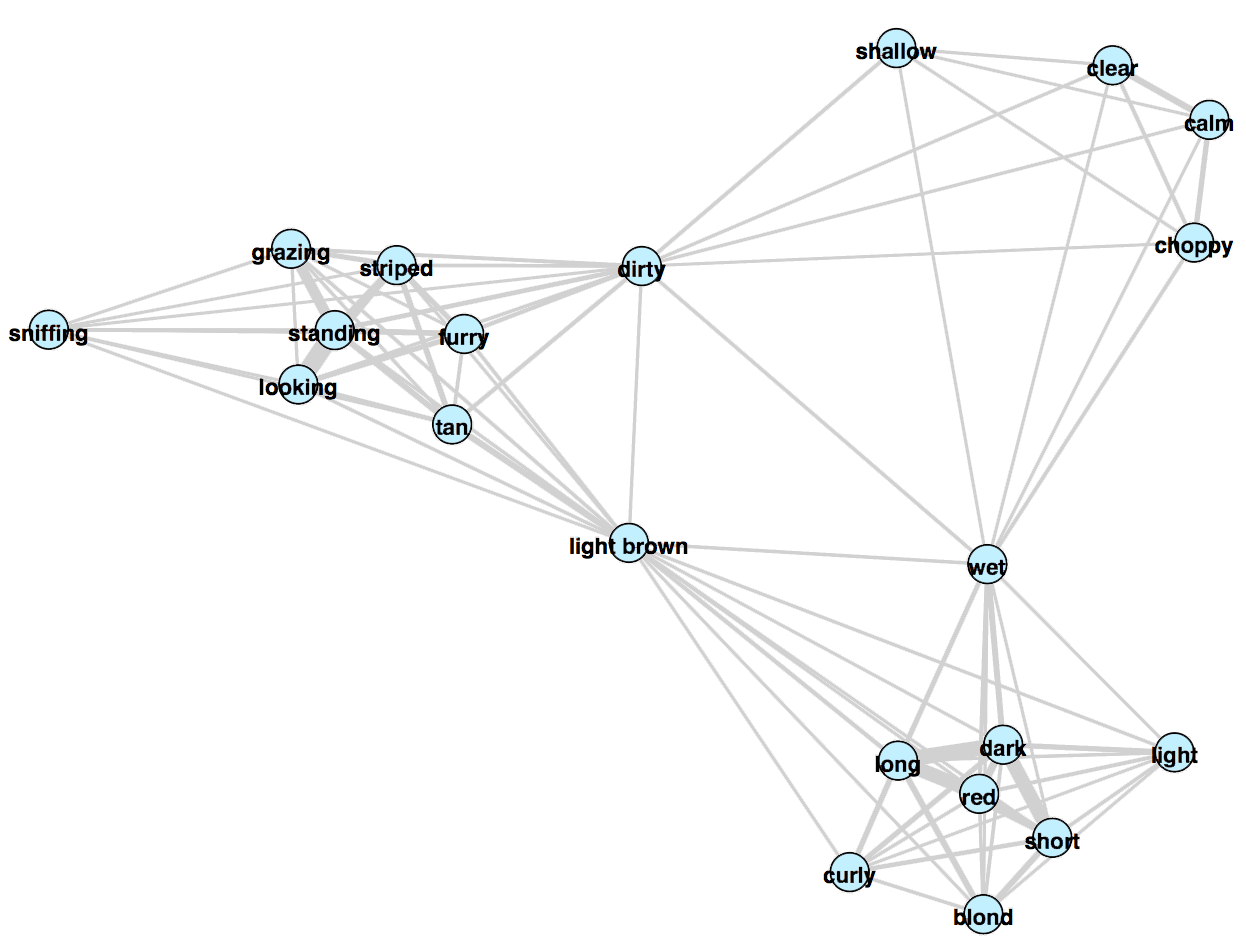}}}
 \qquad
\caption{(a) Graph of the person-describing attributes with the most co-occurrences. Edge thickness represents the frequency of co-occurrence of the two nodes. (b) A subgraph showing the co-occurrences and intersections of three cliques, which appear to describe water (top right), hair (bottom right), and some type of animal (left). Edges between cliques have been removed for clarity.}
\label{fig:attribute_graphs}
\end{figure*}

\vspace{-0.2cm}
\paragraph{Attribute Graphs.} We also qualitatively analyze the attributes in our dataset by constructing co-occurrence graphs, in which nodes are unique attributes and edges connect those attributes that describe the same object. For example, if an image contained a ``large black dog'' (\attribute{large}(\object{dog}), \attribute{black}(\object{dog})) and another image contained a ``large yellow cat'' (\attribute{large}(\object{cat}), \attribute{yellow}(\object{cat})), its attributes would form an incomplete graph with edges (\attribute{large}, \attribute{black}) and (\attribute{large}, \attribute{yellow}). We create two such graphs: one for both the total set of attributes and a second where we consider only objects that refer to people. A subgraph of the 16 most frequently connected (co-occurring) person-related attributes is shown in Figure~\ref{fig:attribute_graphs} (a).

Cliques in these graphs represent groups of attributes in which at least one co-occurrence exists for each pair of attributes in that group. In the previous example, if a third image contained a ``black and yellow taxi'' (\attribute{black}(\object{taxi}), \attribute{yellow}(\object{taxi})), the resulting third edge would create a clique between the attributes \attribute{black}, \attribute{large}, and \attribute{yellow}. When calculated across the entire Visual Genome dataset, these cliques provide insight into commonly perceived traits of different types of objects. Figure~\ref{fig:attribute_graphs} (b) is a selected representation of three example cliques and their overlaps. From just a clique of attributes, we can predict what types of objects are usually referenced. In Figure~\ref{fig:attribute_graphs} (b), we see that these cliques describe an animal (left), water body (top right), and human hair (bottom right). 

Other cliques (not shown) can also uniquely identify objects. In our set, one clique contains \attribute{athletic}, \attribute{young}, \attribute{fit}, \attribute{skateboarding}, \attribute{focused}, \attribute{teenager}, \attribute{male}, \attribute{skinny}, and \attribute{happy}, capturing some of the common traits of \object{skateboarders} in our set. Another such clique has \attribute{shiny}, \attribute{small}, \attribute{metal}, \attribute{silver}, \attribute{rusty}, \attribute{parked}, and \attribute{empty}, most likely describing a subset of cars. From these cliques, we can thus infer distinct objects and object types based solely on their attributes, potentially allowing for highly specific object identification based on selected characteristics.

\clearpage

\subsection{Relationship Statistics}
\label{sec:relationship_stats}
Relationships are the core components that link objects in our scene graphs. Relationships are directional, i.e.\ they involve two objects, one acting as the subject and one as the object of a predicate relationship. We denote all relationships in the form \relationship{subject}{relationship}{object}. For example, if a \object{man} is \predicate{swinging} a \object{bat}, we write \relationship{man}{swinging}{bat}. Relationships can be spatial (e.g.~\predicate{inside\_of}), action (e.g.~\predicate{swinging}), compositional (e.g.~\predicate{part\_of}), etc. More complex relationships such as \predicate{standing\_on}, which includes both an action and a spatial aspect, are also represented. Relationships are extracted from region descriptions by crowd workers, similarly to attributes and objects. Visual Genome contains a total of $13,894$ unique relationships, with over $1.8$ million total relationships.

\begin{figure}[t]%
    \centering
    \subfloat[]{{\includegraphics[width=0.35\textwidth]{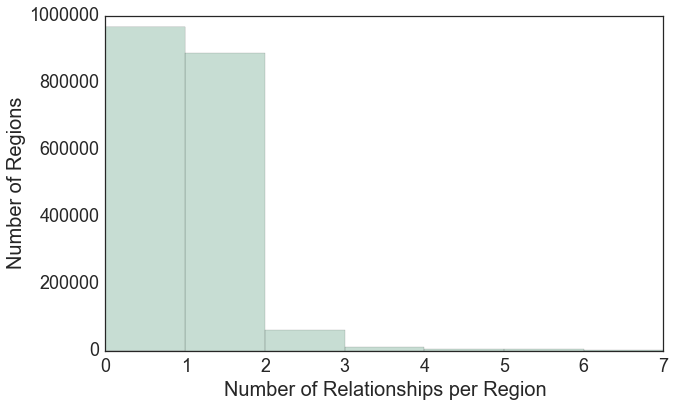} }}%
    \qquad
    \vspace{-2mm}
    \subfloat[]{{\includegraphics[width=0.35\textwidth]{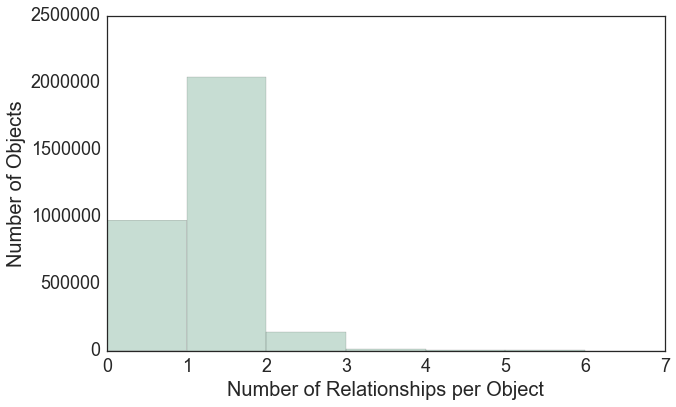} }}%
    \qquad
    \vspace{-2mm}
    \subfloat[]{{\includegraphics[width=0.35\textwidth]{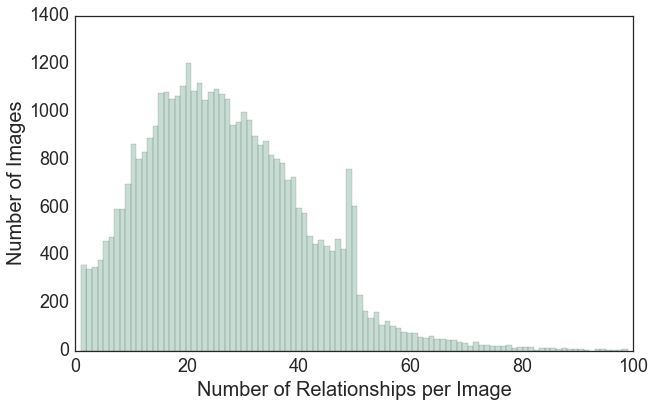} }}%
    \caption{Distribution of relationships (a) per image region, (b) per image object, (c) per image.}%
    \label{fig:relationship_distribution}%
    \setlength{\belowcaptionskip}{-0.1cm}
\end{figure}

Figure~\ref{fig:relationship_distribution} (a) shows the distribution of relationships per region description. On average, we have $1$ relationship per region, with a maximum of $7$. We also have some descriptions like ``an old, tall man,'' which have multiple attributes associated with the \object{man} but no relationships. Figure~\ref{fig:relationship_distribution} (b) is a distribution of relationships per image object. Finally, Figure~\ref{fig:relationship_distribution} (c) shows the distribution of relationships per image. Each image has an average of $19$ relationships, with a minimum of $1$ relationship and with ax maximum of over $60$ relationships.

\begin{figure*}[t]%
    \centering
    \subfloat[]{{\includegraphics[width=0.46\textwidth]{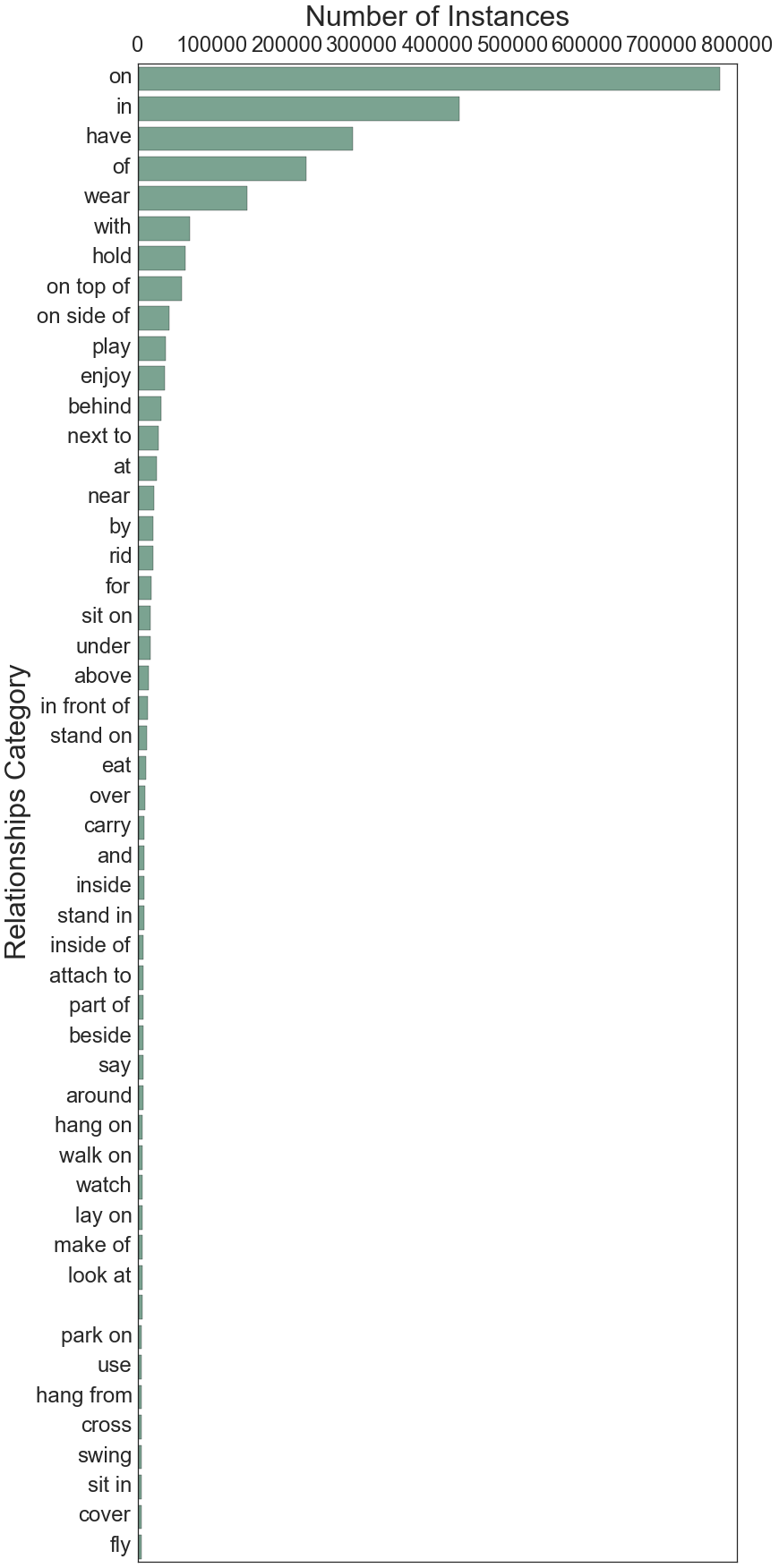} }}%
    \qquad
    \subfloat[]{{\includegraphics[width=0.46\textwidth]{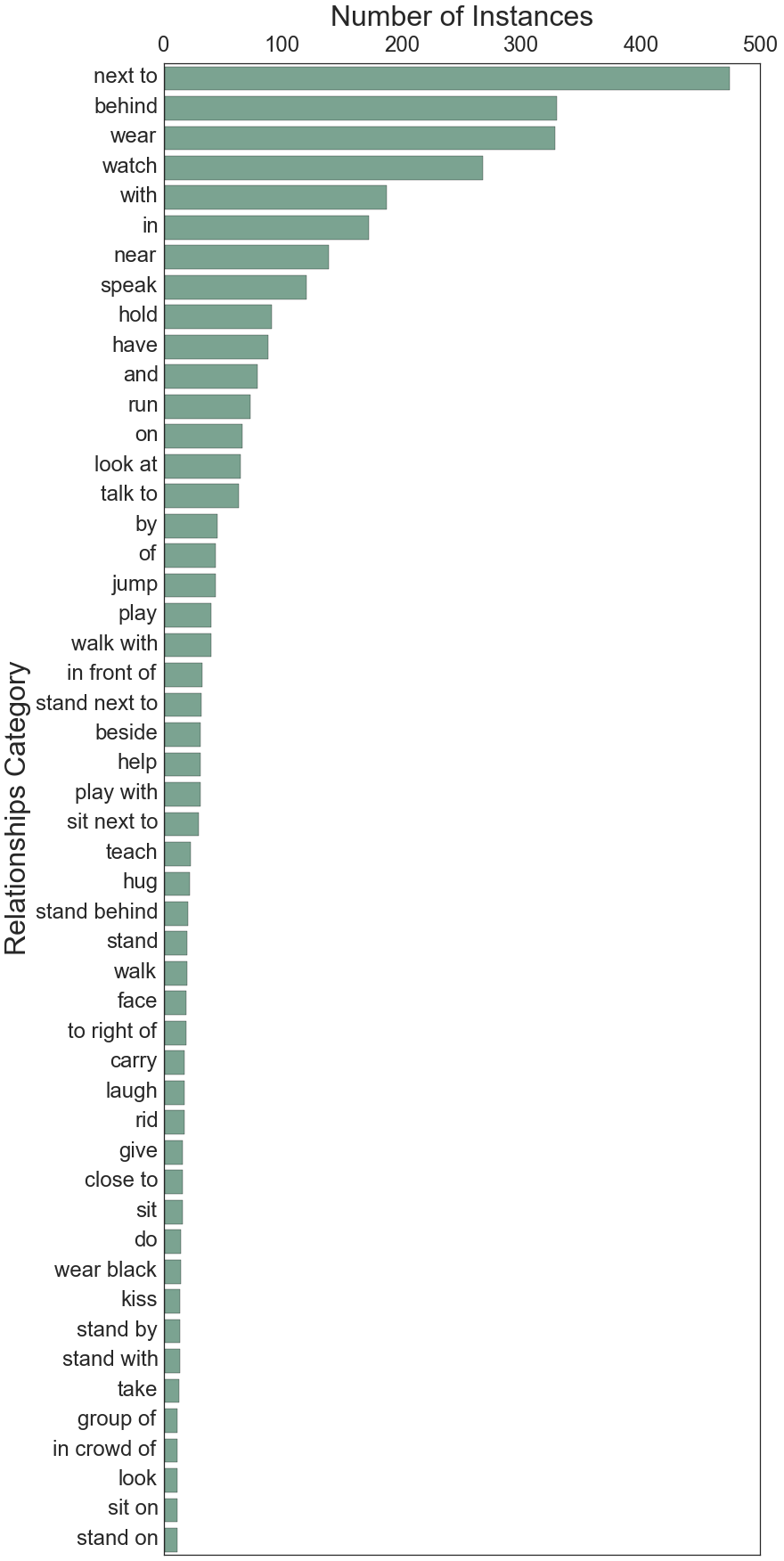} }}%
    \caption{(a) A sample of the most frequent relationships in our dataset. In general, the most common relationships are spatial (\predicate{on top of}, \predicate{on side of}, etc.). (b) A sample of the most frequent relationships involving humans in our dataset. The relationships involving people tend to be more action oriented (\predicate{walk}, \predicate{speak}, \predicate{run}, etc.).}%
    \label{fig:top_predicates}%
\end{figure*}

\vspace{-0.15cm}
\paragraph{Top relationship distributions.} We display the most frequently occurring relationships in Figure~\ref{fig:top_predicates} (a). \predicate{on} is the most common relationship in our dataset. This is primarily because of the flexibility of the word \predicate{on}, which can refer to spatial configuration (\predicate{on top of}), attachment (\predicate{hanging on}), etc. Other common relationships involve actions like \predicate{holding} and \predicate{wearing} and spatial configurations like \predicate{behind}, \predicate{next to}, and \predicate{under}.  Figure~\ref{fig:top_predicates} (b) shows a similar distribution but for relationships involving people. Here we notice more human-centric relationships or actions such as \predicate{kissing}, \predicate{chatting with}, and \predicate{talking to}. The two distributions follow a Zipf distribution”.

\begin{figure*}[t]%
    \centering
    \subfloat[]{{\includegraphics[width=0.46\textwidth]{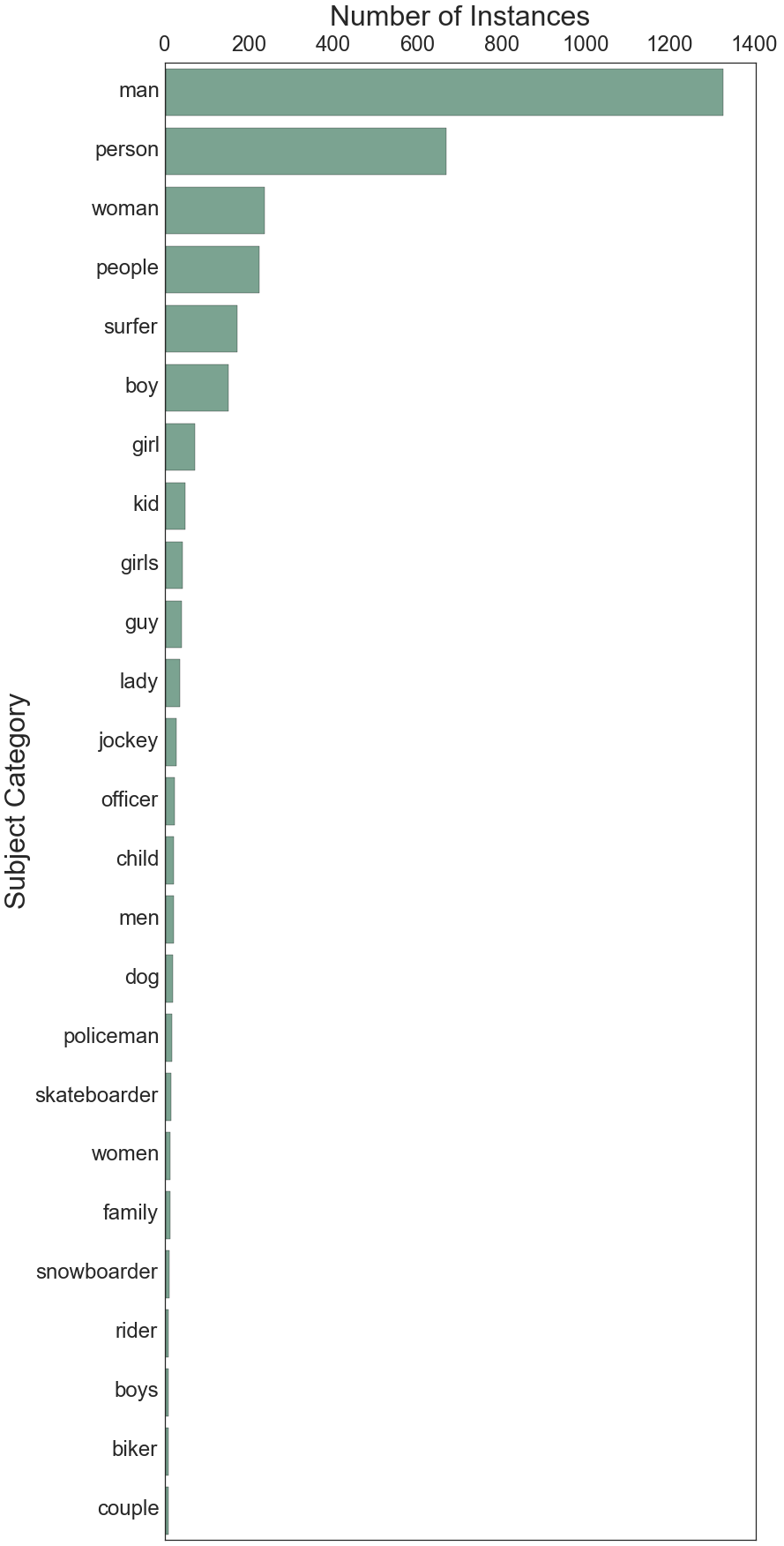} }}%
    \qquad
    \subfloat[]{{\includegraphics[width=0.46\textwidth]{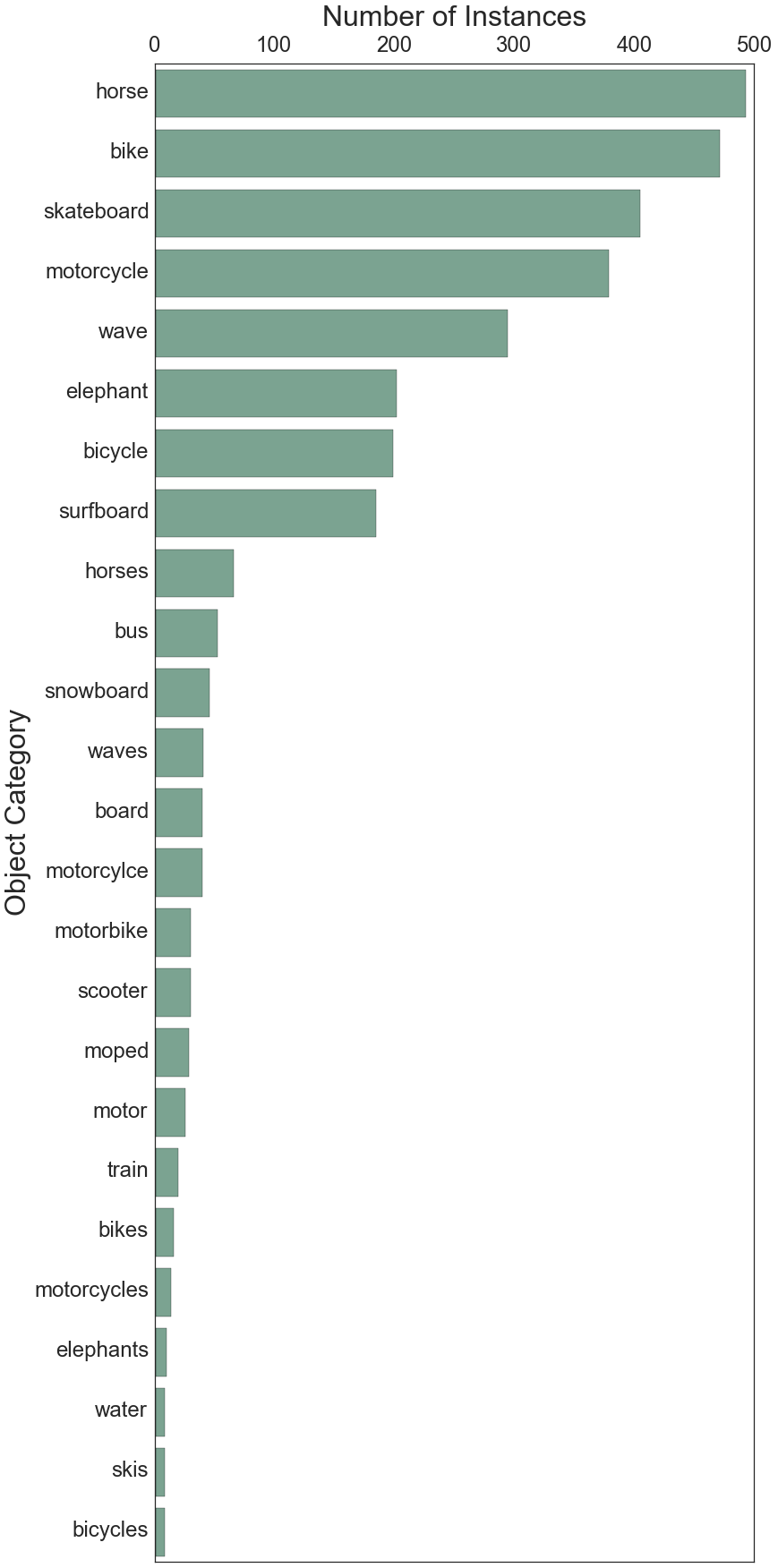} }}%
    \caption{(a) Distribution of subjects for the relationship \predicate{riding}. (b) Distribution of objects for the relationship \predicate{riding}. Subjects comprise of people-like entities like \object{person}, \object{man}, \object{policeman}, \object{boy}, and \object{skateboarder} that can ride other objects. On the other hand, objects like \object{horse}, \object{bike}, \object{elephant} and \object{motorcycle} are entities that can afford \predicate{riding}.}%
    \label{fig:riding_subjects_objects}%
\end{figure*}

\vspace{-0.15cm}
\paragraph{Understanding affordances.} Relationships allow us to also understand the affordances of objects. We show this using a specific distribution of subjects and objects involved in the relationship \predicate{riding} in Figure~\ref{fig:riding_subjects_objects}. Figure~\ref{fig:riding_subjects_objects} (a) shows the distribution for subjects while Figure~\ref{fig:riding_subjects_objects} (b) shows a similar distribution for objects. Comparing the two distributions, we find clear patterns of people-like subject entities such as \object{person}, \object{man}, \object{policeman}, \object{boy}, and \object{skateboarder} that can ride other objects; the other distribution contains objects that afford \predicate{riding}, such as \object{horse}, \object{bike}, \object{elephant}, \object{motorcycle}, and \object{skateboard}. We can also learn specific common-sense knowledge, like that \object{skateboarders} only ride \object{skateboards} and only \object{surfers} ride \object{waves} or \object{surfboards}.

\vspace{-0.15cm}
\paragraph{Related work comparison.} It is also worth mentioning in this section some prior work on relationships. The concept of visual relationships has already been explored in Visual Phrases~\cite{sadeghi2011recognition}, who introduced a dataset of $17$ such relationships such as \relationship{person}{next\_to}{bike} and \relationship{person}{riding}{horse}. However, their dataset is limited to just these $17$ relationships. Similarly, the MS-COCO-a dataset~\cite{2015arXiv150602203R} introduced $140$ actions that humans performed in MS-COCO's dataset~\cite{lin2014microsoft}. However, their dataset is limited to just actions, while our relationships are more general and numerous, with over $13$K unique relationships. Finally, VisKE~\cite{sadeghi2015viske} introduced $6500$ relationships, but in a much smaller dataset of images than Visual Genome.

\begin{figure*}[t]
\begin{center}
    \iftoggle{smallfigs}{
        \includegraphics[width=1.0\textwidth]{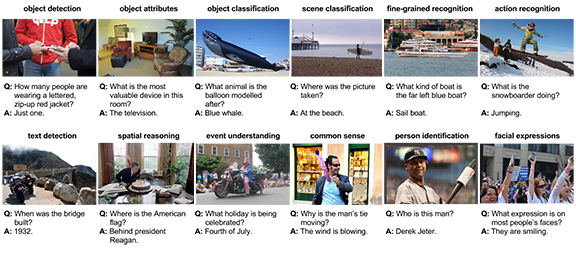}
    }{
        \includegraphics[width=1.0\textwidth]{png_graphics/visual6w_examples.png}
    }
    \caption{Example QA pairs in the Visual Genome dataset. Our QA pairs cover a spectrum of visual tasks from recognition to high-level reasoning.}
\label{fig:visual6w_examples}
\vspace{-1em}
\end{center}
\end{figure*}

\begin{table}%
\centering
\begin{tabular}{lccc}
 & Objects & Attributes & Relationships \\ 
 \midrule
Region Graph & 0.43 & 0.41 & 0.45 \\
Scene Graph & 21.26 & 16.21 & 18.67 \\
\bottomrule
\end{tabular}
\caption{The average number of objects, attributes, and relationships per region graph and per scene graph.}
\label{tab:graph_statistics}
\end{table}

\subsection{Region and Scene Graph Statistics}\label{sec:graph_stats}
We introduce in this paper the largest dataset of scene graphs to date. We use these graph representations of images as a deeper understanding of the visual world. In this section, we analyze the properties of these representations, both at the region level through region graphs and at the image level through scene graphs. We also briefly explore other datasets with scene graphs and provide aggregate statistics on our entire dataset.

Scene graphs by asking humans to write triples about an image~\cite{Johnson2015CVPR}. However, unlike them, we collect graphs at a much more fine-grained level, the region graph. We obtained our graphs by asking workers to create them from the descriptions we collected from our regions. Therefore, we end up with multiple graphs for an image, one for every region description. Together, we can combine all the individual region graphs to aggregate a scene graph for an image. This scene graph is made up of all the individual region graphs. In our scene graph representation, we merge all the objects that referenced by multiple region graphs into one node in the scene graph.

Each of our images has a distribution between $40$ to $50$ region graphs per image, with an average of $42$. Each image has exactly one scene graph. Note that the number of region descriptions and the number of region graphs for an image are not the same. For example, consider the description ``it is a sunny day''. Such a description contains no objects, which are the building blocks of a region graph. Therefore, such descriptions have no region graphs associated with them.

Objects, attributes, and relationships occur as a normal distribution in our data.  Table~\ref{tab:graph_statistics} shows that in a region graph, there are an average of $0.43$ objects, $0.41$ attributes, and $0.45$ relationships. Each scene graph and consequently each image has average of $21.26$ objects, $16.21$ attributes, and $18.67$ relationships.


\subsection{Question Answering Statistics}
\label{sec:qa_stats}
We collected $1,773,258$ question answering (QA) pairs on the Visual Genome images. Each pair consists of a question and its correct answer regarding the content of an image. On average, every image has $17$ QA pairs. Rather than collecting unconstrained QA pairs as previous work has done~\cite{antol2015vqa,gao2015you,malinowski2014multi}, each question in Visual Genome starts with one of the six Ws -- what, where, when, who, why, and how. There are two major benefits to focusing on six types of questions. First, they offer a considerable coverage of question types, ranging from basic perceptual tasks (e.g.\ recognizing objects and scenes) to complex common sense reasoning (e.g.\ inferring motivations of people and causality of events). Second, these categories present a natural and consistent stratification of task difficulty, indicated by the baseline performance in Section~\ref{sec:answer_generation}. For instance, \emph{why} questions that involve complex reasoning lead to the poorest performance ($3.4\%$ top-100 accuracy) of the six categories. This enables us to obtain a better understanding of the strengths and weaknesses of today's computer vision models, which sheds light on future directions in which to proceed.

We now analyze the diversity and quality of our questions and answers. Our goal is to construct a large-scale visual question answering dataset that covers a diverse range of question types, from basic cognition tasks to complex reasoning tasks. We demonstrate the richness and diversity of our QA pairs by examining the distributions of questions and answers in Figure~\ref{fig:visual6w_examples}.

\begin{figure}[t!]
\begin{center}
\includegraphics[width=1.0\linewidth]{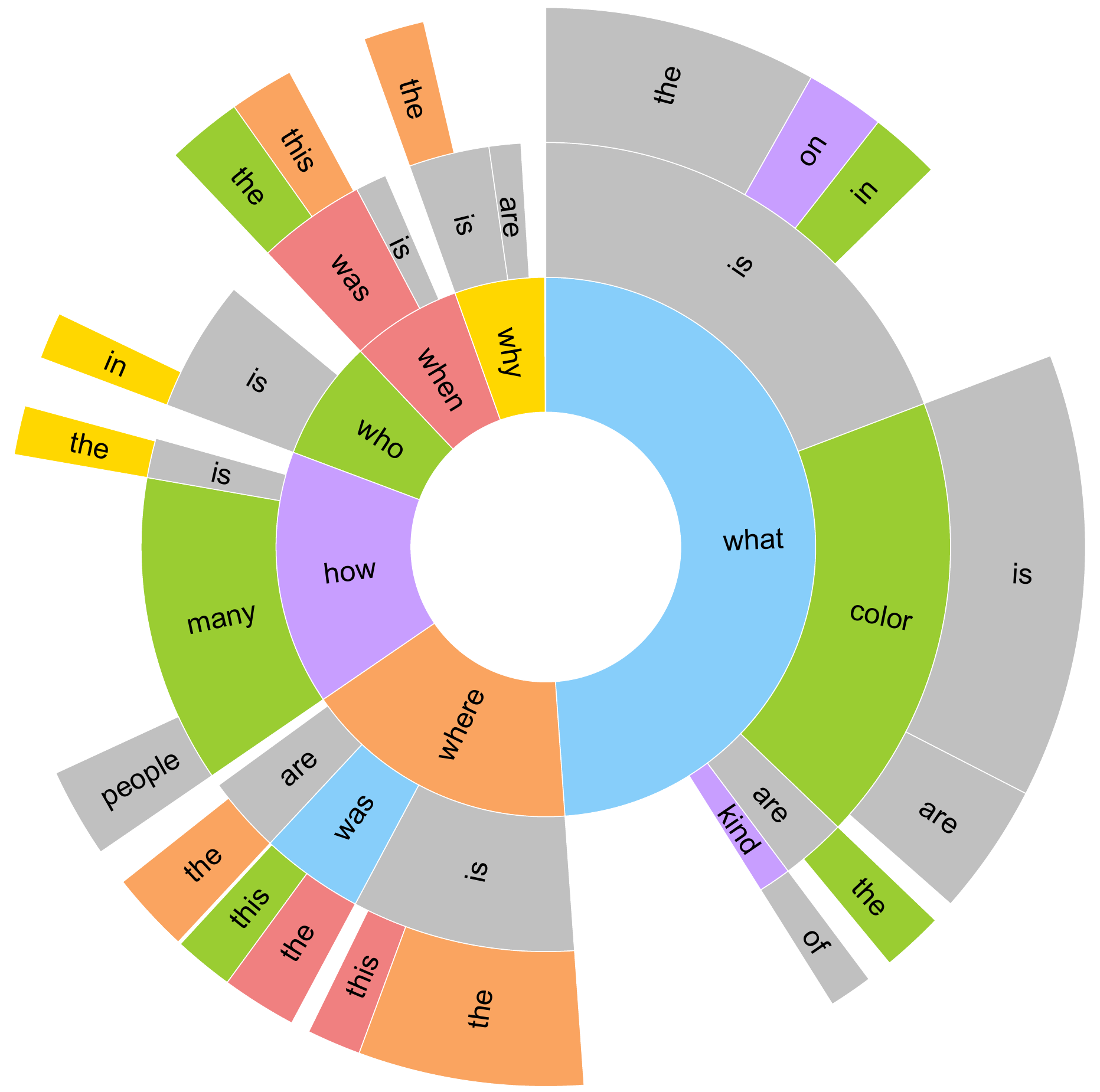}
\caption{Distribution of question types by starting words. This figure shows the distribution of the questions by their first three words. The angles of the regions are proportional to the number of pairs from the corresponding categories. We can see that ``what'' questions are the largest category with nearly half of the QA pairs.}
\label{fig:visual6w-first3-sunburst}
\end{center}
\end{figure}

\paragraph{Question type distributions.} The questions naturally fall into the 6W categories via their interrogative words. Inside each of the categories, the second and following words categorize the questions with increasing granularity. Inspired by VQA~\cite{antol2015vqa}, we show the distributions of the questions by their first three words in Figure~\ref{fig:visual6w-first3-sunburst}. We can see that ``what'' is the most common of the six categories. A notable difference between our question distribution and VQA's is that we focus on ensuring that all 7 question categories are adequately represented, while in VQA, $32.37\%$ of the questions are yes/no binary questions. As a result, a trivial model can achieve a reasonable performance by just predicting ``yes'' or ``no'' as answers. We encourage more difficult QA pairs by ruling out binary questions.

\begin{figure}[t]
\centering
\includegraphics[width=1.0\linewidth]{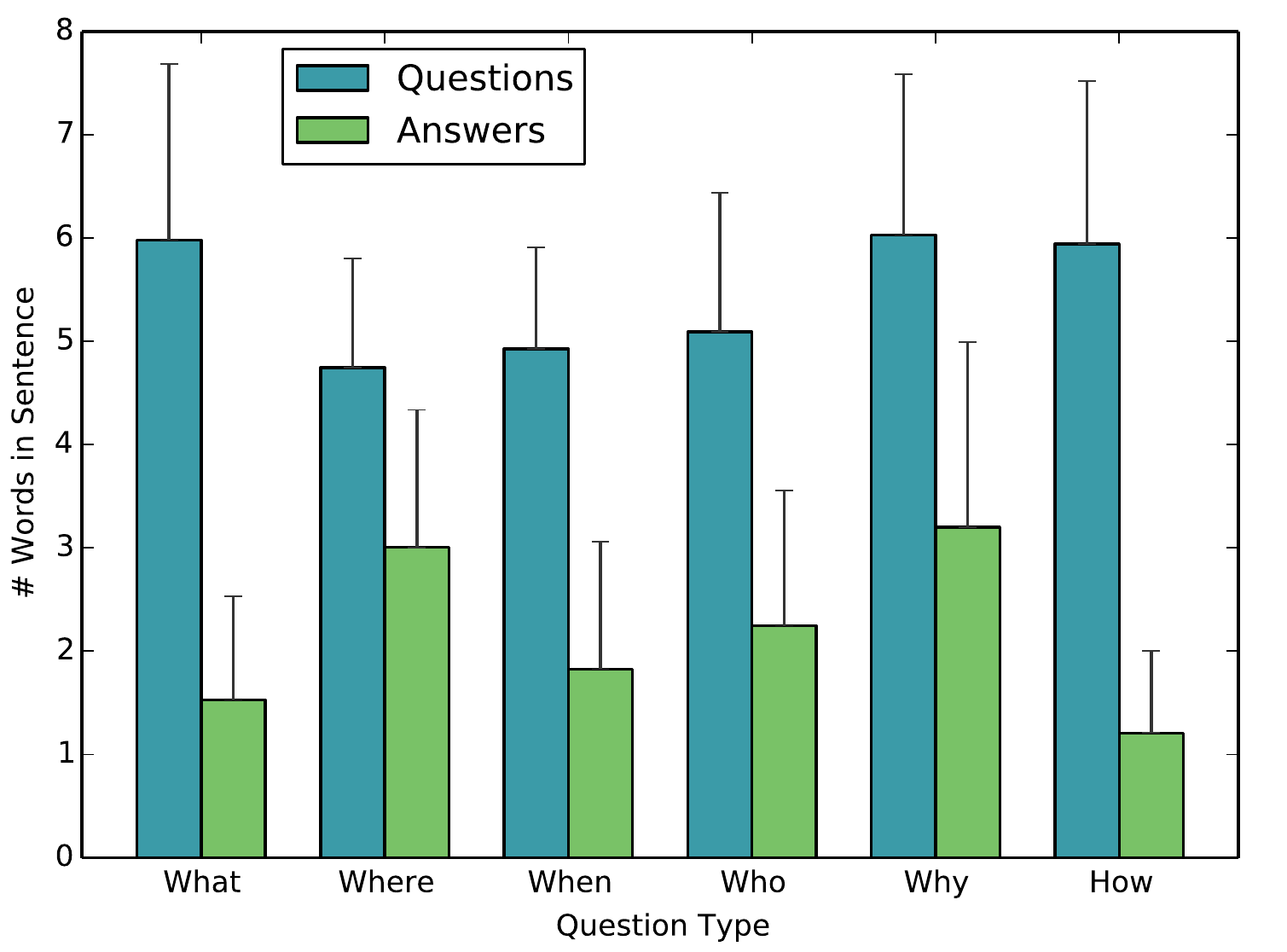}
\caption{Question and answer lengths by question type. The bars show the average question and answer lengths of each question type. The whiskers show the standard deviations. The factual questions, such as ``what'' and ``how'' questions, usually come with short answers of a single object or a number. This is only because ``how'' questions are disproportionately counting questions that start with ``how many''. Questions from the ``where'' and ``why'' categories usually have phrases and sentences as answers.}
\label{fig:visual6w-sentence_lengths_by_question_type}
\end{figure}

\paragraph{Question and answer length distributions.} We also analyze the question and answer lengths of each 6W category. Figure~\ref{fig:visual6w-sentence_lengths_by_question_type} shows the average question and answer lengths of each category. Overall, the average question and answer lengths are 5.7 and 1.8 words respectively. In contrast to the VQA dataset, where $.88\%$, $8.38\%$, and $3.25\%$ of the answers consist of one, two, or three words, our answers exhibit a long-tail distribution where $57.3\%$, $18.1\%$, and $15.7\%$ of the answers have one, two, or three words respectively. 
We avoid verbosity by instructing the workers to write answers as concisely as possible. The coverage of long answers means that many answers contain a short description that contains more details than merely an object or an attribute. It shows the richness and complexity of our visual QA tasks beyond object-centric recognition tasks. We foresee that these long-tail questions can motivate future research in common-sense reasoning and high-level image understanding.

\begin{figure*}[ht]
 \centering
    \iftoggle{smallfigs}{
        \includegraphics[width=1.0\linewidth]{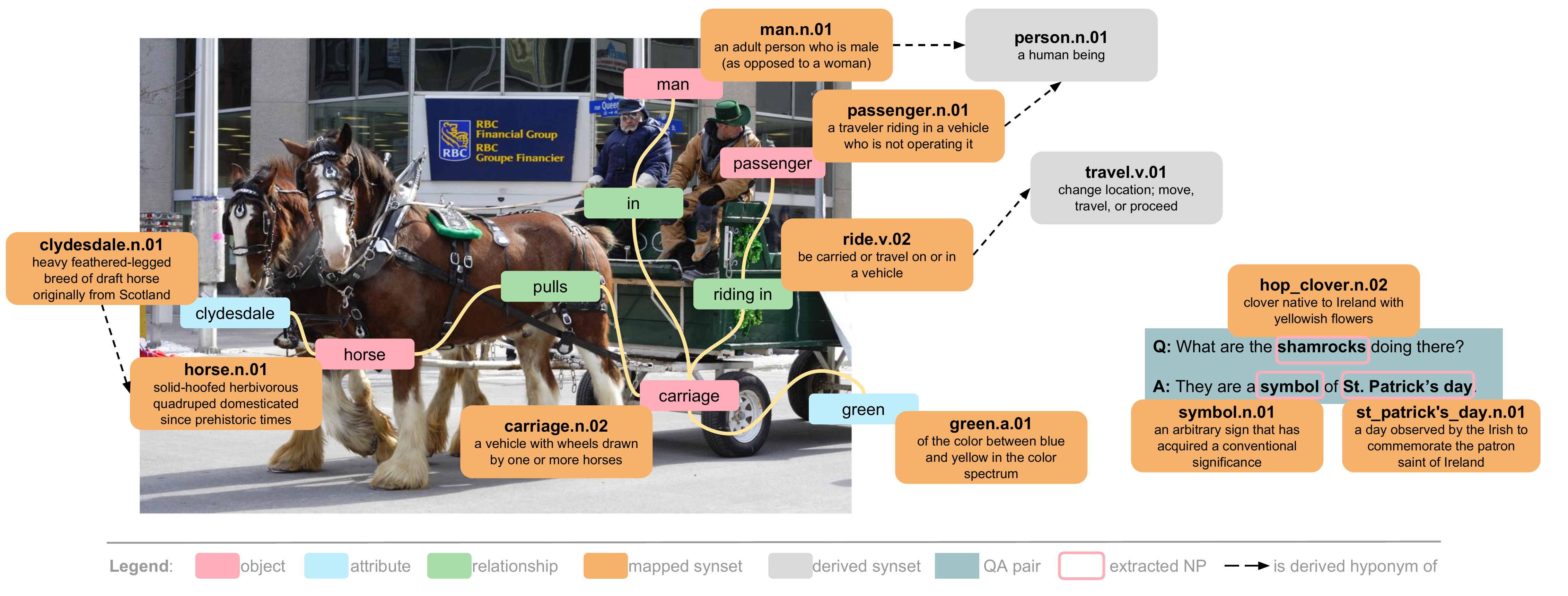}
    }{
        \includegraphics[width=1.0\linewidth]{png_graphics/canonicalization_pipeline.png}
    }
    \caption{An example image from the Visual Genome dataset with its region descriptions, QA, objects, attributes, and relationships canonicalized. The large text boxes are WordNet synsets referenced by this image. For example, the \object{carriage} is mapped to \synset{carriage.n.02: a vehicle with wheels drawn by one or more horses.} We do not show the bounding boxes for the objects in order to allow readers to see the image clearly. We also only show a subset of the scene graph for this image to avoid cluttering the figure.}
\label{fig:canonicalization_pipeline}
\end{figure*}

\subsection{Canonicalization Statistics}
\label{sec:canonicalization_stats}
In order to reduce the ambiguity in the concepts of our dataset and connect it to other resources used by the research community, we canonicalize the semantic meanings of all objects, relationships, and attributes in Visual Genome. By ``canonicalization,'' we refer to word sense disambiguation (WSD) by mapping the components in our dataset to their respective synsets in the WordNet ontology~\cite{miller1995WordNet}. This mapping reduces the noise in the concepts contained in the dataset and also facilitates the linkage between Visual Genome and other data sources such as ImageNet~\cite{deng2009imagenet}, which is built on top of the WordNet ontology.

Figure~\ref{fig:canonicalization_pipeline} shows an example image from the Visual Genome dataset with its components canonicalized. For example, \object{horse} is canonicalized as \synset{horse.n.01: solid-hoofed herbivorous quadruped domesticated since prehistoric times}. Its attribute, \attribute{clydesdale}, is canonicalized as its breed \synset{clydesdale.n.01: heavy feathered-legged breed of draft horse originally from Scotland}. We also show an example of a QA from which we extract the nouns \object{shamrocks}, \object{symbol}, and \object{St. Patrick's day}, all of which we canonicalize to WordNet as well. 

\vspace{-1em}
\paragraph{Related work.} Canonicalization, or WSD~\cite{pal2015word}, has been used in numerous applications, including machine translation, information retrieval, and information extraction~\cite{rothe2015autoextend, leacock1998using}. In English sentences, sentences like ``He scored a goal'' and ``It was his goal in life'' carry different meanings for the word ``goal.'' Understanding these differences is crucial for translating languages and for returning correct results for a query. Similarly, in Visual Genome, we ensure that all our components are canonicalized to understand how different objects are related to each other; for example, ``person'' is a hypernym of ``man'' and ``woman.'' Most past canonicalization models use precision, recall, and F1 score to evaluate on the Semeval dataset~\cite{mihalcea2004senseval}. The current state-of-the-art performance on Semeval is an F1 score of $75.8\%$~\cite{chen2014unified}. Since our canonicalization setup is different from the Semeval benchmark (we have an open vocabulary and no annotated ground truth for evaluation), our canonicalization method is not directly comparable to these existing methods. We do however, achieve a similar precision and recall score on a held-out test set described below.

\begin{figure*}[htbp]
\centering
\subfloat[]{{\includegraphics[width=0.5\linewidth]{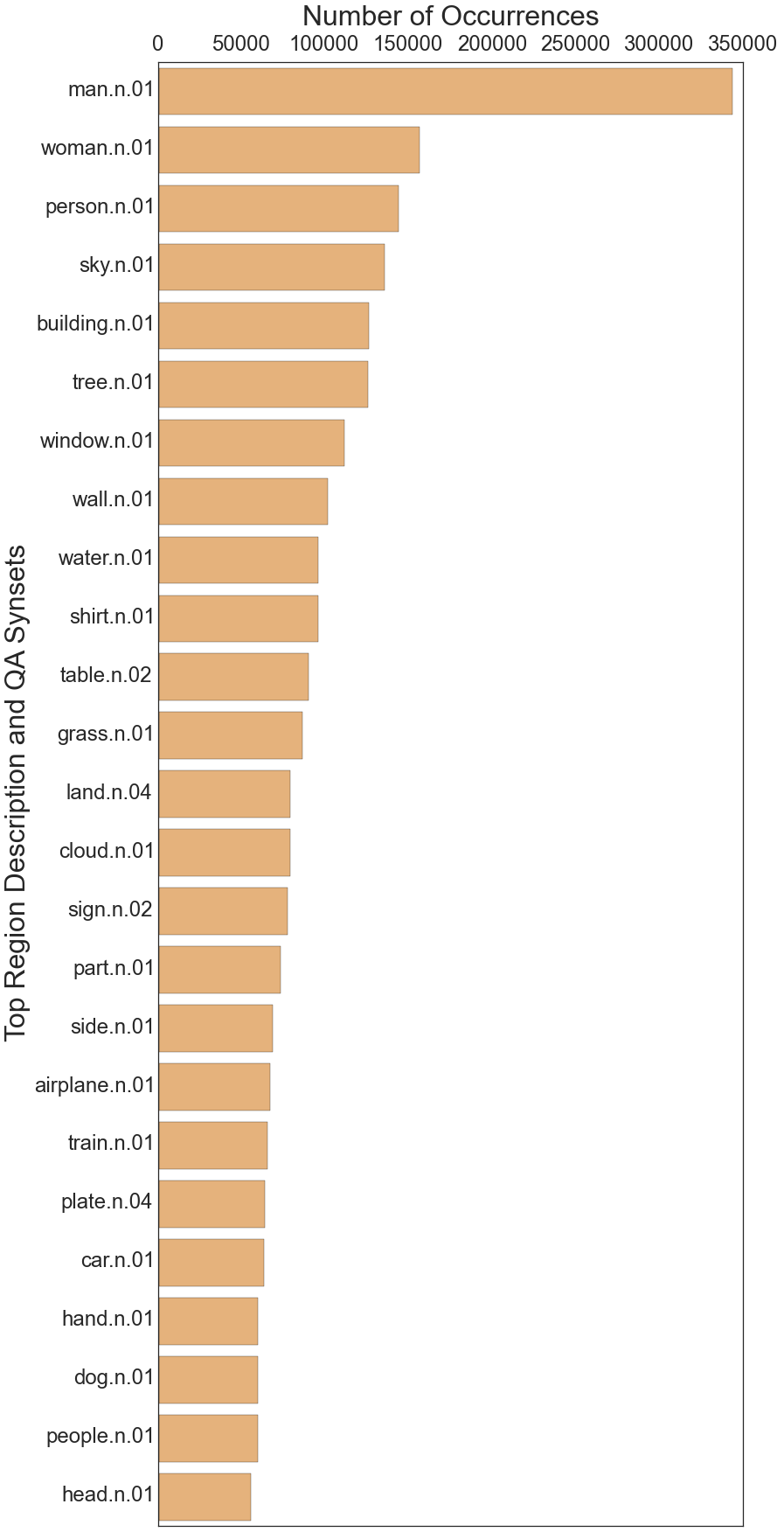}}}
\subfloat[]{{\includegraphics[width=0.5\linewidth]{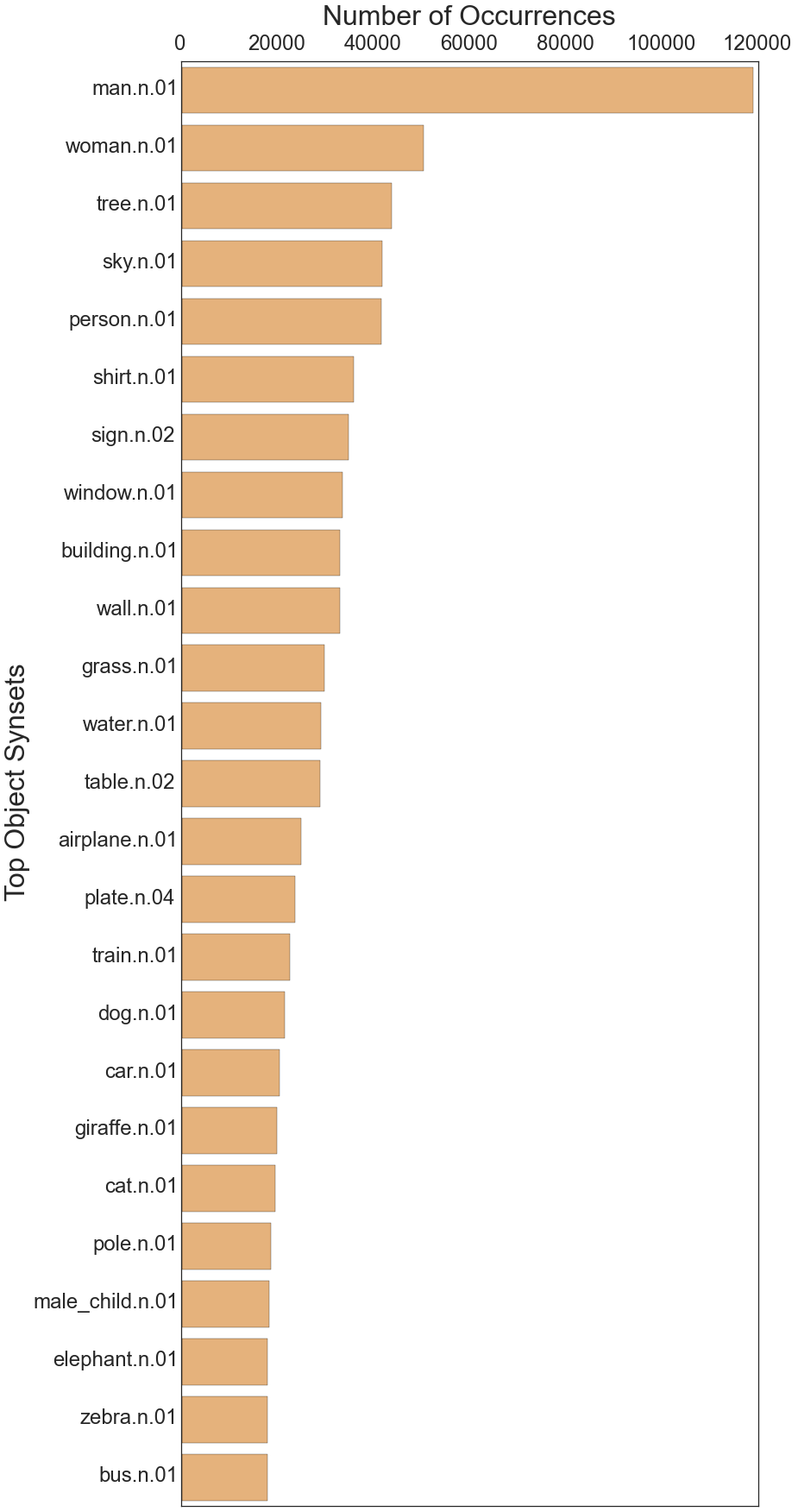}}}
\caption{Distribution of the 25 most common synsets mapped from (a) region descriptions and question answers and (b) objects.}
\label{fig:region_object_synset_distributions}
\end{figure*}

\paragraph{Region descriptions and QAs.} We canonicalize all objects mentioned in all region descriptions and QA pairs. Because objects need to be extracted from the phrase text, we use Stanford NLP tools~\cite{manning-EtAl:2014:P14-5} to extract the noun phrases in each region description and QA, resulting in $99\%$ recall of noun phrases from a subset of $200$ region descriptions we manually annotated. After obtaining the noun phrases, we map each to its most frequent matching synset (according to WordNet lexeme counts). 
This resulted in an overall mapping accuracy of $86\%$ and a recall of $98.5\%$. The most common synsets extracted from region descriptions, QAs, and objects are shown in Figure~\ref{fig:region_object_synset_distributions}.

\begin{table}[t]%
\centering
\begin{tabular}{l c c}
 & Precision & Recall\\
 \midrule
Objects & 88.0 & 98.5\\
Attributes & 85.7 & 95.9\\
Relationships & 92.9 & 88.5\\
\bottomrule
\end{tabular}
\caption{Precision, recall, and mapping accuracy percentages for object, attribute, and relationship canonicalization.}
\label{tab:canon_statistics}
\vspace{-1em}
\end{table}

\begin{figure*}[htbp]
\centering
\subfloat[]{{\includegraphics[width=0.49\linewidth]{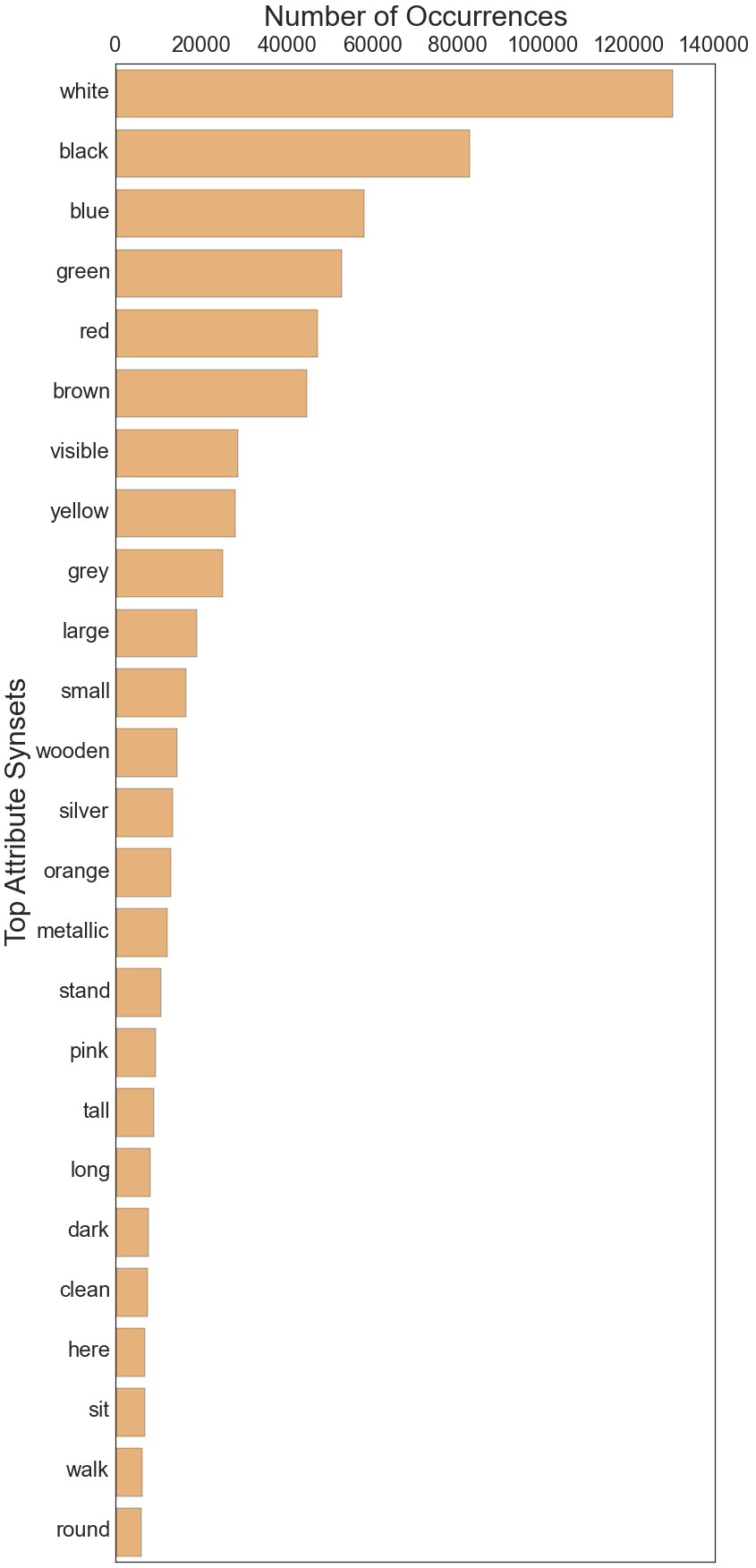}}}
\subfloat[]{{\includegraphics[width=0.51\linewidth]{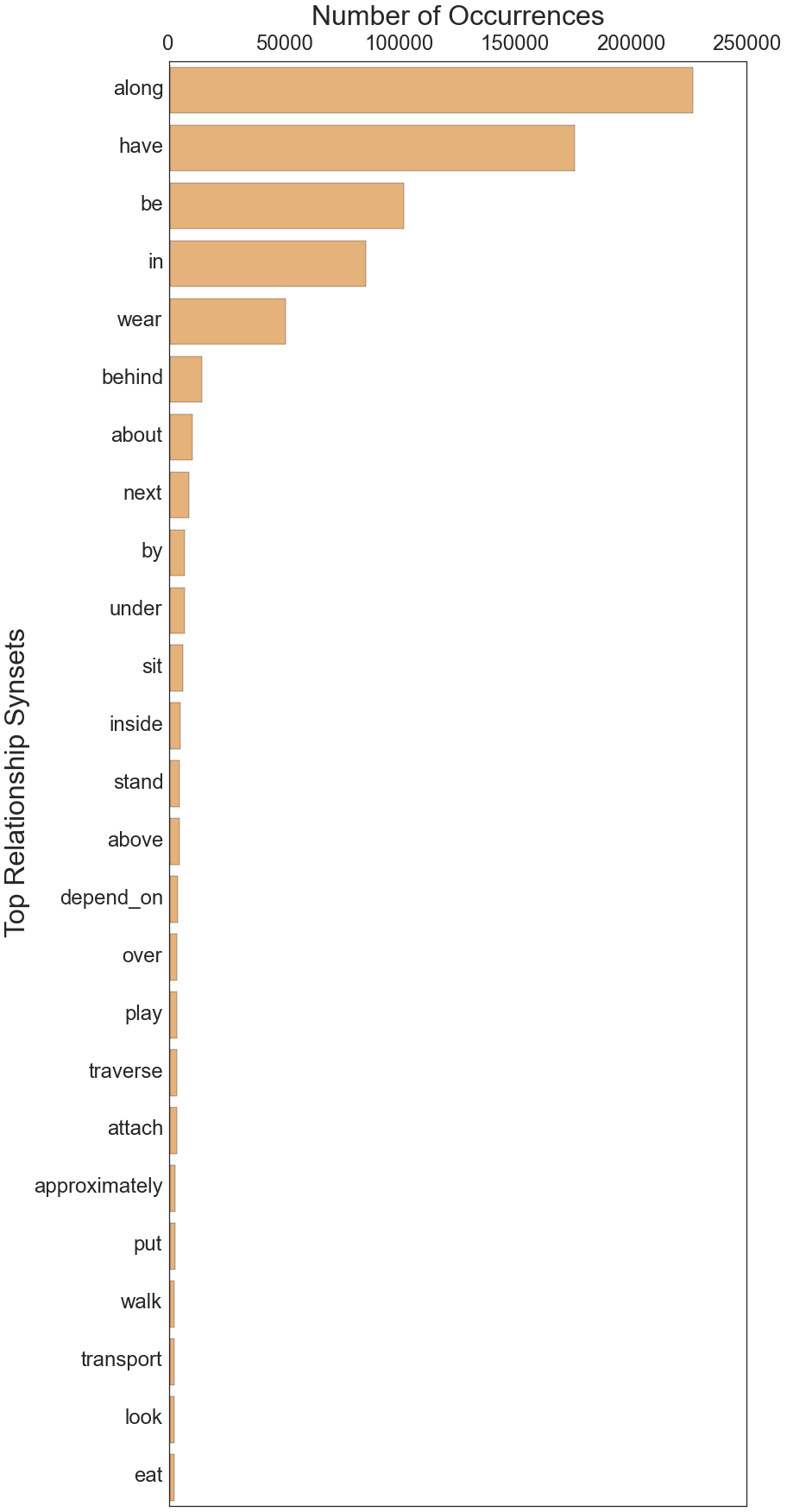}}}
\caption{Distribution of the 25 most common synsets mapped from (a) attributes and (b) relationships.}
\label{fig:attribute_relationship_synset_distributions}
\end{figure*}

\vspace{-1em}
\paragraph{Attributes.} We canonicalize attributes from the crowd-extracted attributes present in our scene graphs. The ``attribute'' designation encompasses a wide range of grammatical parts of speech. Because part-of-speech taggers rely on high-level syntax information and thus fail on the disjoint elements of our scene graphs, we normalize each attribute based on morphology alone (so-called ``stemming''). Then, as with objects, we map each attribute phrase to the most frequent matching WordNet synset. We include 15 hand-mapped rules to address common failure cases in which WordNet's frequency counts prefer abstract senses of words over the spatial senses present in visual data, e.g.\ ``short.a.01: limited in duration'' over \synset{short.a.02: lacking in length}. For verification, we randomly sample $200$ attributes, produce ground-truth mappings by hand, and compare them to the results of our algorithm. This resulted in a recall of $95.9\%$ and a mapping accuracy of $83.5\%$. The most common attribute synsets are shown in Figure~\ref{fig:attribute_relationship_synset_distributions} (a).

\paragraph{Relationships.} As with attributes, we canonicalize the relationships isolated in our scene graphs. We exclude prepositions, which are not recognized in WordNet, leaving a set primarily composed of verb relationships. Since the meanings of verbs are highly dependent upon their morphology and syntactic placement (e.g.\ passive cases, prepositional phrases), we map the structure of each relationship to the appropriate WordNet sentence frame and only consider those WordNet synsets with matching sentence frames. For each verb-synset pair, we then consider the root hypernym of that synset to reduce potential noise from WordNet's fine-grained sense distinctions. We also include 20 hand-mapped rules, again to correct for WordNet's lower representation of concrete or spatial senses; for example, the concrete \synset{hold.v.02: have or hold in one's hand or grip} is less frequent in WordNet than the abstract \synset{hold.v.01: cause to continue in a certain state}. For verification, we again randomly sample $200$ relationships and compare the results of our canonicalization against ground-truth mappings. This resulted in a recall of $88.5\%$ and a mapping accuracy of $77.6\%$. While several datasets, such as VerbNet~\cite{schuler2005verbnet} and FrameNet~\cite{baker1998framenet}, include semantic restrictions or frames to improve classification, there is no comprehensive method of mapping to those restrictions or frames. The most common relationship synsets are shown in Figure~\ref{fig:attribute_relationship_synset_distributions} (b).

\section{Experiments}
\label{sec:experiments}

Thus far, we have presented the Visual Genome dataset and analyzed its individual components. With such rich information provided, numerous perceptual and cognitive tasks can be tackled. In this section, we aim to provide baseline experimental results using components of Visual Genome that have not been extensively studied.
Object detection is already a well-studied problem~\cite{everingham2010pascal, girshick2014rich, sermanet2013overfeat, girshick2015fast, ren2015faster}. Similarly, region graphs and scene graphs have been shown to improve semantic image retrieval~\cite{Johnson2015CVPR, schustergenerating}. We therefore focus on the remaining components, i.e. \textit{attributes}, \textit{relationships}, \textit{region descriptions}, and \textit{question answer pairs}.

In Section~\ref{sec:attribute_classification}, we present results for two experiments on attribute prediction. In the first, we treat attributes independently from objects and train a classifier for each attribute, i.e.\ a classifier for \attribute{red} or a classifier for \attribute{old}, as in~\cite{malisiewicz2008recognition, varma2005statistical, ferrari2007learning, farhadi2009describing, Johnson2015CVPR}. In the second experiment, we learn object and attribute classifiers \emph{jointly} and predict object-attribute pairs (e.g.\ predicting that an \object{apple} is \attribute{red}), as in \cite{sadeghi2011recognition}.

In Section~\ref{sec:relationship_classification}, we present two experiments on relationship prediction. In the first, we aim to predict the predicate between two objects, e.g.\ predicting the predicate \predicate{kicking} or \predicate{wearing} between two objects. This experiment is synonymous with existing work in action recognition~\cite{gupta2009observing, ramanathan2015learning}. In another experiment, we study relationships by classifying jointly the objects and the predicate (e.g.\ pre\-dicting \relationship{man}{kicking}{ball}); we show that this is a very difficult task due to the high variability in the appearance of a relationship (e.g.\ the \object{ball} might be on the ground or in mid-air above the \object{man}). These experiment are generalizations of tasks that study spatial relationships between objects and ones that jointly reason about the interaction of humans with objects~\cite{yao2010modeling, prest2012weakly}.

In Section~\ref{sec:description_generation} we present results for region captioning. This task is closely related to image captioning~\cite{chen2015microsoft}; however, results from the two are not directly comparable, as region descriptions are short, incomplete sentences. We train one of the top 16 state-of-the-art image caption generator~\cite{karpathy2014deep} on (1) our dataset to generate region descriptions and on (2) Flickr30K~\cite{young2014image} to generate sentence descriptions. To compare results between the two training approaches, we use simple templates to convert region descriptions into complete sentences. For a more robust evaluation, we validate the descriptions we generate using human judgment.

Finally, in Section~\ref{sec:answer_generation}, we experiment on visual question answering, i.e.\ given an image and a question, we attempt to provide an answer for the question. We report results on the retrieval of the correct answer from a list of existing answers.

\subsection{Attribute Prediction}
\label{sec:attribute_classification}

\begin{figure*}[t]
    \centering
    \iftoggle{smallfigs}{
        \subfloat[]{{\includegraphics[width=0.46\textwidth]{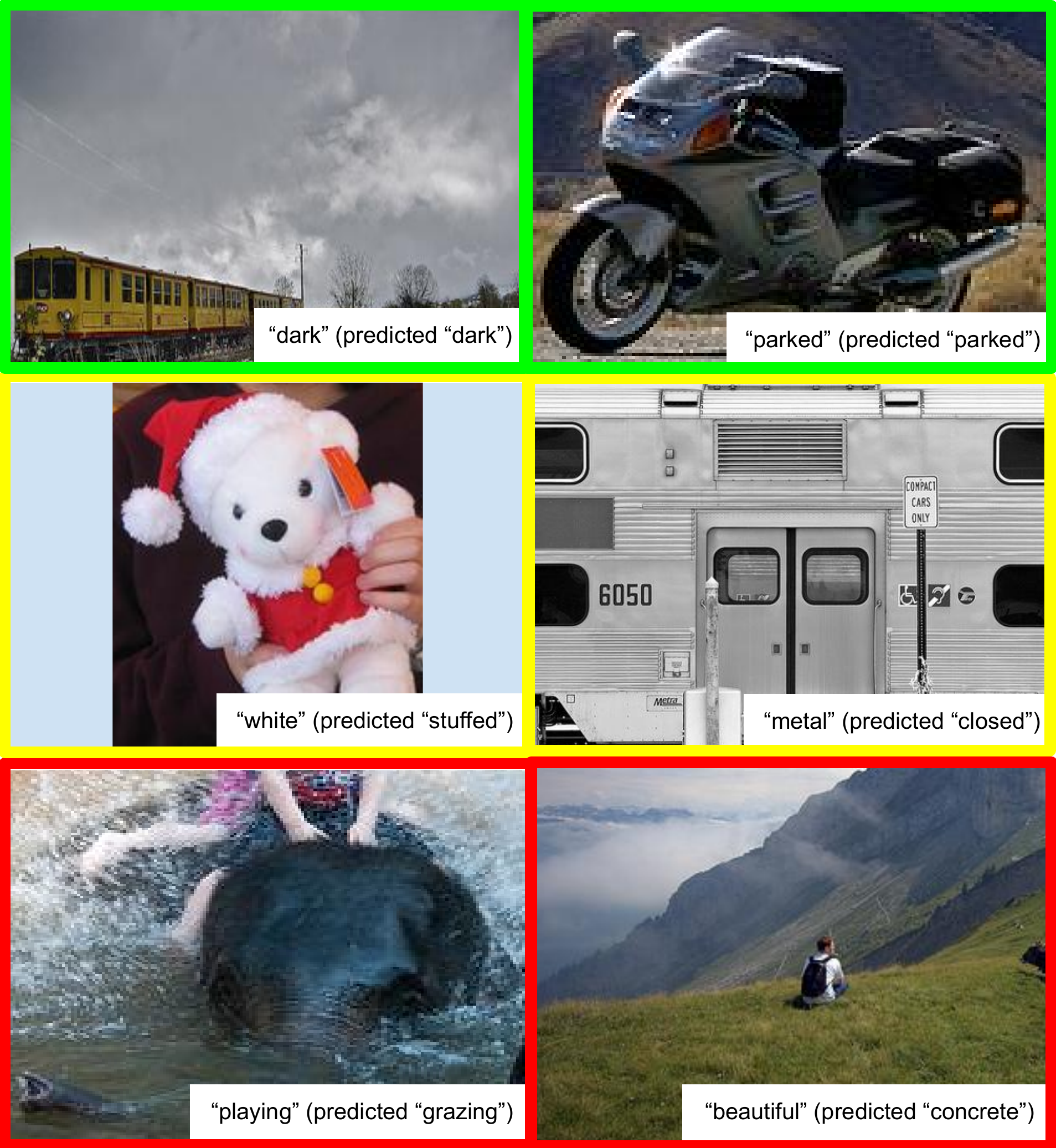} }}
    }{
        \subfloat[]{{\includegraphics[width=0.46\textwidth]{png_graphics/attribute_experiment_examples.png} }}    
    }
    \qquad
    \iftoggle{smallfigs}{
        \subfloat[]{{\includegraphics[width=0.46\textwidth]{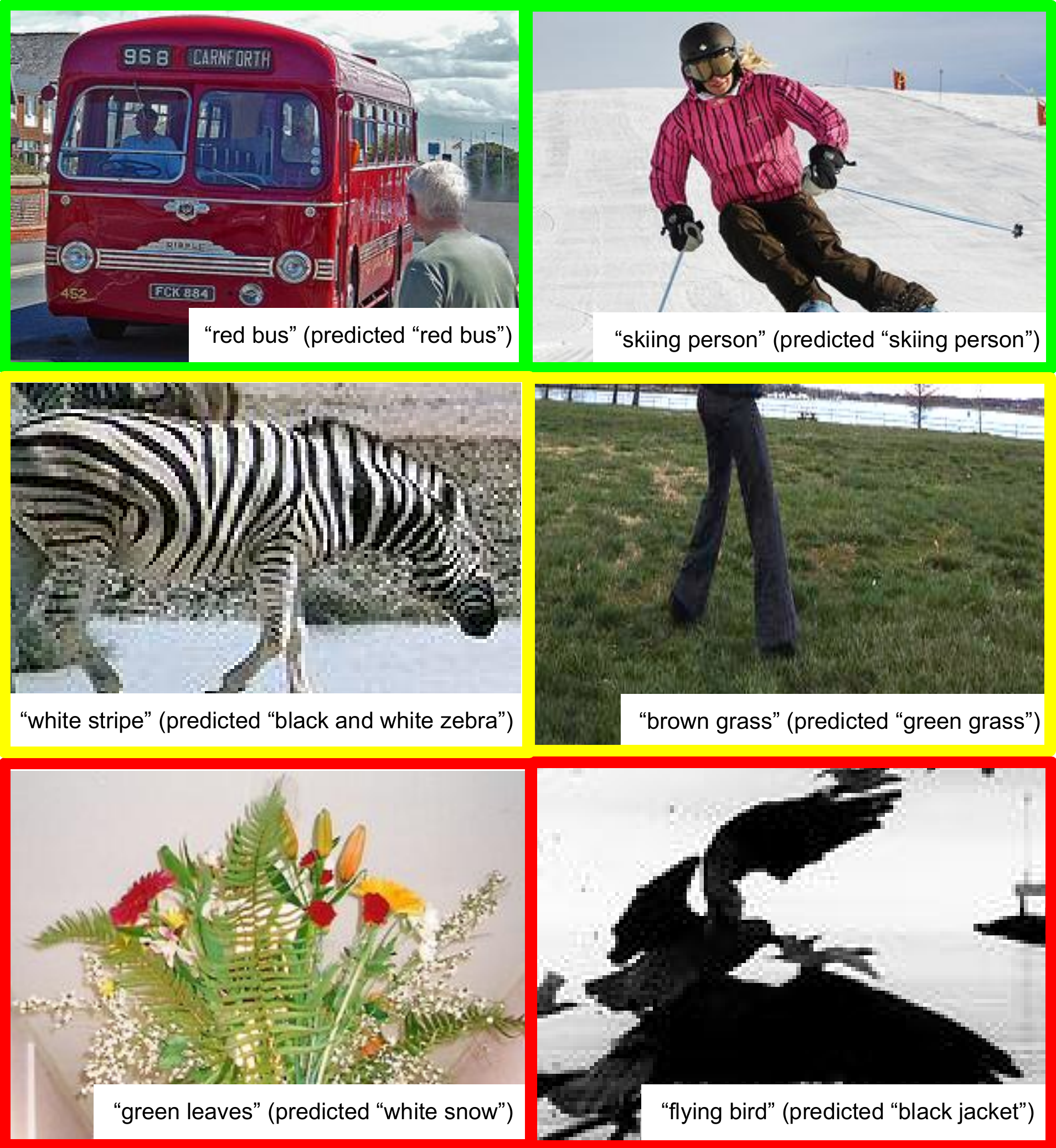}}}
    }{
        \subfloat[]{{\includegraphics[width=0.46\textwidth]{png_graphics/attribute_pair_experiment_examples.png}}}
    }
    \caption{(a) Example predictions from the attribute prediction experiment. Attributes in the first row are predicted correctly, those in the second row differ from the ground truth but still correctly classify an attribute in the image, and those in the third row are classified incorrectly. The model tends to associate objects with attributes (e.g. \object{elephant} with \object{grazing}). (b) Example predictions from the joint object-attribute prediction experiment.}
    \label{fig:attribute_experiment}
\end{figure*}



Attributes are becoming increasingly important in the field of computer vision, as they offer higher-level semantic cues for various problems and lead to a deeper understanding of images. We can express a wide variety of properties through attributes, such as form (\attribute{sliced}), function (\attribute{decorative}), sentiment (\attribute{angry}), and even intention (\attribute{helping}). Distinguishing between similar objects~\cite{isola2015discovering} leads to finer-grained classification, while describing a previously unseen class through attributes shared with known classes can enable ``zero-shot'' learning~\cite{farhadi2009describing, lampert2009learning}. Visual Genome is the largest dataset of attributes, with $18$ attributes per image for a total of $1.8$ million attributes.

\paragraph{Setup.} 

For both experiments, we focus on the $100$ most common attributes in our dataset. We only use objects that occur at least $100$ times and are associated with one of the $100$ attributes in at least one image. For both experiments, we follow a similar data pre-processing pipeline. First, we lowercase, lemmatize, and strip excess whitespace from all attributes. Since the number of examples per attribute class varies, we randomly sample $500$ attributes from each category (if fewer than $500$ are in the class, we take all of them). 


We end up with around $50,000$ attribute instances and $43,000$ object-attribute pair instances in total. We use $80\%$ of the images for training and $10\%$ each for validation and testing. Because each image has about the same number of examples, this results in an approximately $80\%$-$10\%$-$10\%$ split over the attributes themselves. The input data for this experiment is the cropped bounding box of the object associated with each attribute.

We train an attribute predictor by using features learned from a convolutional neural network. Specifically, we fine-tune a 16-layer VGG network~\cite{simonyan2014very} for both of these experiments using the $50,000$ attribute and $43,000$ object-attribute pair instances respectively. We modify the network so that the learning rate of the final fully-connected layer is 10 times that of the other layers, as this improves convergence time. We use a base learning rate of 0.001, which we scale by 0.1 every $200$ iterations, and momentum and weight decays of $0.9$ and $0.0005$ respectively. We use the fine-tuned features from the network and train $100$ individual SVMs~\cite{hearst1998support} to predict each attribute. We output multiple attributes for each bounding box input. For the second experiment, we also output the object class. 

\paragraph{Results.} Table~\ref{tab:attribute_experiment_results} shows results for both experiments. For the first experiment on attribute prediction, we converge after around $700$ iterations with $18.97\%$ top-one accuracy and $43.11\%$ top-five accuracy. Thus, attributes (like objects) are visually distinguishable from each other. For the second experiment where we also predict the object class, we converge after around $400$ iterations with $43.17\%$ top-one accuracy and $71.97\%$ top-five accuracy. Predicting objects jointly with attributes increases the top-one accuracy from $18.97\%$ to $43.17\%$. This implies that some attributes occur exclusively with a small number of objects. Additionally, by jointly learning attributes with objects, we increase the inter-class variance, making the classification process an easier task.

Figure~\ref{fig:attribute_experiment} (a) shows example predictions for the first attribute prediction experiment. In general, the model is good at associating objects with their most salient attributes, for example, \object{animal} with \attribute{stuffed} and \object{elephant} with \attribute{grazing}. However, there is some difference between the user-provided result and the correct ground truth, so the model incorrectly classifies some correct predictions. For example, the \attribute{white} stuffed animal is correct but evaluated as incorrect. 

Figure~\ref{fig:attribute_experiment} (b) shows example predictions for the second experiment in which we also predict the object. While the results in the second row might be considered correct, to keep a consistent evaluation, we mark them as incorrect. For example, the predicted ``green grass'' might be considered subjectively correct even though it is annotated as ``brown grass''. For cases where the objects are not clearly visible but are abstract outlines, our model is unable to predict attributes or objects accurately. For example, it thinks that the ``flying bird'' is actually a ``black jacket''.

The attribute clique graphs in Section~\ref{sec:attribute_stats} clearly show that learning attributes can help us identify types of objects. This experiment strengthens that insight. We learn that studying attributes together with objects can improve attribute prediction.

\begin{table}[t]
\centering
\begin{tabular}{lccc}
 & Top-1 Accuracy & Top-5 Accuracy \\ 
 \midrule
Attribute & 18.97\% & 43.11\% \\
Object-Attribute & 43.17\% & 71.97\% \\
\bottomrule
\end{tabular}
\caption{(First row) Results for the attribute prediction task where we only predict attributes for a given image crop. (Second row) Attribute-object prediction experiment where we predict both the attributes as well as the object from a given crop of the image.}
\label{tab:attribute_experiment_results}
\end{table}

\subsection{Relationship Prediction}
\label{sec:relationship_classification}

\begin{figure*}[t]
    \centering
    \iftoggle{smallfigs}{
        \subfloat[]{{\includegraphics[width=0.46\textwidth]{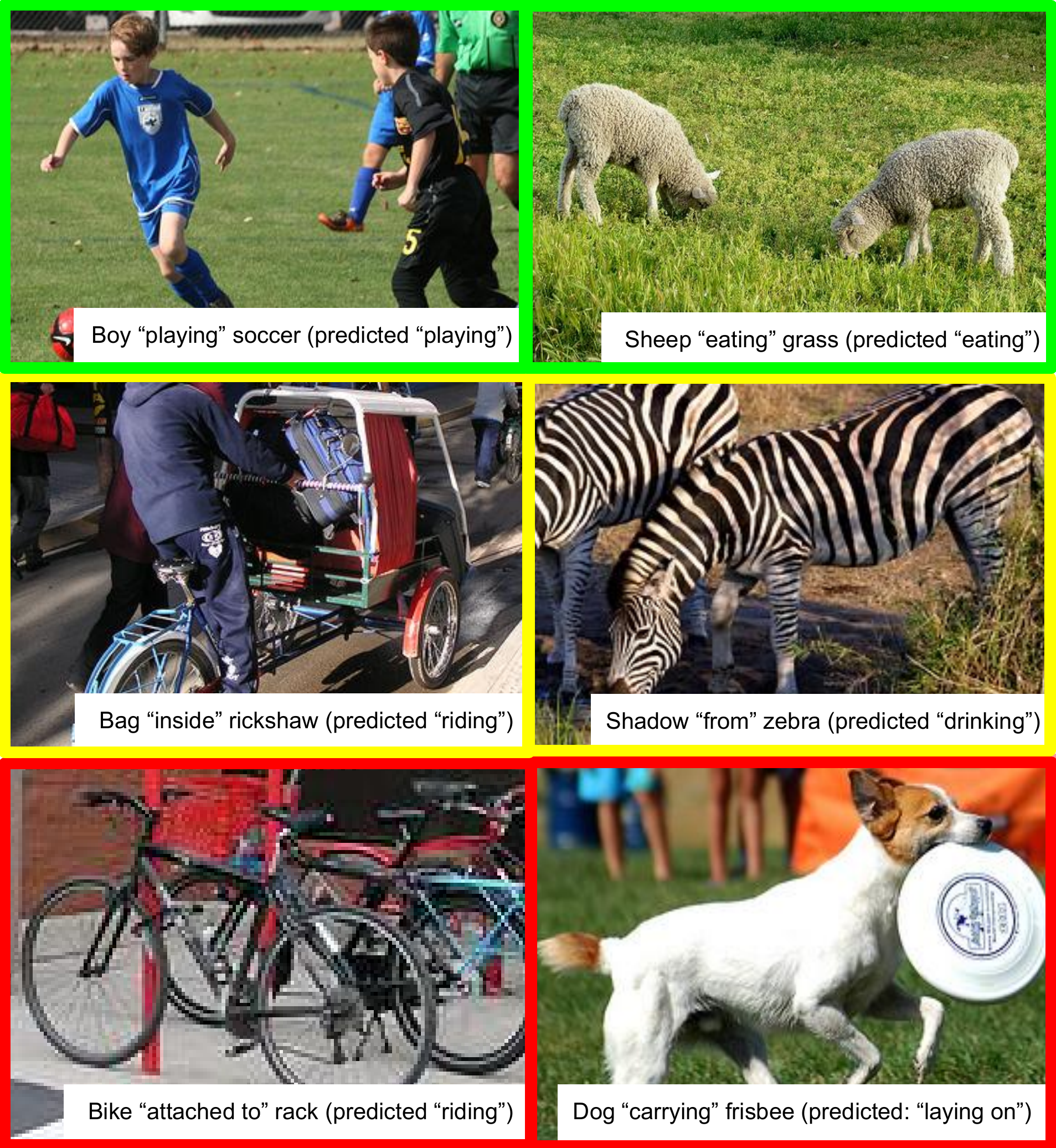}}}
    }{
        \subfloat[]{{\includegraphics[width=0.46\textwidth]{png_graphics/relationship_experiment_examples.png}}}
    }
    \qquad
    \iftoggle{smallfigs}{
        \subfloat[]{{\includegraphics[width=0.46\textwidth]{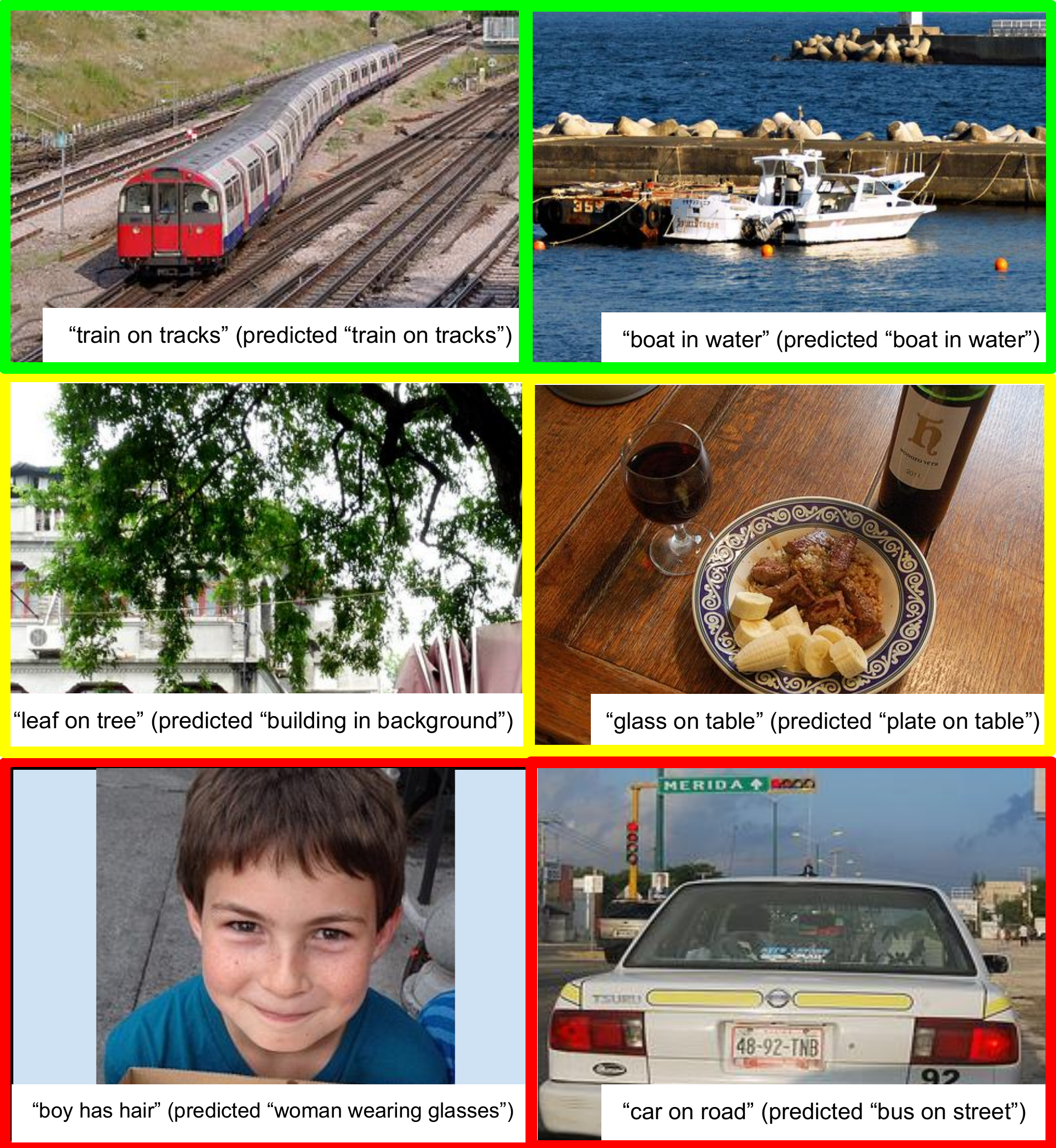}}}
    }{
        \subfloat[]{{\includegraphics[width=0.46\textwidth]{png_graphics/relationship_triple_experiment_examples.png}}}
    }
    \caption{(a) Example predictions from the relationship prediction experiment. Relationships in the first row are predicted correctly, those in the second row differ from the ground truth but still correctly classify a relationship in the image, and those in the third row are classified incorrectly. The model learns to associate animals leaning towards the ground as \predicate{eating} or \predicate{drinking} and bikes with \predicate{riding}. (b) Example predictions from the relationship-objects prediction experiment. The figure is organized in the same way as Figure (a). The model is able to predict the salient features of the image but fails to distinguish between different objects (e.g. \object{boy} and \object{woman} and \object{car} and \object{bus} in the bottom row).}
    \label{fig:relationship_experiment}
    \vspace{-1.0em}
\end{figure*}

While objects are the core building blocks of an image, relationships put them in context. These relationships help distinguish between images that contain the same objects but have different holistic interpretations. For example, an image of ``a man riding a bike'' and ``a man falling off a bike'' both contain \object{man} and \object{bike}, but the relationship (\predicate{riding} vs. \predicate{falling\_off}) changes how we perceive both situations. Visual Genome is the largest known dataset of relationships, with a total of $1.8$ million relationships and an average of $18$ relationships per image.


\paragraph{Setup.} The setups of both experiments are similar to those of the experiments we performed on attributes. We again focus on the top $100$ most frequent relationships. We lowercase, lemmatize, and strip excess whitespace from all relationships. We end up with around $34,000$ relationships and $27,000$ subject-relationship-object triples for training, validation, and testing. The input data to the experiment is the image region containing the union of the bounding boxes of the subject and object (essentially, the bounding box containing the two object boxes). We fine-tune a 16-layer VGG network~\cite{simonyan2014very} with the same learning rates mentioned in Section~\ref{sec:attribute_classification}. 

\paragraph{Results.} Overall, we find that relationships are not visually distinct enough for our discriminative model to learn effectively. Table~\ref{tab:relationship_experiment_results} shows results for both experiments. For relationship classification, we converge after around $800$ iterations with $8.74\%$ top-one accuracy and $29.69\%$ top-five accuracy. Unlike attribute prediction, the accuracy results for relationships are much lower because of the high intra-class variability of most relationships. 
For the second experiment jointly predicting the relationship and its two object classes, we converge after around $450$ iterations with $25.83\%$ top-one accuracy and $65.57\%$ top-five accuracy. We notice that object classification aids relationship prediction. Some relationships occur with some objects and never others; for example, the relationship \predicate{drive} only occurs with the object \object{person} and never with any other objects (\object{dog}, \object{chair}, etc.).

Figure~\ref{fig:relationship_experiment} (a) shows example predictions for the relationship classification experiment. In general, the model associates object categories with certain relationships (e.g.\ animals with \predicate{eating} or \predicate{drinking}, bikes with \predicate{riding}, and kids with \predicate{playing}).

Figure~\ref{fig:relationship_experiment} (b), structured as in Figure~\ref{fig:relationship_experiment} (a), shows example predictions for the joint prediction of relationships with its objects. The model is able to predict the salient features of the image (e.g. ``boat in water'') but fails to distinguish between different objects (e.g. \object{boy} vs. \object{woman} and \object{car} vs. \object{bus} in the bottom row).

\begin{table}[t]
\centering
\begin{tabular}{lccc}
 & Top-1 Accuracy & Top-5 Accuracy \\ 
 \midrule
Relationship & 8.74\% & 26.69\% \\
Sub./Rel./Obj. & 25.83\% & 65.57\% \\
\bottomrule
\end{tabular}
\caption{Results for relationship classification (first row) and joint classification (second row) experiments.}
\label{tab:relationship_experiment_results}
\end{table}

\subsection{Generating Region Descriptions}
\label{sec:description_generation}

\begin{figure}[htbp]
\centering
\iftoggle{smallfigs}{
    \includegraphics[width=0.46\textwidth]{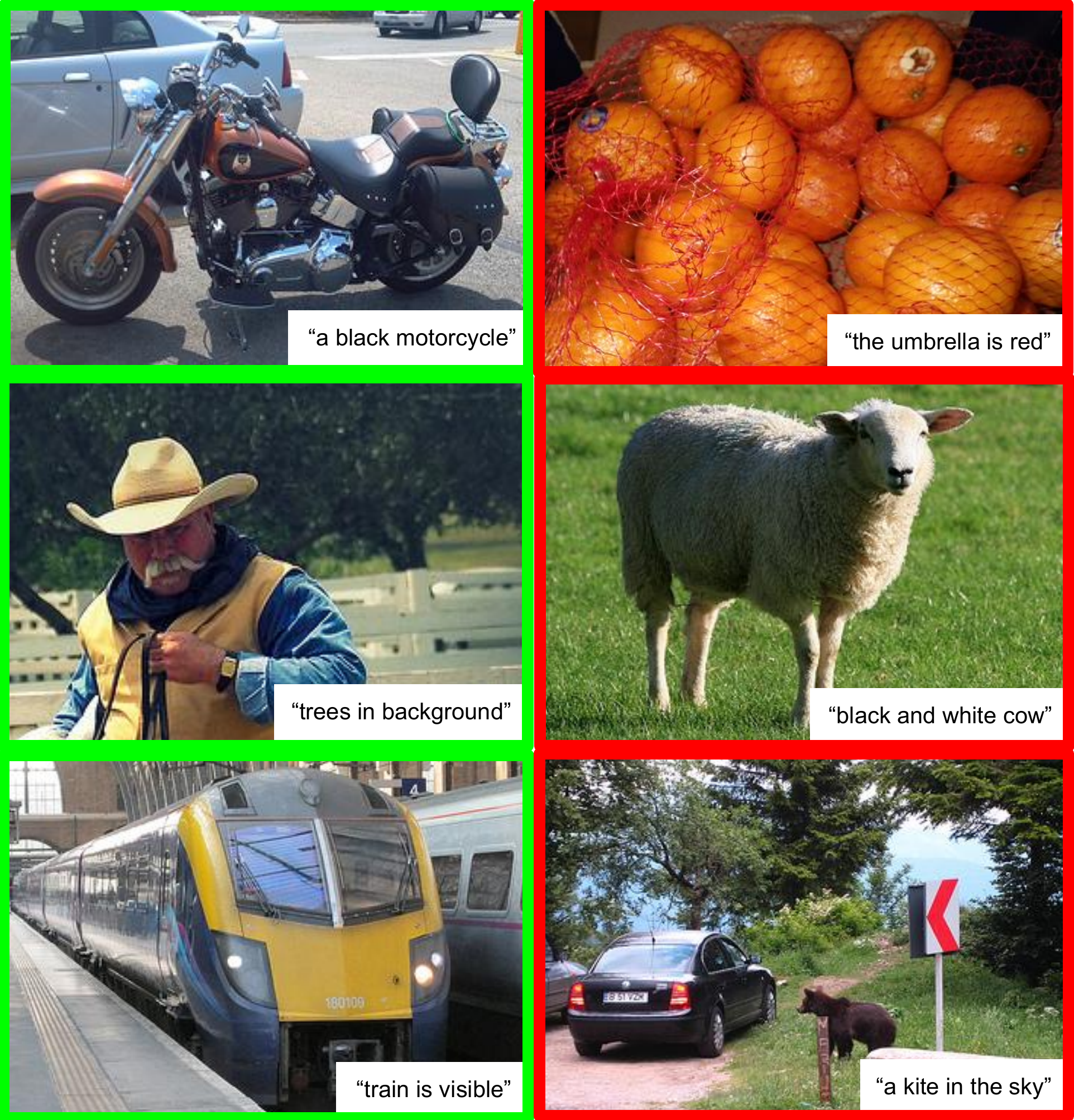}
}{
    \includegraphics[width=0.46\textwidth]{png_graphics/description_generation_examples.png}
}
\caption{Example predictions from the region description generation experiment. Regions in the first column (left) accurately describe the region, and those in the second column (right) are incorrect and unrelated to the corresponding region.}
\label{fig:description_generation_examples}
\end{figure}


Generating sentence descriptions of images has gained popularity as a task in computer vision~\cite{kiros2014multimodal, mao2014explain, karpathy2014deep, vinyals2014show}; however, current state-of-the-art models fail to describe all the different events captured in an image and instead provide only a high-level summary of the image.
In this section, we test how well state-of-the-art models can caption the details of images. For both experiments, we use the NeuralTalk model~\cite{karpathy2014deep}, since it not only provides state-of-the-art results but also is shown to be robust enough for predicting short descriptions. We train NeuralTalk on the Visual Genome dataset for region descriptions and on Flickr30K~\cite{young2014image} for full sentence descriptions.
As a model trained on other datasets would generate complete sentences and would not be comparable~\cite{chen2015microsoft} to our region descriptions, we convert all region descriptions generated by our model into complete sentences using predefined templates~\cite{hou2002template}.

\paragraph{Setup.} For training, we begin by preprocessing region descriptions; we remove all non-alphanumeric characters and lowercase and strip excess whitespace from them. We have $4,158,841$ region descriptions in total. We end up with $3,150,000$ region descriptions for training -- $504,420$ each for validation and testing. Note that we ensure descriptions of regions from the same image are exclusively in the training, validation, or testing set. We feed the bounding boxes of the regions through the pretrained VGG 16-layer network~\cite{simonyan2014very} to get the 4096-dimensional feature vectors of each region. We then use the NeuralTalk~\cite{karpathy2014deep} model to train a long short-term memory (LSTM) network~\cite{hochreiter1997long} to generate descriptions of regions. We use a learning rate of $0.001$ trained with rmsprop~\cite{dauphin2015rmsprop}. The model converges after four days.

For testing, we crop the ground-truth region bounding boxes of images and extract their 4096-dimensional 16-layer VGG network~\cite{simonyan2014very} features. We then feed these vectors through the pretrained NeuralTalk model to get predictions for region descriptions. 

\paragraph{Results.} Table~\ref{tab:description_generation_results} shows the results for the experiment. We calculate BLEU, CIDEr, and METEOR scores~\cite{chen2015microsoft} between the generated descriptions and their ground-truth descriptions. In all cases, the model trained on VisualGenome performs better. Moreover, we asked crowd workers to evaluate whether a generated description was correct---we got $1.6\%$ and $43.03\%$ for models trained on Flickr30K and on Visual Genome, respectively. The large increase in accuracy when the model trained on our data is due to the specificity of our dataset. Our region descriptions are shorter and cover a smaller image area. In comparison, the Flickr30K data are generic descriptions of entire images with multiple events happening in different regions of the image. The model trained on our data is able to make predictions that are more likely to concentrate on the specific part of the image it is looking at, instead of generating a summary description. The objectively low accuracy in both cases illustrates that current models are unable to reason about complex images.

Figure~\ref{fig:description_generation_examples} shows examples of regions and their predicted descriptions. Since many examples have short descriptions, the predicted descriptions are also short as expected; however, this causes the model to fail to produce more descriptive phrases for regions with multiple objects or with distinctive objects (i.e.\ objects with many attributes). While we use templates to convert region descriptions into sentences, future work can explore smarter approaches to combine region descriptions and generate a paragraph connecting all the regions into one coherent description.

\begin{table*}[t]
\centering
\begin{tabular}{lccccccc}
& BLEU-1 & BLEU-2 & BLEU-3 & BLEU-4 & CIDEr & METEOR & Human \\ 
 \midrule
Flickr8K & 0.09 & 0.01 & 0.002 & 0.0004 & 0.05 & 0.04 & 1.6\% \\
VG & 0.17 & 0.05 & 0.02 & 0.01 & 0.30 & 0.09 & 43.03\% \\
\bottomrule
\end{tabular}
\caption{Results for the region description generation experiment. Scores in the first row are for the region descriptions generated from the NeuralTalk model trained on Flickr8K, and those in the second row are for those generated by the model trained on Visual Genome data. BLEU, CIDEr, and METEOR scores all compare the predicted description to a ground truth in different ways.}
\label{tab:description_generation_results}
\vspace{1.5em}
\end{table*}

\subsection{Question Answering}
\label{sec:answer_generation}
Visual Genome is currently the largest dataset of visual question answers with $1.7$ million question and answer pairs. Each of our $108,249$ images contains an average of $17$ question answer pairs. Answering questions requires a deeper understanding of an image than generic image captioning. Question answering can involve fine-grained recognition (e.g. ``What is the breed of the dog?''), object detection (e.g. ``Where is the kite in the image?''), activity recognition (e.g. ``What is this man doing?''), knowledge base reasoning (e.g. ``Is this glass full?''), and common-sense reasoning (e.g. ``What street will we be on if we turn right?'').

By leveraging the detailed annotations in the scene graphs in Visual Genome, we envision building smart models that can answer a myriad of visual questions. While we encourage the construction of smart models, in this paper, we provide some baseline metrics to help others compare their models.

\paragraph{Setup.} We split the QA pairs into a training set ($60\%$) and a test set ($40\%$). We ensure that all images are exclusive to either the training set or the test set. We implement a simple baseline model that relies on answer frequency. The model counts the top $k$ most frequent answers (similar to the ImageNet challenge~\cite{ILSVRC15}) in the training set as the predictions for all the test questions, where $k=100$, $500$, and $1000$. We let a model make $k$ different predictions. We say the model is correct on a QA if one of the $k$ predictions matches exactly with the ground-truth answer. We report the accuracy over all test questions. This evaluation method works well when the answers are short, especially for single-word answers. However, it causes problems when the answers are long phrases and sentences. Other evaluation methods require word ontologies~\cite{malinowski2014multi}, multiple choices~\cite{antol2015vqa,VisualMadlibs}, or human judges~\cite{gao2015you}.

\begin{table}[t!]
\centering
\begin{tabular}{l c c c}
 & top-100 & top-500 & top-1000\\
\midrule
What & 0.420 & 0.602 & 0.672\\
Where & 0.096 & 0.324 & 0.418\\
When & 0.714 & 0.809 & 0.834\\
Who & 0.355 & 0.493 & 0.605\\
Why & 0.034 & 0.118 & 0.187\\
How & 0.780 & 0.827 & 0.846\\
\midrule
Overall\qquad & 0.411 & 0.573 & 0.641\\
\bottomrule
\end{tabular}
\caption{Baseline QA performances (in accuracy).}
\label{tab:qa-baseline}
\end{table}

\paragraph{Results.} Table~\ref{tab:qa-baseline} shows the performance of the open-ended visual question answering task. These baseline results imply the long-tail distribution of the answers. Long-tail distribution is common in existing QA datasets as well~\cite{antol2015vqa, malinowski2014multi}. The top 100, 500, and 1000 most frequent answers only cover $41.1\%$, $57.3\%$, and $64.1\%$ of the correct answers. In comparison, the corresponding sets of frequent answers in VQA~\cite{antol2015vqa} cover $63\%$, $75\%$, and $80\%$ of the test set answers. The ``where'' and ``why'' questions, which tend to involve spatial and common sense reasoning, tend to have more diverse answers and hence perform poorly, with performances of $0.096\%$ and $0.024\%$ top-100 respectively. The top 1000 frequent answers cover only $41.8\%$ and $18.7\%$ of the correct answers from these two question types respectively.


\section{Future Applications}
We have analyzed the individual components of this dataset and presented experiments with baseline results for tasks such as attribute classification, relationship classification, description generation, and question answering. There are, however, more applications and experiments for which our dataset can be used. In this section, we note a few potential applications that our dataset can enable.

\paragraph{Dense image captioning.} We have seen numerous image captioning papers~\cite{kiros2014multimodal, mao2014explain, karpathy2014deep, vinyals2014show} that attempt to describe an entire image with a single caption. However, these captions do not exhaustively describe every part of the scene. An natural extension to this application, which the Visual Genome dataset enables, is the ability to create dense captioning models that describe parts of the scene.

\paragraph{Visual question answering.} While visual question answering has been studied as a standalone task~\cite{VisualMadlibs,ren2015image,antol2015vqa,gao2015you}, we introduce a dataset that combines all of our question answers with descriptions and scene graphs. Future work can build supervised models that utilize various components of Visual Genome to tackle question answering.

\paragraph{Image understanding.} While we have seen a surge of image captioning~\cite{kiros2014multimodal} and question answering~\cite{antol2015vqa} models, there has been little work on creating more comprehensive evaluation metrics to measure how well these models are performing. Such models are usually evaluated using BLEU, CIDEr, or METEOR and other similar metrics that do not effectively measure how well these models understand the image~\cite{chen2015microsoft}. The Visual Genome scene graphs can be used as a measurement for image understanding. Generated descriptions and answers can be matched against the ground truth scene graph of an image to evaluate its corresponding model.

\paragraph{Relationship extraction.} Relationship extraction has been extensively studied in information retrieval and natural language processing~\cite{zhou122007tree, guodong2005exploring, culotta2004dependency, socher2012semantic}. Visual Genome is the first large-scale visual relationship dataset. This dataset can be used to study the extraction of visual relationships\cite{sadeghi2015viske} from images, and its interactions between objects can also be used to study action recognition~\cite{yao2010modeling, ramanathan2015learning} and spatial orientation between objects~\cite{gupta2009observing, prest2012weakly}.

\paragraph{Semantic image retrieval.} Previous work has already shown that scene graphs can be used to improve semantic image search~\cite{Johnson2015CVPR, schustergenerating}. Further methods can be explored using our region descriptions combined with region graphs. Attention-based search methods can also be explored where the area of interest specified by a query is also localized in the retrieved images.

\section{Conclusion}
\label{sec:conlusion}


Visual Genome provides a multi-layered understanding of pictures. It allows for a multi-perspective study of an image, from pixel-level information like objects, to relationships that require further inference, and to even deeper cognitive tasks like question answering. It is a comprehensive dataset for training and benchmarking the next generation of computer vision models. With Visual Genome, we expect these models to develop a broader understanding of our visual world, complementing computers' capacities to detect objects with abilities to describe those objects and explain their interactions and relationships. Visual Genome is a large \textit{formalized knowledge representation} for visual understanding and a more \textit{complete set of descriptions and question answers} that \textit{grounds visual concepts to language}.

\begin{acknowledgements}
We would like to start by thanking our sponsors: Stanford Computer Science Department, Yahoo Labs!, The Brown Institute for Media Innovation, Toyota and Adobe. Next, we specially thank Michael Stark, Yutian Li, Frederic Ren, Sherman Leung, Michelle Guo and Gavin Mai for their contributions. We thank Carsten Rother from the University of Dresden for facilitating Oliver Groth's involvement. We also thank all the thousands of crowd workers for their diligent contribution to Visual Genome. Finally, we thank all members of the Stanford Vision Lab and HCI Lab for their useful comments and discussions.
\end{acknowledgements}

\bibliographystyle{apalike}      

\bibliography{egbib}   

\end{document}